%% file: main.tex
\documentclass{article} 
\usepackage{collas2022_conference,times}

\usepackage{etoolbox}
\makeatletter
\patchcmd{\maketitle}
 {\def\@makefnmark}
 {\def\@makefnmark{}\def\useless@macro}
 {}{}
\makeatother

\usepackage{microtype}      
\usepackage{microtype}
\usepackage{graphicx}
\usepackage{amsfonts}       
\usepackage{booktabs} 
\usepackage{multirow}
\usepackage{caption}
\usepackage{hyperref}
\usepackage{url}
\usepackage{amsmath}
\usepackage{bbm}
\usepackage{todonotes}
\usepackage[ruled,vlined]{algorithm2e}
\usepackage{bbm}
\usepackage{color,soul} 
\usepackage{float}
\usepackage{subcaption}
\usepackage{soul}

\input{math_commands.tex}

\title{Improving Meta-Learning generalization \\with activation-based early-stopping}

\author{Simon Guiroy\\
Mila - Quebec AI Institute\\
Universit\'e de Montr\'eal\thanks{Correspondence to: Simon Guiroy, \textless \href{mailto:simon.guiroy@umontreal.ca}{simon.guiroy@umontreal.ca}\textgreater \newline Code available at: \url{https://github.com/simonguiroy/ABE}}\\
\And 
Christopher Pal\\
Mila - Quebec AI Institute\\
Polytechnique Montr\'eal\\
CIFAR AI Chair\\
\And 
\hspace{-1mm}Gon\c{c}alo Mordido\\
\hspace{-1mm}Mila - Quebec AI Institute\\
\hspace{-1mm}Polytechnique Montr\'eal\\
\And
Sarath Chandar\\
Mila - Quebec AI Institute\\
Polytechnique Montr\'eal\\
CIFAR AI Chair\\
}

\collasfinalcopy 

\begin{document}

\maketitle

\begin{abstract}
        Meta-Learning algorithms for few-shot learning aim to train neural networks capable of generalizing to novel tasks using only a few examples. Early-stopping is critical for performance, halting model training when it reaches optimal generalization to the new task distribution. Early-stopping mechanisms in Meta-Learning typically rely on measuring the model performance on labeled examples from a meta-validation set drawn from the training (source) dataset. This is problematic in few-shot transfer learning settings, where the meta-test set comes from a different target dataset (OOD) and can potentially have a large distributional shift with the meta-validation set. In this work, we propose Activation Based Early-stopping (ABE), an alternative to using validation-based early-stopping for meta-learning. Specifically, we analyze the evolution, during meta-training, of the neural activations at each hidden layer, on a small set of unlabelled support examples from a single task of the target tasks distribution, as this constitutes a minimal and justifiably accessible information from the target problem. Our experiments show that simple, label agnostic statistics on the activations offer an effective way to estimate how the target generalization evolves over time. At each hidden layer, we characterize the activation distributions, from their first and second order moments, then further summarized along the feature dimensions, resulting in a compact yet intuitive characterization in a four-dimensional space. Detecting when, throughout training time, and at which layer, the target activation trajectory diverges from the activation trajectory of the source data, allows us to perform early-stopping and improve generalization in a large array of few-shot transfer learning settings, across different algorithms, source and target datasets.
\end{abstract}

\section{Introduction}
\begin{figure}[ht]
    \centering
    \includegraphics[width=0.75\linewidth]{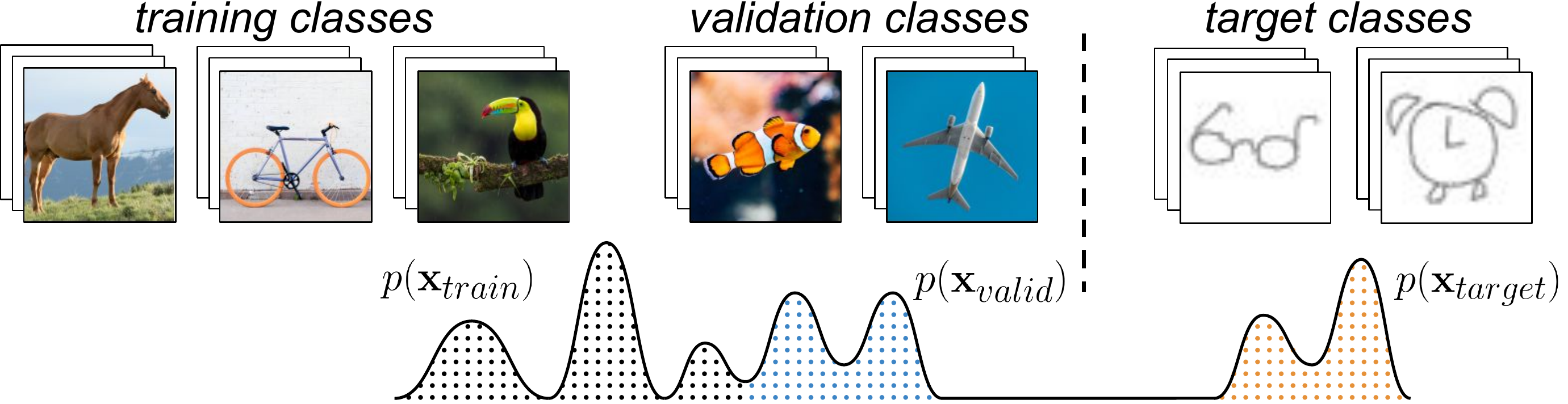}
    \caption{A \textit{task} is created by randomly picking a subset of classes from a set $\mathcal{C}$, and belongs to a \textit{task distribution} $p(\mathcal{T})$ (e.g. $\mathcal{T}_i^{train}\sim p(\mathcal{T}_{train})$ : classify between ``horse'' and ``bicycle''). Training, validation, and target tasks are made of different classes. In the framework of \textit{few-shot classification}, there is a limited amount of labeled examples to solve each target task. In \textit{few-shot transfer learning}, not only the target classes, but also the target inputs are drawn from a different data distribution than the training data.}
    \label{fig:few_shot_transfer_learning}
\end{figure}

Machine learning research has been successful at producing algorithms and models that, when optimized on a distribution of training examples, generalize well to previously unseen examples drawn from that same distribution. Meta-learning is in a way, a natural extension of this aim, where the model has to generalize to not only new data points but entirely new tasks.
Meta-learning algorithms generally aim to train a model $f(\mathbf{x}; \theta)$ on a set of source problems, often presented as a distribution over tasks $p(\mathcal{T}_{train})$, in such a way that the model is capable of generalizing to new, previously unseen tasks from a target distribution $p(\mathcal{T}_{target})$. When applied to classification, Meta-Learning has often been formulated in the past by defining a task $\mathcal{T}$ that involves the $m$-way classification of input examples $\mathbf{x}$ among $m$ distinct classes. The tasks from $p(\mathcal{T}_{train})$ and $p(\mathcal{T}_{target})$ are made of classes drawn from two disjoint sets $\mathcal{C}_{train}$ and $\mathcal{C}_{target}$. A novel task thus involves new classes not seen during training. In \textit{few-shot transfer learning}, not only are the class sets $\mathcal{C}_{train}$ and $\mathcal{C}_{target}$ disjoint, but the marginal $p(\mathbf{x}_{target})$ can be arbitrarily different from $p(\mathbf{x}_{train})$, \textit{e.g.} from a different image dataset. This is illustrated in Fig.~\ref{fig:few_shot_transfer_learning}.

Important practical progress has been made in meta-learning over the past few years \citep{Lake2015HumanlevelCL,Koch2015SiameseNN,DBLP:journals/corr/AndrychowiczDGH16,Rezende:2016:OGD:3045390.3045551,pmlr-v48-santoro16,DBLP:journals/corr/MishraRCA17,prototypical,DBLP:journals/corr/VinyalsBLKW16,DBLP:journals/corr/abs-1711-06025,2017arXiv171104043G,DBLP:journals/corr/FinnAL17,DBLP:journals/corr/abs-1810-03642,DBLP:journals/corr/abs-1904-03758,sachin,DBLP:journals/corr/WichrowskaMHCDF17,2018arXiv180400222M,2019arXiv190203356P,Maclaurin:2015:GHO:3045118.3045343,2018arXiv180604910F,Lian2020Towards}. Yet, it remains sparsely understood what are the underlying phenomena behind the transitioning of a neural network's generalization to novel tasks, from the underfitting to the overfitting regime. Early-stopping, a fundamental element of machine learning practice, maximizes generalization by aiming to halt the training at the frontier between those two regimes when generalization is optimal. 

Usually, the optimal stopping point is computed on a validation set, made of held-out examples from the training data, which serves as a proxy for the test data. However, in meta-learning, adopting such an early-stopping strategy may be problematic due to the arbitrarily large distributional shift between the meta-validation tasks and the meta-test tasks, prevalent in few-shot transfer learning settings. While this problem has been acknowledged in a related field studying out-of-distribution generalization \citep{2020arXiv200701434G}, performing early-stopping based on validation set performance remains the \textit{de facto} approach in meta-learning.

In this work, we propose an alternative to using validation-based early-stopping for meta-learning. Specifically, we analyze the evolution, during meta-training, of the neural activations on a small set of unlabelled examples from the target input distribution. To illustrate the shortcomings of using validation-based metrics for early-stopping, we focus on few-shot transfer learning settings, where the target input data may differ significantly from the source data. Our experiments show that simple, label agnostic statistics on the activations offer an effective way to estimate how the target generalization evolves over time, and detecting when the target activation trajectory diverges from the source trajectory allows us to perform early-stopping.

Our main contributions may be summarized as follows:

1. We propose a novel method for early-stopping in Meta-Learning based on the observation that neural activation dynamics on a few unlabelled target examples may be used to infer generalization capabilities (Sec.~\ref{sec:neural_activation_dynamics}).

2. We empirically show the superiority of using our method for early-stopping compared to validation set performance. We improve the overall generalization performance of few-shot transfer learning on multiple image classification datasets (Sec.~\ref{sec:performance}). On average, our method closes 47.8\% of the generalization gap between the validation baseline and optimal performance. Notably, when the baseline generalization gap is larger (baseline performance is $<$ 0.95 $\times$ optimal accuracy), we close the generalization gap by 69.2\%. and when this gap is already small (baseline performance $\geq$ 0.95 $\times$  optimal accuracy), we still manage to slightly outperform the validation baseline, closing the gap by 14.5\% on average.

\section{Motivation}
\label{sec:issue_meta-val}

In \textit{few-shot classification}, the inputs $\mathbf{x}$ of the training and target tasks originate from the same distribution $p(\mathbf{x})$, \textit{e.g.} an image dataset, but conditioned on different training and target classes: $p(\mathbf{x}_{train}) = p(\mathbf{x} | \mathbf{y} \in \mathcal{C}_{train})$ and $p(\mathbf{x}_{target}) = p(\mathbf{x} | \mathbf{y} \in \mathcal{C}_{target})$, respectively. The \textit{few-shot} aspect means that, for a given new task $\mathcal{T}_{target}$, only a very few labelled examples are available, typically $k$ examples for each of the $m$ classes. The model, then, uses this \textit{support set} $\mathcal{S} = \{(\mathbf{x}, \mathbf{y})\}_{1..k}$ (of $m \times k$ examples) to adapt its parameters $\theta$ to $\mathcal{T}_{target}$. Finally, its accuracy is evaluated on new \textit{query} examples from $\mathcal{T}_{target}$. 
Considering multiple target tasks from $p(\mathcal{T}_{target})$, the Meta-Learning \textit{generalization} $Acc_{target}$ for a model $f(\mathbf{x}; \theta_t)$ after $t$ training iterations is its average query accuracy:
\begin{equation}
\begin{array}{c}
    Acc_{target} \doteq 
    \mathop{\mathbb{E}}_{\mathcal{T}_i \sim p(\mathcal{T}_{target})} 
    \left[ \mathop{\mathbb{E}}_{(\mathbf{x}, \mathbf{y}) \sim \mathcal{T}_i \setminus \mathcal{S}_i} \left[ \mathbbm{1} \{ \mathrm{argmax}(f(\mathbf{x};\theta_t^i)) = \mathbf{y} \} \right] \right] \, ,
\end{array}\label{eq:generalization}
\end{equation} 
where, for each new task $\mathcal{T}_i$, the adapted solution $\theta_t^i$ may be obtained by performing $T$ steps of gradient descent (full-batch) on the cross-entropy loss $\mathcal{L}(f, \mathcal{S}_i)$ with respect to $\theta_t$. The time of optimal generalization is defined as $t^* = \mathop{\mathrm{argmax}}_t Acc_{target}(t)$. In a standard supervised learning setup, a subset of examples is held out from the training data to constitute a validation set. Since the validation accuracy is a good proxy for the test accuracy, early-stopping is performed by halting training when the validation accuracy reaches its maximum. In Meta-Learning for few-shot classification, the validation set is made of held out classes from the training data to constitute the validation task distribution $p(\mathcal{T}_{valid})$. Using validation set performance, the early-stopping time for optimal generalization, $t^*$, is estimated by $t^*_{valid} = \mathrm{argmax}_t \, Acc_{valid}$. 

However in transfer learning, $Acc_{valid}$ may not be a good proxy for $Acc_{target}$ due to the potentially large distributional shift between $p(\mathcal{T}_{target})$ and $p(\mathcal{T}_{valid})$. In the end, this may lead a potentially large difference between $t^*$ and $t^*_{valid}$, and thus to sub-optimal generalization (as later shown in Sec.~\ref{sec:performance}).
Hence, estimating $t^*$ some minimal amount of information about $p(\mathcal{T}_{target})$. However, the availability of data from $p(\mathcal{T}_{target})$ is often highly limited, introducing the few-shot learning paradigm. Since the model doesn't control how many new tasks will actually be presented, as there could be either several thousands or very few, the model will need to solve at the very least a single task from $p(\mathcal{T}_{target})$, and thus has access to at least a single support set $\mathcal{S}$. Moreover, since labelled data is often more difficult to access, our method only uses the few examples from a single target task, without their labels. We thus propose to only use a few examples: the support set of a single new task (\textit{e.g.} 5 images). Our end goal is to improve overall generalization to tasks from $p(\mathcal{T}_{target})$. It is important to note that the target task used to estimate $t^*$ is excluded from $p(\mathcal{T}_{target})$ to perform this evaluation. For the justification of why it is reasonable to have access to a few unlabelled support examples from the target distribution, we can think a real-world examples. Consider an application with an autonomous car that can detect signs in North America, trained on data there, then these cars are deployed in another country. They have collected data, but you have not yet labelled it, and it is unlabelled data. The model could thus use those examples to decide, using our proposed method, which of the checkpoints of the state of the neural network should be picked. For further motivation on the necessity of using a minimal hint of information about the target task distribution, see App. \ref{sec:appendix:exp_results:issue_meta-val}.

\section{Estimating generalization via neural activation dynamics}
\label{sec:neural_activation_dynamics}

\begin{figure}[H]
    \centering
    \subfloat[For a deep neural network $f$ composed of a feature extractor $\varphi$ of $L$ hidden-layers (followed by a classifier $g$), and a given input distribution $p(\mathbf{x})$, we analyze the neural activation distributions and track how their trajectories evolve during training.]{\includegraphics[width=0.37\textwidth]{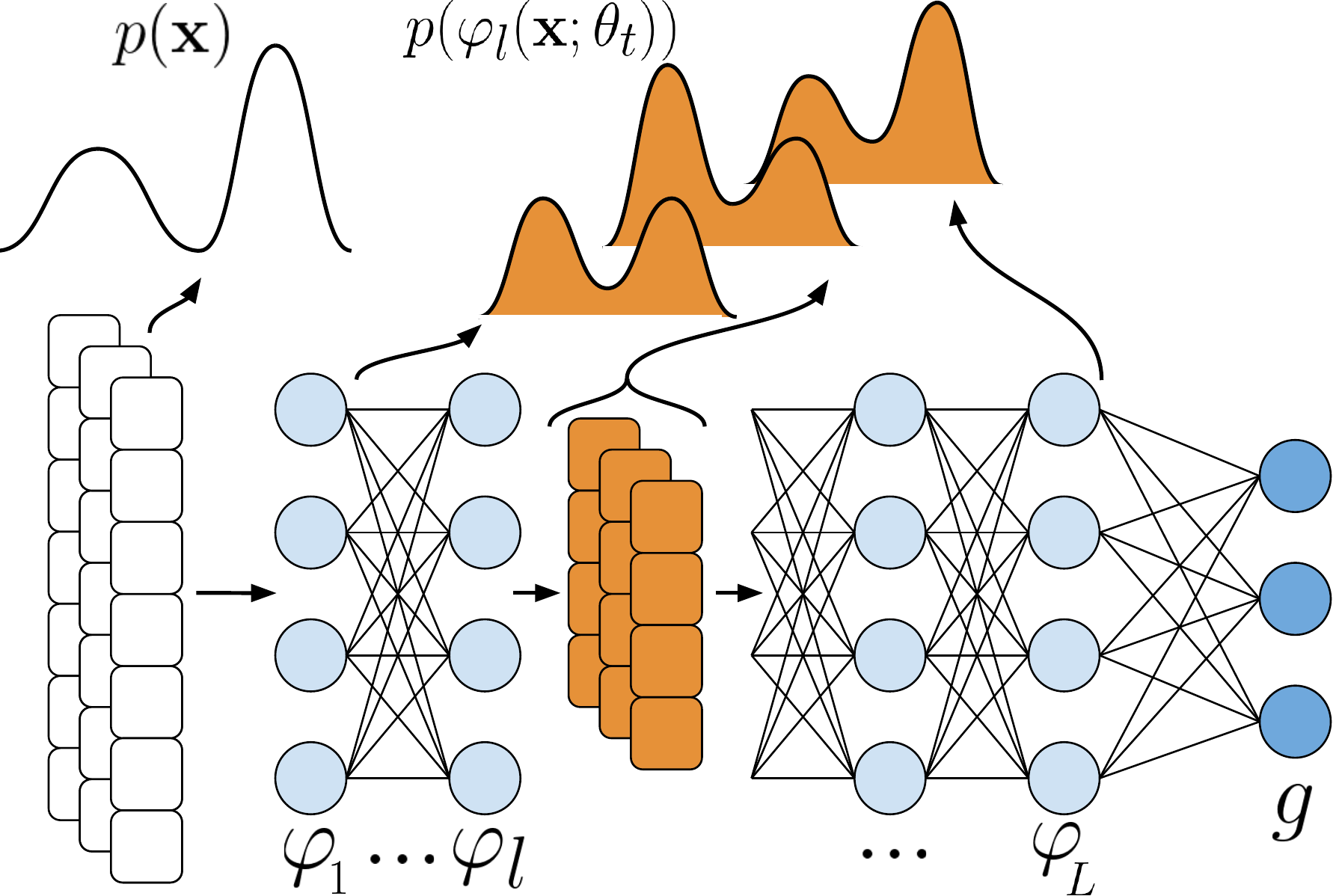}}\quad
    \subfloat[We characterize the activation distributions from their first and second order statistical moments. and project their trajectory in $\psi(\mathbf{x},t)$.
    We hypothesize that time $\hat{t}$, defined when the target trajectory $\psi(\mathbf{x}_{target},t)$ diverges from the source trajectory $\psi(\mathbf{x}_{valid},t)$ (left figure), coincides with the time $t^*$, defined when the target accuracy starts to degrade while the validation accuracy still increases (right figure). Our method uses $\hat{t}$ as an estimation of $t^*$ to halt training.
    ]{\includegraphics[width=0.6\textwidth]{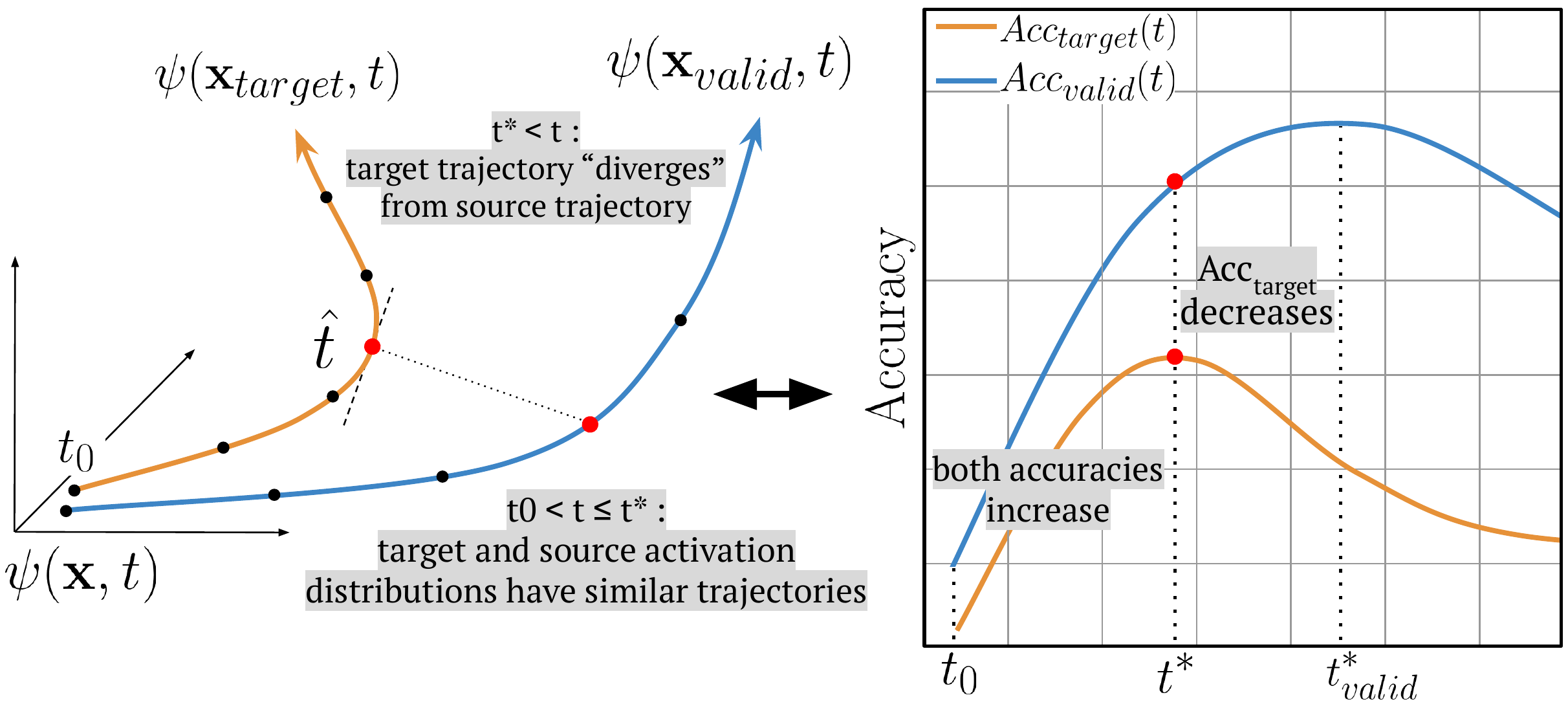}}
    \caption{Activation-Based Early-stopping (ABE).}
    \label{fig:trajectory_divergence}
\end{figure}

In this work, we propose to leverage observable properties based on the neural activations in deep neural networks to infer Meta-Learning generalization to a given target problem as training time $t$ progresses.
More specifically, we want to estimate $t^* = \mathop{\mathrm{argmax}}_t Acc_{target}(t)$ using only a few target examples, \textit{i.e.} the support set of a single task, to accommodate for few-shot learning settings. Our goal is to improve generalization when validation-based early-stopping is suboptimal, particularly in transfer learning scenarios that require out-of-distribution generalization.

Our experiments in a wide range of Meta-Learning settings for few-shot classification suggest that, for a deep neural network $f$ with a feature extractor $\varphi$ of $L$ hidden-layers, the generalization $Acc_{target}$ to a novel task distribution $p(\mathcal{T}_{target})$ can often be estimated from the distribution of neural activations with respect to the target input data from $p(\mathbf{x}_{target})$.
In deep learning, the feature extractor is responsible for learning deep representations of the data, often comprising the majority of the neural architecture, hence it makes sense that it is critical for the generalization performance. Other authors made similar observations \citep{2019arXiv190909157R,DBLP:journals/corr/abs-2002-06753} regarding the feature extractor.
Since the neural activations characterize the feature extractor as a function of the input, we propose to study the trajectory of their evolution to analyze how the feature extractor changes w.r.t the target data. We empirically observe that $Acc_{target}(t)$ is often linked to the dynamics throughout training time $t$ of the neural activations distribution $p(\varphi_l(\mathbf{x}_{target}; \theta_t))$ at a specific layer $l$ of $\varphi$, which we call a \textit{critical layer}. Such layers may be at a different network depth depending on the experimental settings, \textit{e.g.} the meta-training algorithm, architecture, or source and target datasets. This bears similarities to what \cite{DBLP:journals/corr/YosinskiCBL14} observed for deep networks in transfer learning (not few-shot). In few-shot transfer learning, our proposed method of Activation-Based Early-stopping consists of halting the model training when we detect that the trajectory of target activations diverges from the source trajectory.

\subsection{Characterizing the trajectory of neural activations dynamics}

To model the trajectory of the distribution of neural activations, we characterize this distribution based on its sequence of moments. The probability distribution of a random variable, under certain conditions (Carleman's condition), can be uniquely determined by its sequence of moments \citep{2019arXiv191200160Y}, meaning that no other distribution has the same moment sequence (e.g. a Gaussian is uniquely determined by its first and second-order moments). For a random vector, there are very few known examples of multivariate distributions that are not determined by their moment sequence, according to \cite{KLEIBER20137}. For an arbitrary random variable $Z$ with probability distribution $p(z)$, its n-th order moment is defined as  : $\mathbf{m}_n(z) = \mathbb{E}_{p(z)}[z^n]$. For a random vector $\mathbf{z}$ in $D$ dimensions, the sequence of moments $(\mathbf{m}_1(\mathbf{z}), \mathbf{m}_2(\mathbf{z}), ...)$ is made of the first-order moments, which are the elements of the mean vector $\mathbb{E}[\mathbf{z}]$, followed by the second-order moments which are the elements of the autocorrelation matrix $\mathbb{E}[\mathbf{z}\mathbf{z}^{\mathrm{T}}]$, and so on:

\begin{equation}
\mathbf{m}_1(\mathbf{z}) \;,\; \mathbf{m}_2(\mathbf{z}) \;,\; ...  \;\; = \;\;  \mathbb{E}_{p(\mathbf{z})}[\mathbf{z}] \;,\; \mathbb{E}_{p(\mathbf{z})}[\mathbf{z}\mathbf{z}^{\mathrm{T}}] \;,\; ... \;\; = \;\;
\mathbb{E}
\begin{bmatrix}
z_1 \\
z_2 \\
\vdots \\
z_D \end{bmatrix} ,\quad
\mathbb{E}
\begin{bmatrix}
z_1^2 & z_1z_2 & ... & z_1z_D \\
z_2z_1 & z_2^2 & ... & z_2z_D \\
\vdots  & \vdots & \ddots  & \vdots \\
z_Dz_1 & z_Dz_2 & ... & z_D^2 \\
\end{bmatrix} ,
\quad ...
\nonumber
\end{equation}
We observed that Meta-Learning generalization may be estimated from the first and second moments of the distribution of target activations. We therefore cut off the estimated moment sequence after the second-order moments. Moreover, higher-order moments are harder to estimate, bearing higher standard error, especially since we use only a few target examples for the estimations. Note that $\mathbf{m}_1 \in \mathbb{R}^D$ and $\mathbf{m}_2 \in \mathbb{R}^{D^2}$, and that we compute them on the activation vectors, which can be in very high dimensions $D$. To project the distribution trajectories in a space of reduced dimensionality, we aggregate those moments across their $D$ feature dimensions using summary statistics. 

Within the first and second order moments, we compute aggregate statistics on three distinct types: the elements of $\mathbb{E}[\mathbf{z}]$ (feature means), the diagonal elements of $\mathbb{E}[\mathbf{z}\mathbf{z}^{\mathrm{T}}]$ (raw feature variances), and the non-diagonal elements of $\mathbb{E}[\mathbf{z}\mathbf{z}^{\mathrm{T}}]$ (raw feature covariances). Particularly, for the first moment, we compute the empirical mean across feature dimensions ($\hat{m}_1)$ as well as the raw variance across feature dimensions ($\hat{m}_2$). For the second moments, we only compute the empirical mean for the diagonal and non-diagonal elements of $\mathbb{E}[\mathbf{z}\mathbf{z}^{\mathrm{T}}]$, \textit{i.e.} $\hat{m}_3$ and $\hat{m}_4$, respectively, since the variance of second moments leads to a higher-order statistic, which is harder to estimate. For an arbitrary random vector $\mathbf{z}$, we define the following \textit{aggregated moments}:
\begin{equation}
 \hat{m}_1(\mathbf{z}) = \sum_{i}[(\mathbf{m}_1(\mathbf{z}))_i],
    \hat{m}_2(\mathbf{z}) = \sum_i[(\mathbf{m}_1(\mathbf{z}))_i^2], 
    \hat{m}_3(\mathbf{z}) = \sum_{i=j}[(\mathbf{m}_2(\mathbf{z}))_{i,j}], \text{ and }
    \hat{m}_4(\mathbf{z}) = \sum_{i \neq j}[(\mathbf{m}_2(\mathbf{z}))_{i,j}]       
\end{equation}
 
where $(\mathbf{m}_1(\mathbf{z}))_i$ is the i-th feature of $\mathbf{m}_1(\mathbf{z})$, and $(\mathbf{m}_2(\mathbf{z}))_{i,j}$ is the feature at the i-th row and j-th column of $\mathbf{m}_2(\mathbf{z})$. We characterize as $\psi(\mathbf{x};t)$ the trajectory of neural activations dynamics, with respect to a given input distribution $p(\mathbf{x})$, using these aggregated moments on the activation vectors, i.e.  $\mathbf{z} = \varphi_l(\mathbf{x};t)$ and $\mathbf{x} \sim p(\mathbf{x})$, such that:
\begin{equation}
    \psi(\mathbf{x}, t) =
    \begin{bmatrix}
\hat{m}_1(\varphi_1(\mathbf{x}; t)) & \hat{m}_2(\varphi_1(\mathbf{x}; t)) & \hat{m}_3(\varphi_1(\mathbf{x}; t)) & \hat{m}_4(\varphi_1(\mathbf{x}; t)) \\
\hat{m}_1(\varphi_2(\mathbf{x}; t)) & \hat{m}_2(\varphi_2(\mathbf{x}; t)) & \hat{m}_3(\varphi_2(\mathbf{x}; t)) & \hat{m}_4(\varphi_2(\mathbf{x}; t)) \\
\vdots & \vdots & \vdots & \vdots \\
\hat{m}_1(\varphi_L(\mathbf{x}; t)) & \hat{m}_2(\varphi_L(\mathbf{x}; t)) & \hat{m}_3(\varphi_L(\mathbf{x}; t)) & \hat{m}_4(\varphi_L(\mathbf{x}; t)) \\
\end{bmatrix}.
\label{eq:collas:neural_activation_dynamics}
\end{equation}

Our experiments show that the target generalization $Acc_{target}(t)$ in few-shot learning is often proportional to a linear combination of the presented aggregated moments when computed on a few unlabelled examples of the target data $p(\mathbf{x}_{target})$. The function space spanned by the possible linear combinations of $\hat{m}_1$ to $\hat{m}_4$, while being simple to compute, is general enough to express many functions of the activation vectors, such as their average $l_2$ norm, pair-wise inner products, pair-wise $l_2$ distances, collective variance, and average variance between individual features. This suggests that there may be multiple factors linked to Meta-Learning generalization, and such factors are likely dependent on the experimental setting. In App.~\ref{sec:appendix:exp_results:neural_activation_dynamics} we show empirical motivation for how we characterize the neural activation dynamics, and for the need to consider all layers of the feature extractor.

\subsection{Activation-based early-stopping for few-shot transfer learning}

In few-shot transfer learning, since we cannot directly infer from $Acc_{valid}(t)$ which combination of the aggregated moments coincides with $Acc_{target}(t)$, we propose to compare the trajectory of the target activations, $\psi(\mathbf{x}_{target},t)$, to the trajectory of activations of the source data, $\psi(\mathbf{x}_{valid},t)$, within the time interval $t_0 < t \leq t^*_{valid}$, where $t_0$ is the initial training time and $t^*_{valid}$ is the time when the validation accuracy reaches its optimum. If at a given time $\hat{t}$ we detect that the target trajectory starts to diverge from the source trajectory, we halt training at $\hat{t}$. In other words, we perform early-stopping at time $\hat{t}$, which serves as an estimate of the optimal early-stopping time for optimal generalization ($t^*$).

The intuition for the proposed early-stopping criterion is that as the model is being trained, learned knowledge from the source data is only useful for solving the target tasks up to a certain time, which we refer to as $\hat{t}$. After $\hat{t}$, the learning becomes overspecialized to the source tasks, harming target generalization. As previously mentioned, our experiments indicate that the divergence in activation distribution trajectories is often most prominent at a particular layer $l^*$ of the feature extractor, which we refer to as the \textit{critical layer}. Moreover, we observed that the trajectory divergence is usually mainly driven by a specific aggregated moment $\hat{m}^*$ at the critical layer, and we refer to $\hat{m}^*$ as the \textit{critical moment}. 

\begin{figure}[H]
    \centering
    \subfloat[Evolution of the source and target activation trajectories (omitting $\hat{m}_4$ for 3D visualization). Here their divergence is  highest at layer 4, along $\hat{m}_2$. ]{\includegraphics[width=0.36\textwidth]{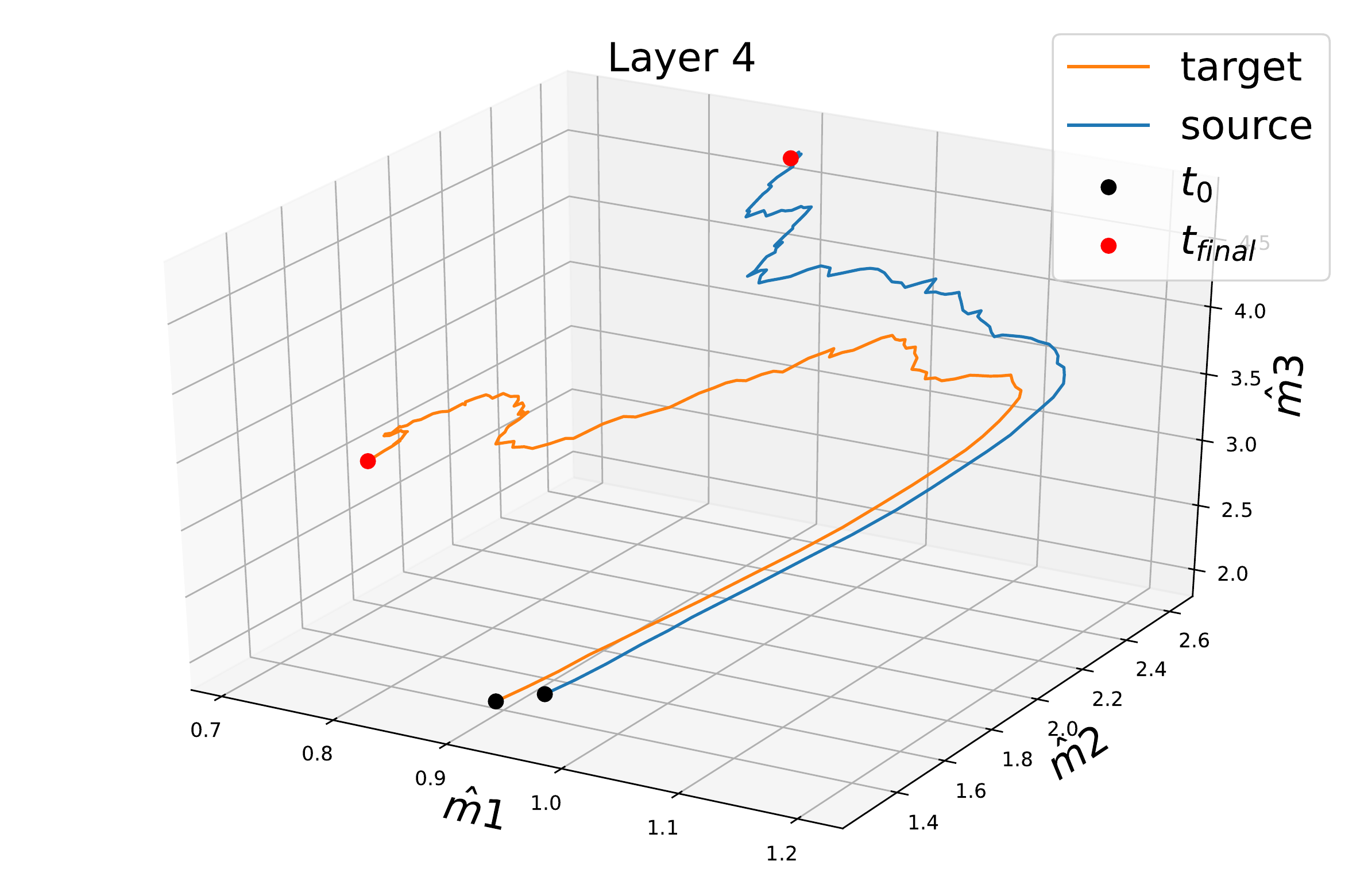}\label{fig:exp:divergence:finding_psi_and_l}.}
    \hskip 0.1in
    \subfloat[Evaluating the target and source activations trajectories at layer 4 and along $\hat{m}_2$, their divergence is detected at time  $\hat{t}$, after which their plots become negatively correlated.]{\includegraphics[width=0.28\textwidth]{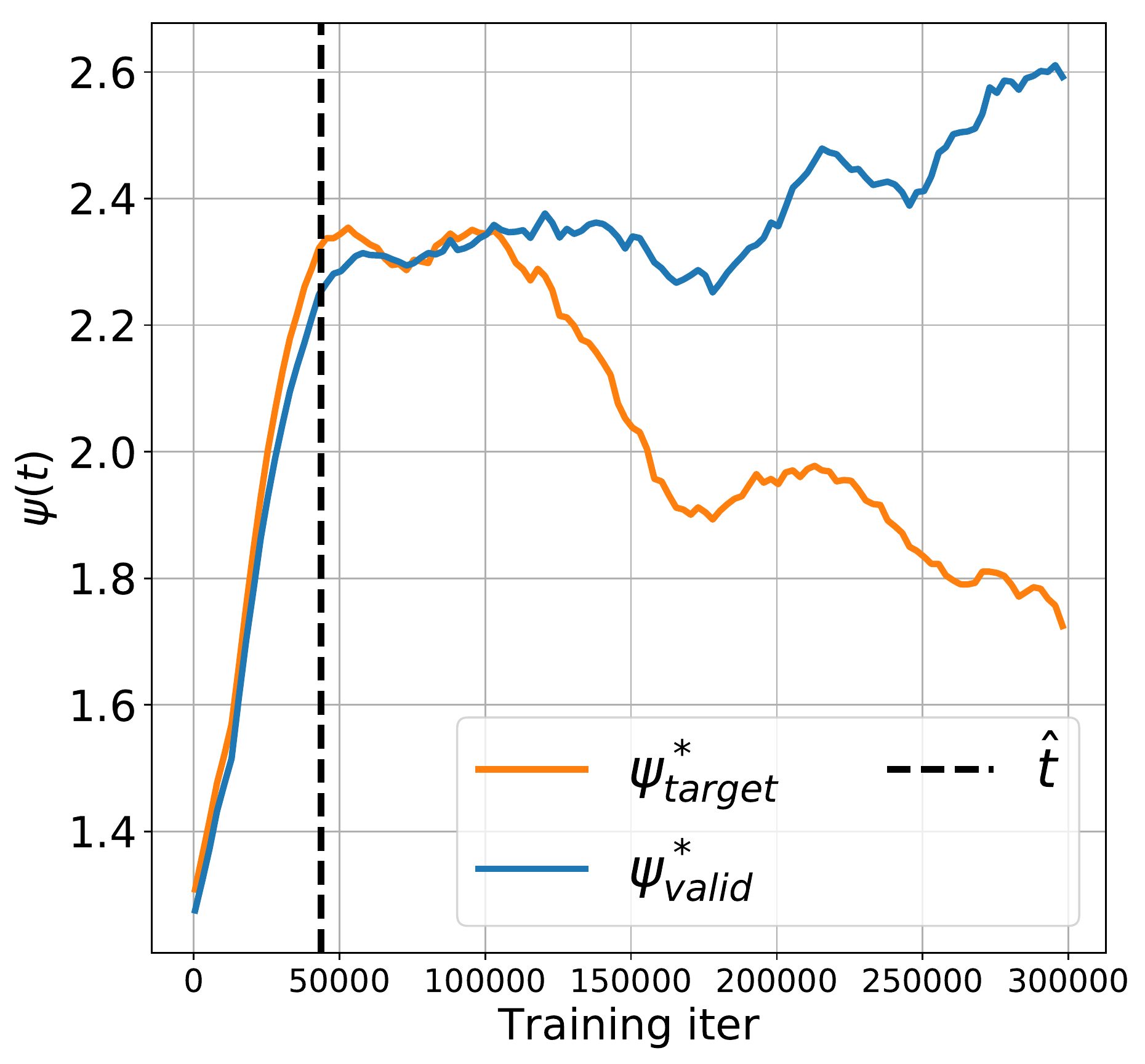}\label{fig:exp:divergence:finding_psi_and_l}}
    \hskip 0.1in
    \subfloat[ABE here improves the target generalization by early-stopping at time $\hat{t}$, as it is much closer to the true optimum $t^*$, compared to $t^*_{valid}$ of the validation accuracy.]{\includegraphics[width=0.285
    \textwidth]{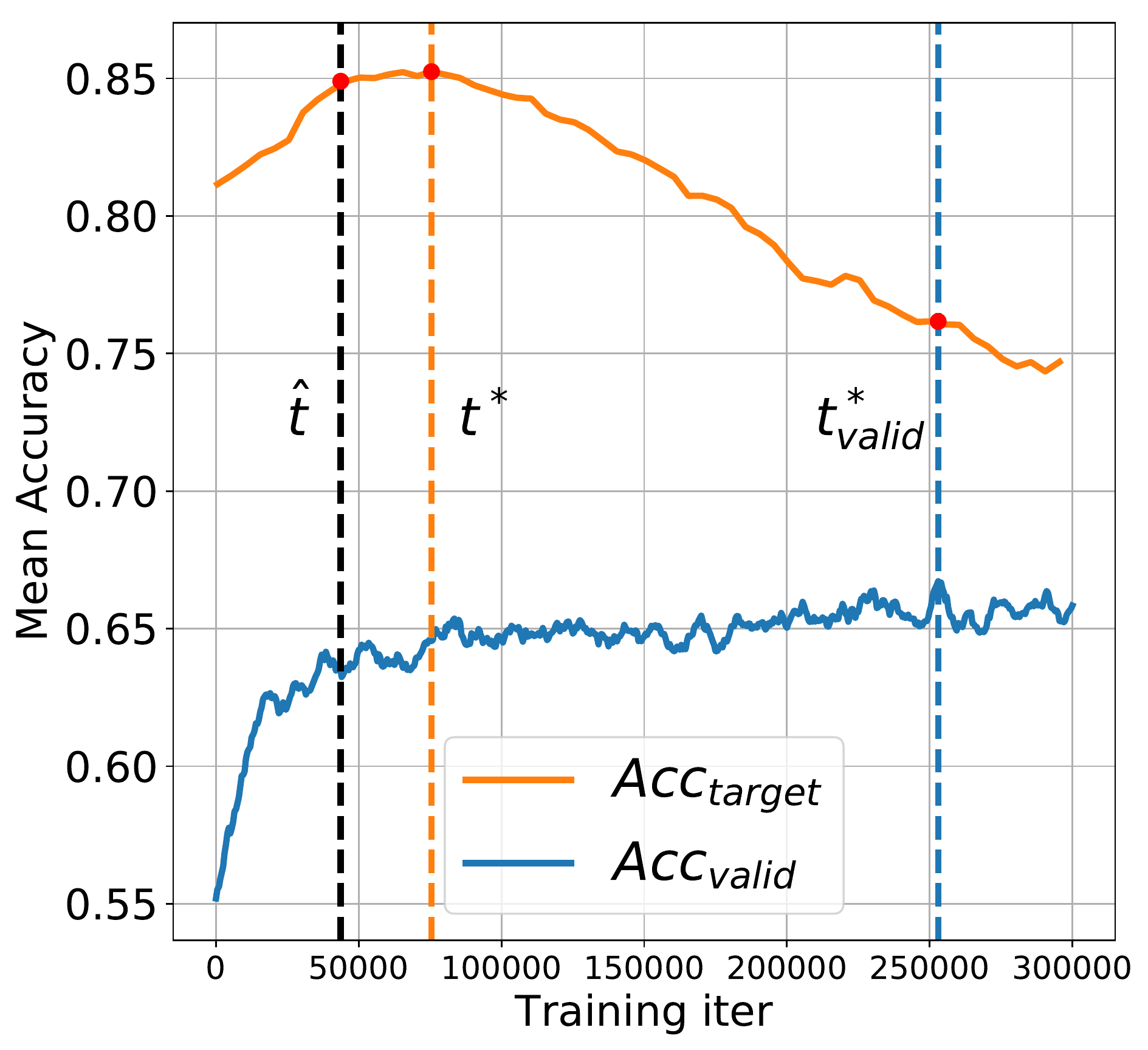}\label{fig:exp:divergence:finding_t}}
    \caption{Inferring when to stop in few-shot transfer learning from the unlabelled examples (5) of a single target task. Setting shown: MAML, CNN, Quickdraw to Omniglot, 5-way 1-shot. We first identify the critical layer $l^*$ and critical moment $\hat{m}^*$, where neural activation dynamics of the target inputs diverge the most from those of the source domain (Eq.~\ref{eq:collas:critical_layer_and_moment}). The strongest divergence is detected at layer 4 along $\hat{m}_2$. We then early-stop at $\hat{t}$ (Eq.~\ref{eq:collas:critical_time}), i.e. when the Pearson correlation of $\psi(\mathbf{x}_{target}, t)$ and $\psi(\mathbf{x}_{valid}, t)$, both evaluated at $l^*$ and $\hat{m}^*$, flips from being positive between $t_0$ and $\hat{t}$, to being negative after $\hat{t}$. Here, activation-based early-stopping at $\hat{t}$ achieves much greater target generalization than validation-based early-stopping at $t^*_{valid}$.}
    \label{fig:analysis:activation-based_early-stopping}
\end{figure}

We quantify the divergence $d$ between the source and target trajectories by their negative correlation across time. More precisely, for a given layer $l$ and moment $\hat{m}$, a strong divergence $d$ corresponds to a large time interval $[t_1, t_2]$ within which the source and target trajectories, both evaluated at $l$ and $\hat{m}$, show a strong negative Pearson correlation $\rho$ through time $t$:
\begin{equation}
    d\Big(\psi_{l, \hat{m}}(\mathbf{x}_{target}, t), \psi_{l, \hat{m}}(\mathbf{x}_{valid}, t), t_1, t_2\Big) \overset{\Delta}{=} - \bigg( \rho_{t_1 < t \leq t_2}\Big(\psi_{l, \hat{m}}(\mathbf{x}_{target}, t), \psi_{l, \hat{m}}(\mathbf{x}_{valid}, t)\Big) \bigg) \times (t_2 - t_1) \, .
\label{eq:collas:trajectory_divergence}
\end{equation}
The critical layer $l^*$ and critical moment $\hat{m}^*$ are where we observe the strongest divergence as defined by $d$, with the time interval bound between $t_0$ and $t^*_{valid}$:
\begin{equation}
    l^*, \hat{m}^* = \mathop{\mathrm{argmax}}_{l, \hat{m}} \max_t d\Big(\psi_{l, \hat{m}}(\mathbf{x}_{target}, t), \psi_{l, \hat{m}}(\mathbf{x}_{valid}, t), t, t^*_{valid}\Big) \, .
\label{eq:collas:critical_layer_and_moment}
\end{equation}
Once we have identified $l^*$ and $\hat{m}^*$, our estimated early-stopping time $\hat{t}$ is simply the moment when this strongest observable divergence begins (see Fig.~\ref{fig:analysis:activation-based_early-stopping}):
\begin{equation}
   \hat{t} = \mathop{\mathrm{argmax}}_t d\Big(\psi_{l^*, \hat{m}^*}(\mathbf{x}_{target}, t), \psi_{l^*, \hat{m}^*}(\mathbf{x}_{valid}, t), t, t^*_{valid}\Big)\, .
\label{eq:collas:critical_time}
\end{equation}
Note that if no divergence is detected, the correlation between the target and source trajectories remains positive throughout $t$, thus $d$ is maximized by selecting out stopping time as $\hat{t} = t^*_{valid}$. Hence, our method is designed to perform on par with the validation baseline in such situations.

\section{Experimental results}
\label{sec:performance}

We now present experimental results of the performance of our proposed method, ABE. For each experiment, tasks are 5-way 1-shot classification, and we only use the unlabelled input examples from the support set of a single target task. In other words, only 5 unlabelled images of a single target task are used to analyze the neural activation dynamics. At the beginning of an experiment, we randomly sample a task $\mathcal{T}_i$ from $p(\mathcal{T}_{target})$ and keep only its set of support input examples, which we use for early-stopping, and evaluate the resulting target accuracy (performance). We repeat this for 50 independently and identically distributed support sets from $p(\mathcal{T}_{target})$, and report the average performance. This is then repeated for 3 independent training runs. As the baseline for direct comparison, we use the common early-stopping approach based on performance on the validation set. For each experiment, validation and target accuracies are averaged over 600 tasks. Note that the examples used to measure the neural activations and perform early-stopping are \textit{not used} in the evaluation of generalization. To benchmark our method, we use the standard 4-layer convolutional neural network architecture described by \cite{DBLP:journals/corr/VinyalsBLKW16} trained on 3 different Meta-Learning algorithms: MAML \citep{DBLP:journals/corr/FinnAL17}, Prototypical Network \citep{prototypical}, and Matching Network \citep{DBLP:journals/corr/VinyalsBLKW16}. Our source and target datasets are taken from the Meta-Dataset \citep{Triantafillou2020Meta-Dataset}. We refer to App.~\ref{sec:appendix:exp_details} for full experimental details and App.~\ref{sec:appendix:exp_results} for additional experiments.

Across our experiments, our results show that our method outperforms validation-based early-stopping. On average, our method closes 47.8\% of the generalization gap between the validation baseline and optimal performance. Notably, when the baseline generalization gap is larger (baseline performance is $<$ 0.95 $\times$ optimal accuracy), we close the generalization gap by 69.2\%. Moreover, when this gap is already small (baseline performance $\geq$ 0.95 $\times$  optimal accuracy), we still manage to slightly outperform the validation baseline, closing the gap by 14.5\% on average. More detailed experimental discussions are provided below. 

\begin{figure}[H]
    \centering
    \subfloat[MAML]{
        \includegraphics[width=0.33\linewidth]{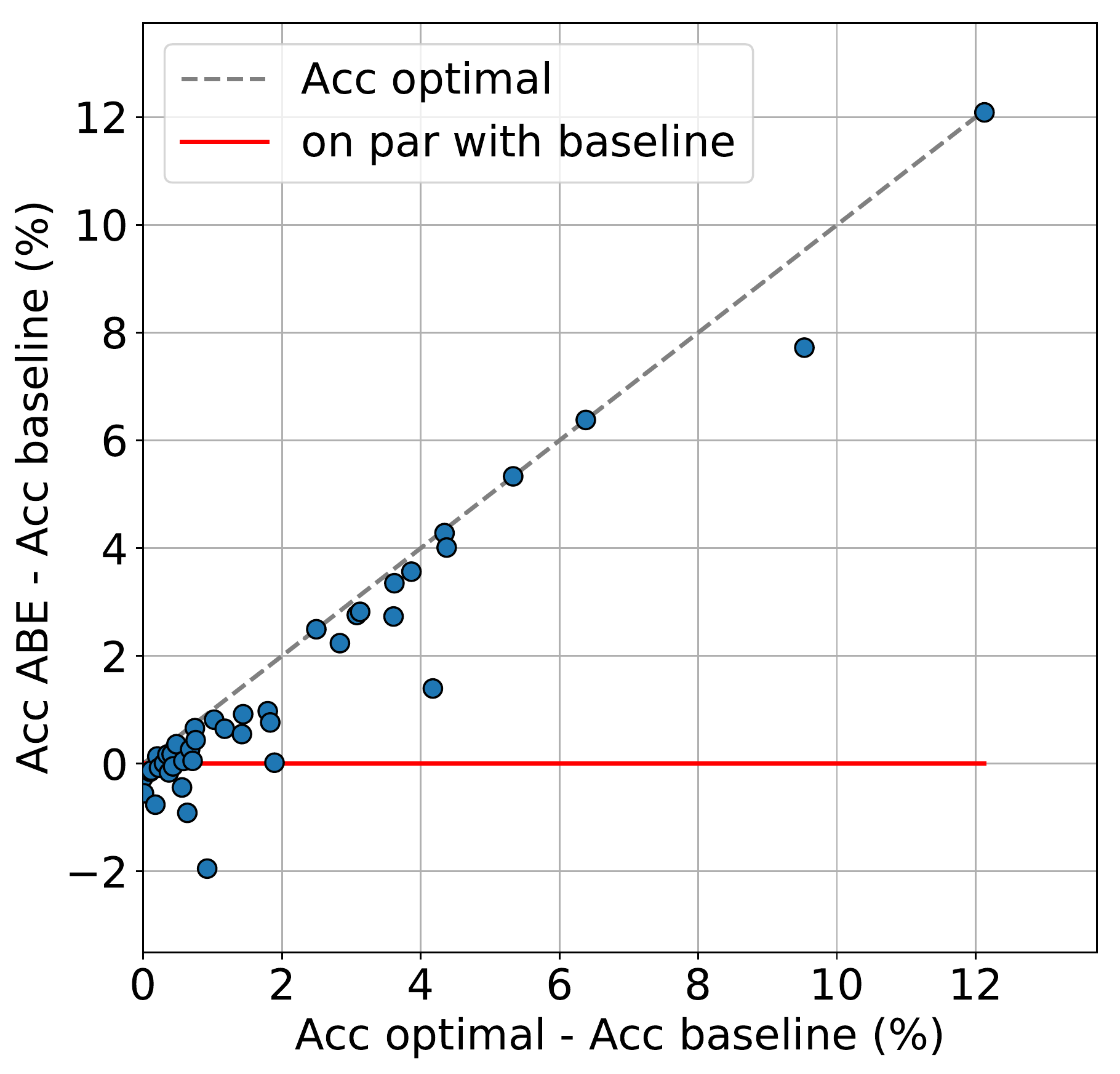}
        \label{fig:scatter:our_method-baseline:maml}
    }
    \subfloat[Prototypical Network]{%
\includegraphics[width=0.33\linewidth]{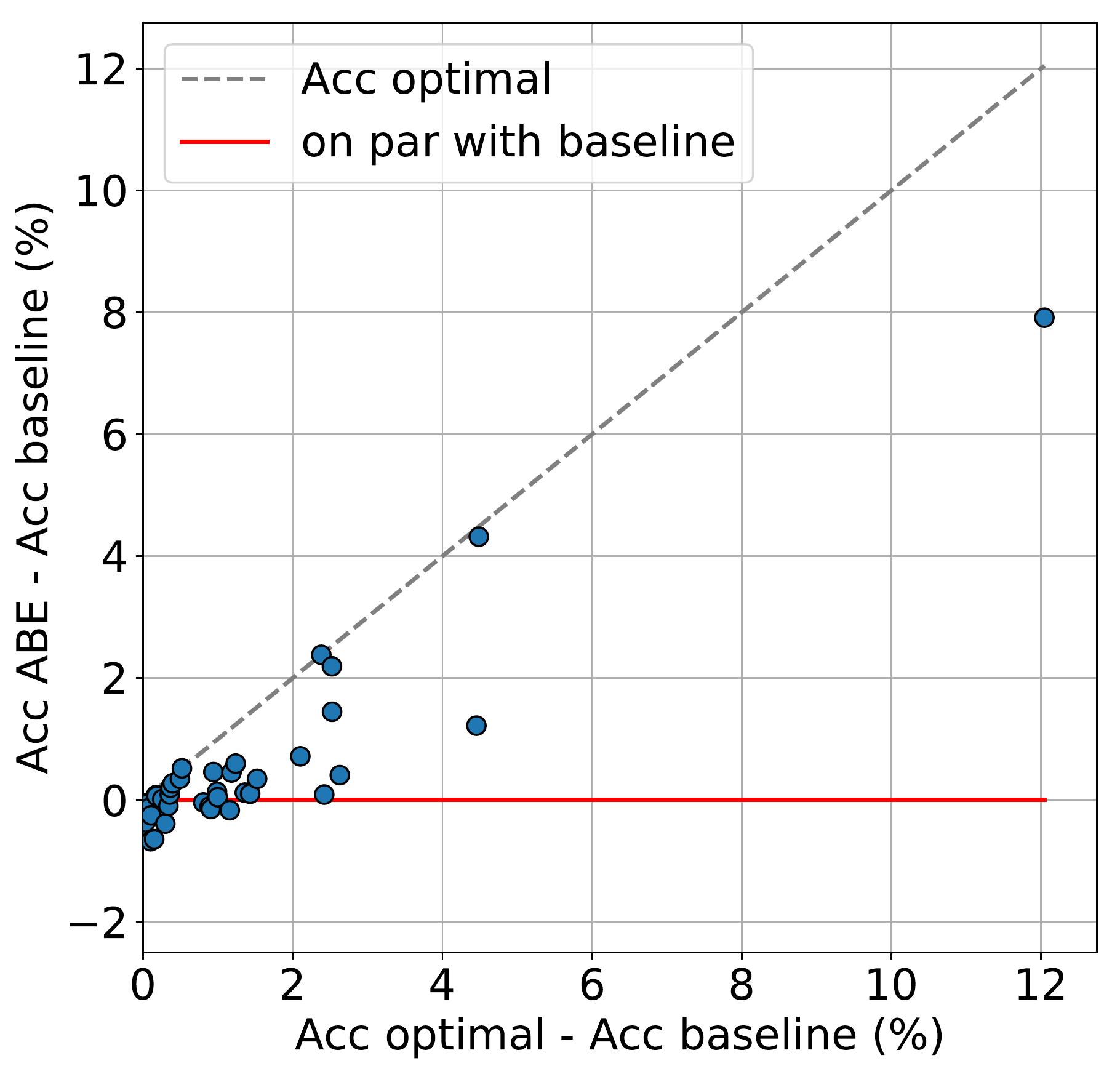}
        \label{fig:scatter:our_method-baseline:prototypical}
    }
    \subfloat[Matching Network]{%
        \includegraphics[width=0.33\linewidth]{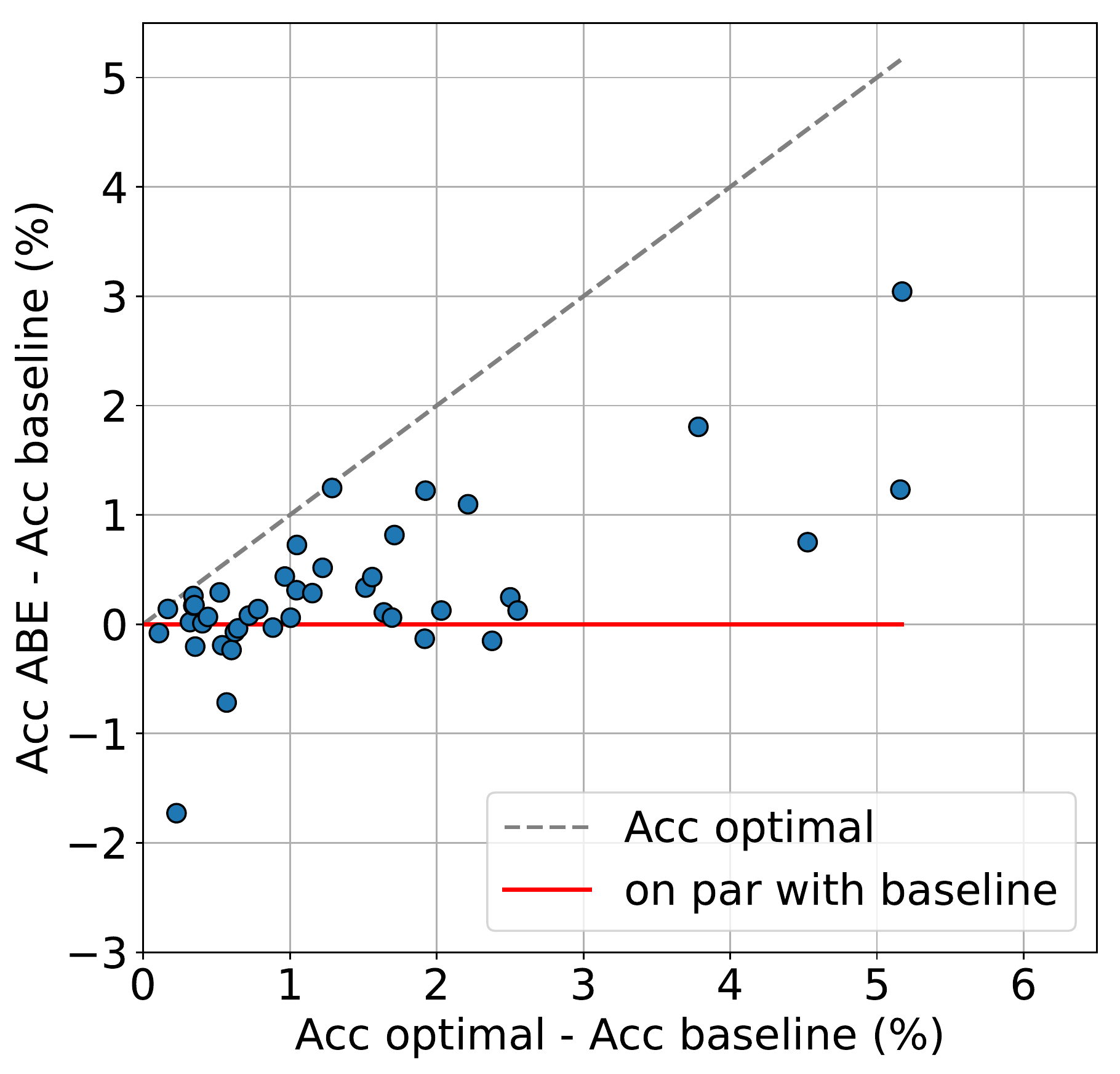}
        \label{fig:scatter:our_method-baseline:matching}
    }
    \caption{Gain in target accuracy from our early-stopping method (ABE), compared to validation-based early-stopping (baseline). The performance gain (y-axis) is shown against the baseline generalization gap (x-axis). For each experiment, we plot the accuracy difference of our method relative to the baseline. The red horizontal bar indicates \textit{on par} with the baseline (no difference). The grey dashed line indicates the maximum achievable performance (completely filling the generalization gap of the baseline). Results show that our method generally outperforms the baseline. }\label{fig:scatter:our_method-baseline}
\end{figure}

In Fig.~\ref{fig:scatter:our_method-baseline} we show the gain in target accuracy from our early-stopping method (ABE), compared to validation-based early-stopping (baseline). We also show the optimal performance, \textit{i.e.} the performance obtained if early-stopping had been optimal in each experiment. For each setting (algorithm, source dataset, and target dataset), we show the accuracy difference against the generalization gap exhibited by the validation-based early-stopping. Results show that our method outperforms the baseline. When the generalization gap of the baseline is small and there is not much gain to have, our method is generally on par with validation-based early-stopping, and only in a very few instances do we suffer from a slight generalization drop. As the baseline generalization gap increases, our method tends to fill this gap. This results in better performance on the target task distribution as observed across all tested Meta-Learning algorithms. In Tab.~\ref{tab:performance} we show the average generalization performance of ABE, compared against validation-based early-stopping, and the optimal performance. Results indicate that ABE outperforms validation-based early-stopping, closing the generalization gap towards optimal performance. We break the results into two categories : performance per source dataset, averaged over all Meta-Learning algorithms and target datasets (Tab.~\ref{tab:performance:source_datasets}); performance per Meta-Learning algorithm (Tab.~\ref{tab:performance:algorithms}).

\renewcommand{\tabcolsep}{3pt}
\begin{table}
    \centering
    \begin{minipage}[c]{0.63\textwidth}
    \subfloat[Performance per source dataset]{
    \resizebox{\textwidth}{!}{%
        \begin{tabular}{|c|cccccc|}
\hline
\multirow{2}{*}{Acc. (\%)} & \multicolumn{6}{c|}{Source dataset} \\ \cline{2-7} 
 & \multicolumn{1}{c|}{cu\_birds} & \multicolumn{1}{c|}{aircraft} & \multicolumn{1}{c|}{omniglot} & \multicolumn{1}{c|}{vgg\_flower} & \multicolumn{1}{c|}{mini-imagenet} & quickdraw \\ \hline
Optimal & \multicolumn{1}{c|}{43.74} & \multicolumn{1}{c|}{40.34} & \multicolumn{1}{c|}{34.27} & \multicolumn{1}{c|}{40.55} & \multicolumn{1}{c|}{47.21} & 38.74 \\ \hline \hline
Baseline & \multicolumn{1}{c|}{41.75} & \multicolumn{1}{c|}{38.71} & \multicolumn{1}{c|}{33.26} & \multicolumn{1}{c|}{39.59} & \multicolumn{1}{c|}{45.71} & 36.31 \\ \hline
\textbf{\begin{tabular}[c]{@{}c@{}}ABE (ours)\end{tabular}} & \multicolumn{1}{c|}{\textbf{43.11}} & \multicolumn{1}{c|}{\textbf{39.27}} & \multicolumn{1}{c|}{\textbf{33.77}} & \multicolumn{1}{c|}{\textbf{40.09}} & \multicolumn{1}{c|}{\textbf{46.20}} & \textbf{37.44} \\ \hline
\end{tabular}}%
        \label{tab:performance:source_datasets}
        }
    \end{minipage}
    \begin{minipage}[c]{0.36\textwidth}
    \subfloat[Performance per Meta-Learning algorithm]{
    \resizebox{\textwidth}{!}{%
\begin{tabular}{|cccc|}
\hline
\multicolumn{1}{|c|}{\multirow{2}{*}{Acc. (\%)}} & \multicolumn{3}{c|}{Meta-learning algorithm} \\ \cline{2-4} 
\multicolumn{1}{|c|}{} & \multicolumn{1}{c|}{MAML} & \multicolumn{1}{c|}{ProtoNet} & Matching Net \\ \hline
\multicolumn{1}{|c|}{Optimal} & \multicolumn{1}{c|}{41.11} & \multicolumn{1}{c|}{38.13} & 43.19 \\ \hline \hline
\multicolumn{1}{|c|}{Baseline} & \multicolumn{1}{c|}{39.01} & \multicolumn{1}{c|}{36.87} & 41.78 \\ \hline
\multicolumn{1}{|c|}{\textbf{\begin{tabular}[c]{@{}c@{}}ABE (ours)\end{tabular}}} & \multicolumn{1}{c|}{\textbf{40.51}} & \multicolumn{1}{c|}{\textbf{37.35}} & \textbf{42.09} \\ \hline
\end{tabular}%
}\label{tab:performance:algorithms}
}
    \end{minipage}
    \caption{Generalization performance of ABE, our proposed early-stopping method, compared to validation-based early-stopping (Baseline). For each setting, ``Optimal" is the maximum achievable generalization performance, if early-stopping had been optimal (with an oracle). We show results per Meta-Learning algorithm (averaged over all source and target datasets), and per source dataset (averaged over all Meta-Learning algorithms and target datasets). The first row shows the maximum achievable target accuracy. Results show that for each source dataset and Meta-Learning algorithm, our method consistently outperforms the validation baseline. To measure the variability in performance of our method, for each experiment (algo, source dataset, target dataset), we compute the standard deviation among the generalization performances obtained with the 50 different target tasks (independent trials). We then average those standard deviations across all experiments. We present them on a per-algorithm basis. Those standard deviations are (in percentages of accuracy): 0.35\% for MAML,  0.13\% for Prototypical Network, and 0.38\% for Matching Network.}
    \label{tab:performance}
\end{table}

\begin{figure}[H]
    \centering
    \includegraphics[width=0.99\linewidth]{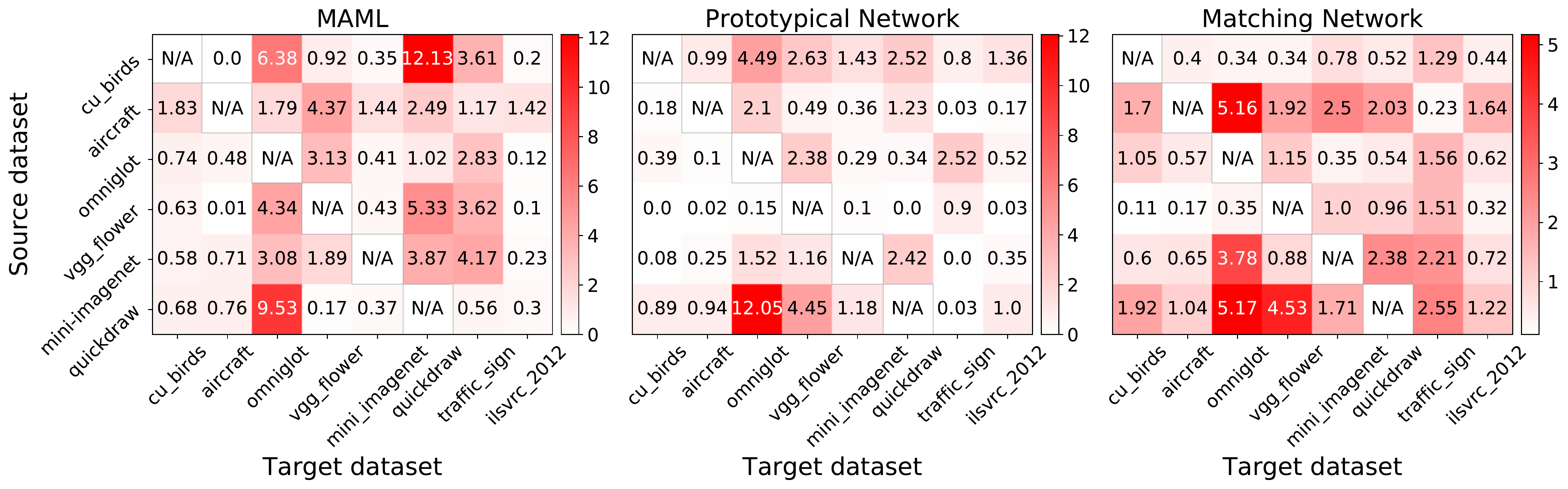}
    \caption{Degradation in target performance (red) using validation-based early-stopping, with respect to the maximum possible accuracy -- we refer to this as the generalization gap. For each experiment, we compute the target accuracy obtained when early-stopping with the validation baseline, and the actual maximum target accuracy that was achievable during the experiment, if early-stopping had been optimal, and we compute the difference.}
    \label{fig:baseline_gap_heatmap:horizontal}
\end{figure}
\begin{figure}[H]
    \centering
    \includegraphics[width=0.99\linewidth]{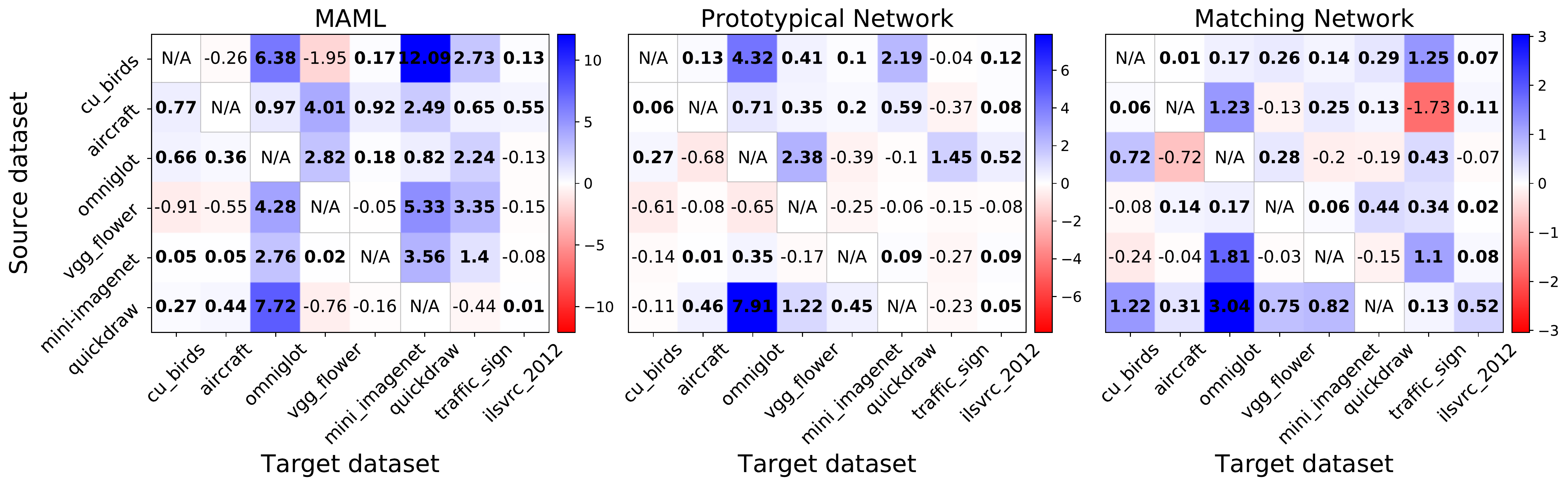}
    \caption{Improvement in generalization performance (blue) of ABE (our method) compared to the validation baseline, versus degradation (red). Difference in generalization performance between ABE and the baseline of validation early-stopping. For each experiment, we compute the target accuracy obtained when early-stopping with the validation baseline, and target accuracy obtained when early-stopping with the validation baseline, and we compute the difference. }
    \label{fig:performance_heatmap:method:horizontal}
\end{figure}

In Fig.~\ref{fig:baseline_gap_heatmap:horizontal} we present the degradation in target performance from using validation-based early-stopping (baseline), across all our settings, namely for each pair of source and target datasets, and across all three Meta-Learning algorithms. This degradation, also referred generalization gap, is the difference between the optimal target accuracy (if early-stopping had been optimal) and the target accuracy obtained from validation-based early-stopping, \textit{i.e.} : $Acc_{optimal} - Acc_{baseline}$.

In Fig.~\ref{fig:performance_heatmap:method:horizontal}, we show the generalization gains of ABE, and how it is able to close this generalization gap, where blue indicates an improvement over the baseline, and red indicates a deterioration. We observe that, overall, our method improves over the baseline, with very few instances of notable deterioration of generalization. Second, by comparing with Fig.~\ref{fig:baseline_gap_heatmap:horizontal}, we see that our method offers gains (blue) where the baseline suffers more severely (red). When the baseline performance is close to the optimum (white and light color) and there isn't much to gain, we perform on par with the baseline (white and light color), as desired. A comparison of the numerical values reveals that we often fill a large portion of the baseline generalization gap, sometimes even completely.

\textbf{Other neural architectures} In addition, we tested our ABE on a different neural architecture: a deep residual network (ResNet-18), as used in the original Meta-Dataset paper. We tested it on a subset of our experiments. Our results show our method outperforming validation-based early-stopping, for all three algorithms. See Tab. \ref{tab:resnet} in App. \ref{sec:appendix:exp_results}.

\textbf{Early-stopping using a single OOD task : ABE vs. support loss} Since ABE has access to the unlabelled data from a single target task, we compare its performance to another baseline : tracking the support loss of the single target task, after fine-tuning on its  examples (since we use 1-shot, we can randomly label the examples). This captures how well can the model classifies the support examples of the task. We use the cross-entropy loss instead of the classification accuracy which can easily saturate even after a single step. We performed this analysis using MAML (as it is not applicable with the other two algorithms). Results show that this baseline doesn't not perform as well as ABE, and in fact performs worse than the validation-based early-stopping, as it increases the generalization gap by 54.1\% on average, while with MAML, our method closes the generalization gap by 71.4\% on average. We also observed a much greater variance between the estimated stopping times for this baseline, when using different tasks, and thus a higher standard deviation in performance of 0.74\% (in accuracy, on average) compared to only 0.35\% for ABE used with MAML. See results in Fig. \ref{fig:performance_heatmap:single-task_baseline:horizontal} of Appendix \ref{sec:appendix:exp_results:baseline_single-task_support_loss}, there are mostly degradation of generalization (red color) and very few instances of improvement (blue color), as opposed to the left most subfigure of  Fig. \ref{fig:performance_heatmap:method:horizontal}. This reinforces some of the insights of our work, for instance : To infer target generalization, one might have to " look under the hood " and inspect the activations at lower layers of the feature extractor. Indeed, in Fig.~\ref{fig:hist:critical_layers_and_moments} of Appendix \ref{sec:appendix:exp_results:critical_layer_and_moment}, we present observations on the frequency, across all of our experiments, of which layer and which aggregated moments are critical, \textit{i.e.} where we observed the strongest divergence between the trajectories of the target and source activations. In Fig.~\ref{fig:hist:critical_layers}, we observe that the strongest divergence may happen at all layers, despite happening more frequently in earlier layers. Moreover, Fig.~\ref{fig:hist:critical_moments} suggests that all of the aggregated moments that we introduced, from $\hat{m}_1$ to $\hat{m}_4$, might be critical in driving the divergence between the activation trajectories. This further motivates our use of these four moments to characterize the trajectories of neural activations.

\textbf{Early-stopping from an additional (oracle) OOD task distribution}
What happens if we perform early-stopping using a third distribution, i.e another dataset previously unseen (assuming we have access to it) ? Should its distributional shift with the training distribution result in a better early-stopping time ? Results in Appendix \ref{sec:appendix:exp_results:issue_meta-val} Tab. \ref{tab:ood-validation_oracle} that while this OOD validation early-stopping can outperform the in-distribution counterpart, our method generally outperforms it, the resulting early-stopping time can vary greatly depending on which dataset is used. The performance is generally lower too, compared to ABE, highlighting the importance of considering the specific target problem at hand. 

\textbf{Using more unlabelled examples} We also tested ABE when using more data to compute the moments on the activations. When using more data, the average performance slightly increases, but more notably, the variance in performance is significantly reduced and generalization becomes more consistent. See Appendix \ref{sec:appendix:exp_results:ABE_more_data}.

\section{Related Work}

In recent years, some works have started to analyze theoretical aspects of gradient-based Meta-Learning. Particularly, \cite{DBLP:journals/corr/abs-1902-08438} provided a theoretical upper bound for the regret of MAML in an online Meta-Learning setting, where an agent faces a sequence of tasks.

Moreover, \cite{denevi2019learning} studied Meta-Learning through the perspective of biased regularization, where the model adapts to new tasks by starting from a biased parameter vector, which we refer to as the meta-training solution.
When learning simple tasks such as linear regression and binary classification with stochastic gradient descent, they theoretically prove the advantage of starting from the meta-training solution, under the assumption that similar tasks lead to similar weight parameterization. 
Considering online convex optimization, where the model learns from a stream of tasks, \cite{DBLP:journals/corr/abs-1902-10644} stated that the optimal solution for each task lies in a small subset of the parameter space. Then, under this assumption, they proposed an algorithm using Reptile, a first-order Meta-Learning method, such that the task-averaged regret scales with the diameter of the aforementioned subset \citep{DBLP:journals/corr/abs-1803-02999}.

Bearing a stronger relation to our approach, \cite{DBLP:journals/corr/abs-1907-07287} empirically studied the objective landscapes of gradient-based Meta-Learning, with a focus on few-shot classification. They notably observed that average generalization to new tasks appears correlated with the average inner product between their gradient vectors. In other words, as gradients appear more similar in the inner product, the model will, on average, better generalize to new tasks, after following a step of gradient descent. Similar findings on the relation between local gradient statistics and generalization have also been previously discussed by \cite{mahsereci2017early}. 

More recently, a few works have studied the properties of the feature extractor $\varphi$ in the context of Meta-Learning. Notably, \cite{2019arXiv190909157R} empirically showed that when neural networks adapt to novel tasks in a few-shot setting, the feature extractor network is approximately invariant, while the final linear classifier undergoes significant functional changes. They then performed experiments where $\varphi$ is frozen at meta-test time, while only the classifier $g$ is fine-tuned, without noticing significant drops in generalization performance compared to the regular fine-tuning procedure. Intuitively, these results suggest that the generalization variation over time might be predominantly driven by some evolving but unknown property of the feature extractor. 
\cite{DBLP:journals/corr/abs-2002-06753} observed that generalization in few-shot learning was related to how tightly embeddings from new tasks were clustered around their respective classes. However, \cite{DBLP:journals/corr/abs-1909-02729} observed that the embeddings at the output of the feature extractor of the last layer, $\varphi_L$, were poorly clustered around their classes, despite mentioning that clustering may be important when measuring the logit outputs of the classifier $g$. This is similar to what \cite{2019arXiv190201889F} observed when dealing with new out-of-distribution examples. These findings suggest that if generalization is related to a property of the feature extractor, this property might be class agnostic. It is also worth mentioning that in earlier attempts to perform transfer learning, intermediate layers of $\varphi$ were shown to be critical in the ability of the model to transfer knowledge \citep{DBLP:journals/corr/YosinskiCBL14}.

Finally, while other alternatives to validation-based early-stopping have been recently proposed \cite{zhang2021optimization,bonet2021channel,yao2022metalearning}, they mostly focus on in-distribution generalization instead of the few-shot transfer-learning settings considered in our work. Without information about the target data, these methods are likely to suffer from the same lack of out-of-distribution generalization of early-stopping based on validation set performance presented throughout our work.

\section{Conclusion and Future Work}
In this work, we have presented empirical evidence that the overfitting point of  meta-learning for deep neural networks for few-shot classification can often be estimated from simple statistics of neural activations and how they evolve throughout meta-training time. Our results suggest that key properties or statistics of how feature extractors respond to the target input distribution can be found which are simple enough to be estimated from just a few \emph{unlabelled} target input examples. However, the specific function of the activations, and the layer at which to measure them, need to be inferred. We demonstrate that these functions and layers of interest can be inferred and used to guide early stopping -- leading to a new, and effective method for early stopping which represents a significant departure from the \emph{de facto} standard practice of using a validation set. In few-shot learning, these ingredients can be inferred from how the neural activation dynamics of the validation data relate to the validation accuracy. More importantly, in few-shot transfer learning where the validation performance is not a good proxy for target generalization, they are inferred through searching for which function (in a given function space) and at which layer, that the activation dynamics of the target input domain ``diverge'' the most from those of the source domain. Finally, we have demonstrated have used these insights to propose Activation-Based Early-stopping (ABE) to improve overall generalization to distributions of novel few-shot classification tasks while only using unlabelled support examples from a single target task (5 images). We empirically demonstrated, through a large array of meta-learning experiments, the advantage of our method over the baseline of validation-based early-stopping.

While we proposed our method for estimating target accuracy (and its related early-stopping criterion) in the framework on few-shot learning and meta-learning, our approach is in principle applicable to the broader range of out-of-distribution generalization setups. For this reason, we propose to study the applicability of our method to such setups. We also want to emphasize that the core proposition of our method is to estimate the target generalization from simple statistics (small order moments), by inspecting all layers of the feature extractor, and here we presented a successful implementation of this core idea for meta-learning in few-shot classification, but the exact implementation may have to change if applied in a drastically different scenario. For example, it is possible that when predicting the generalization of a new task distribution but in a continual learning framework, some elements of our implementation might have to change (ex: how we identify the critical layer). We do encourage other authors to consider using our approach, and adapting it if needed, to try to improve generalization in their own setup. 

\bibliography{main}
\bibliographystyle{collas2022_conference}

\newpage

\appendix

\section{Experimental details}\label{sec:appendix:exp_details}

\textbf{CNN}: We use the architecture proposed by \cite{DBLP:journals/corr/VinyalsBLKW16} and used by \cite{DBLP:journals/corr/FinnAL17}, consisting of 4 modules stacked on each other, each being composed of 64 filters of of 3 $\times$ 3 convolution, followed by a batch normalization layer, a ReLU activation layer, and a 2 $\times$ 2 max-pooling layer. With Omniglot, strided convolution is used instead of max-pooling, and images are downsampled to 28 $\times$ 28. With MiniImagenet, we used fewer filters to reduce overfitting but used 48 while MAML used 32. As a loss function to minimize, we use cross-entropy between the predicted classes and the target classes.

\textbf{ResNet-18}: We use the same implementation of the Residual Network as in \cite{Triantafillou2020Meta-Dataset}.

For most of the hyperparameters, we follow the setup of \cite{Triantafillou2020Meta-Dataset}, but we set the main few-shot learning hyperparameters so as to follow the original MAML setting more closely, and in each setting, we consider a single target dataset at a time, with a fixed number of shots and classification ways. We use 5 steps of gradient descent for the task adaptations, 15 shots of query examples to evaluate the test accuracy of tasks. We don't use any learning rate decay during meta-training, and step-size of 0.01 when finetuning the models to new tasks.

\textbf{Datasets}: We use the MiniImagenet and Omniglot datasets, as well as the many datasets included in the Meta-Dataset benchmark \citep{Triantafillou2020Meta-Dataset}.

\section{Additional experimental results}\label{sec:appendix:exp_results}

\subsection{The relation between the neural activation dynamics and generalization to novel tasks}\label{sec:appendix:exp_results:neural_activation_dynamics}

\subsubsection{Relation between the representation space of the feature extractor and target generalization}\label{sec:appendix:exp_results:neural_activation_dynamics:obs_1}

\begin{figure}[H]
\centering
    \subfloat[5-way 1-shot 5-step]{%
        \includegraphics[width=0.33\linewidth]{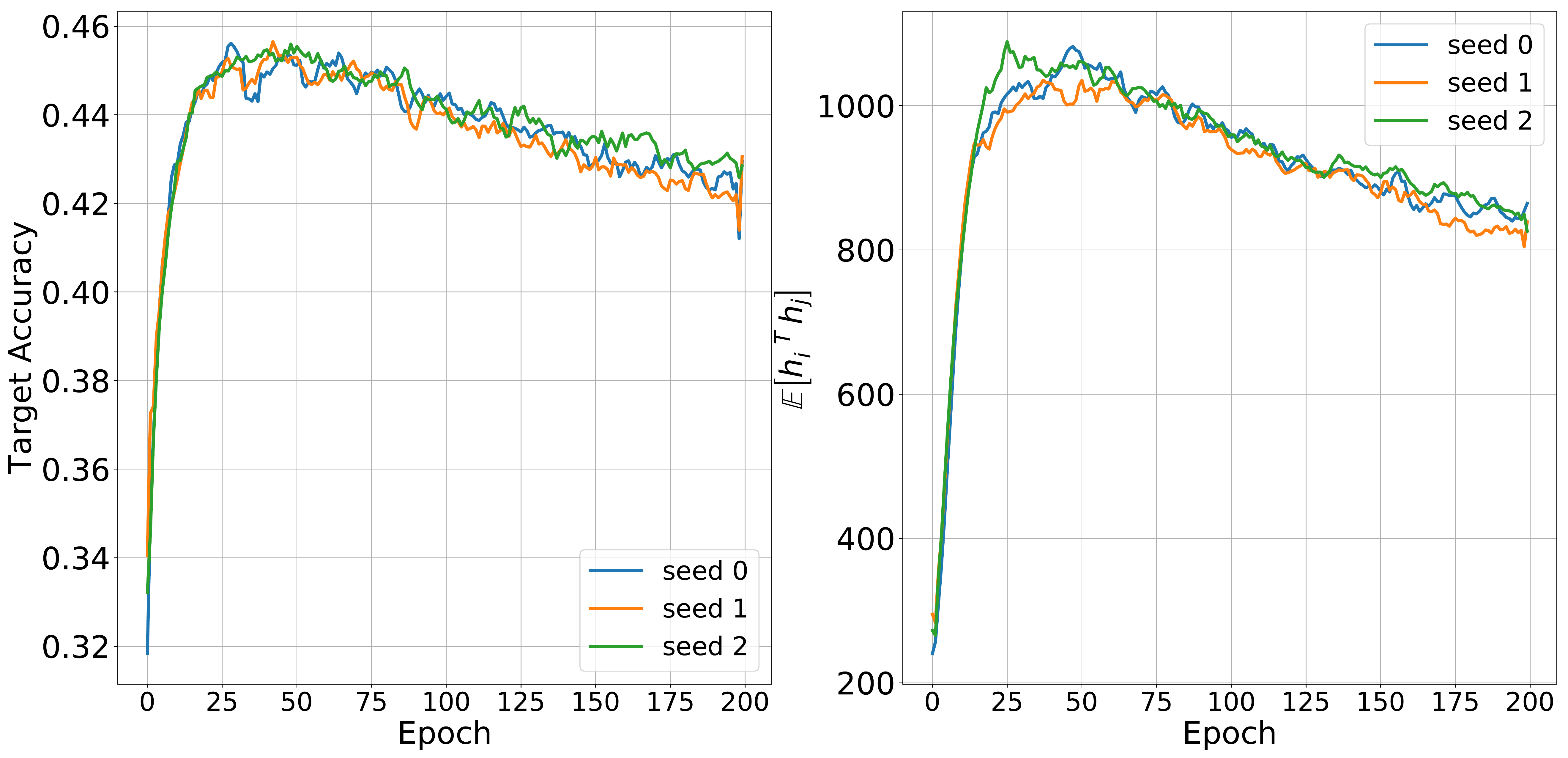}
        \label{fig:model_selection:mini_5w1s5s_fo}
    }
    \subfloat[5-way, 1-shot, 1-step]{%
        \includegraphics[width=0.33\linewidth]{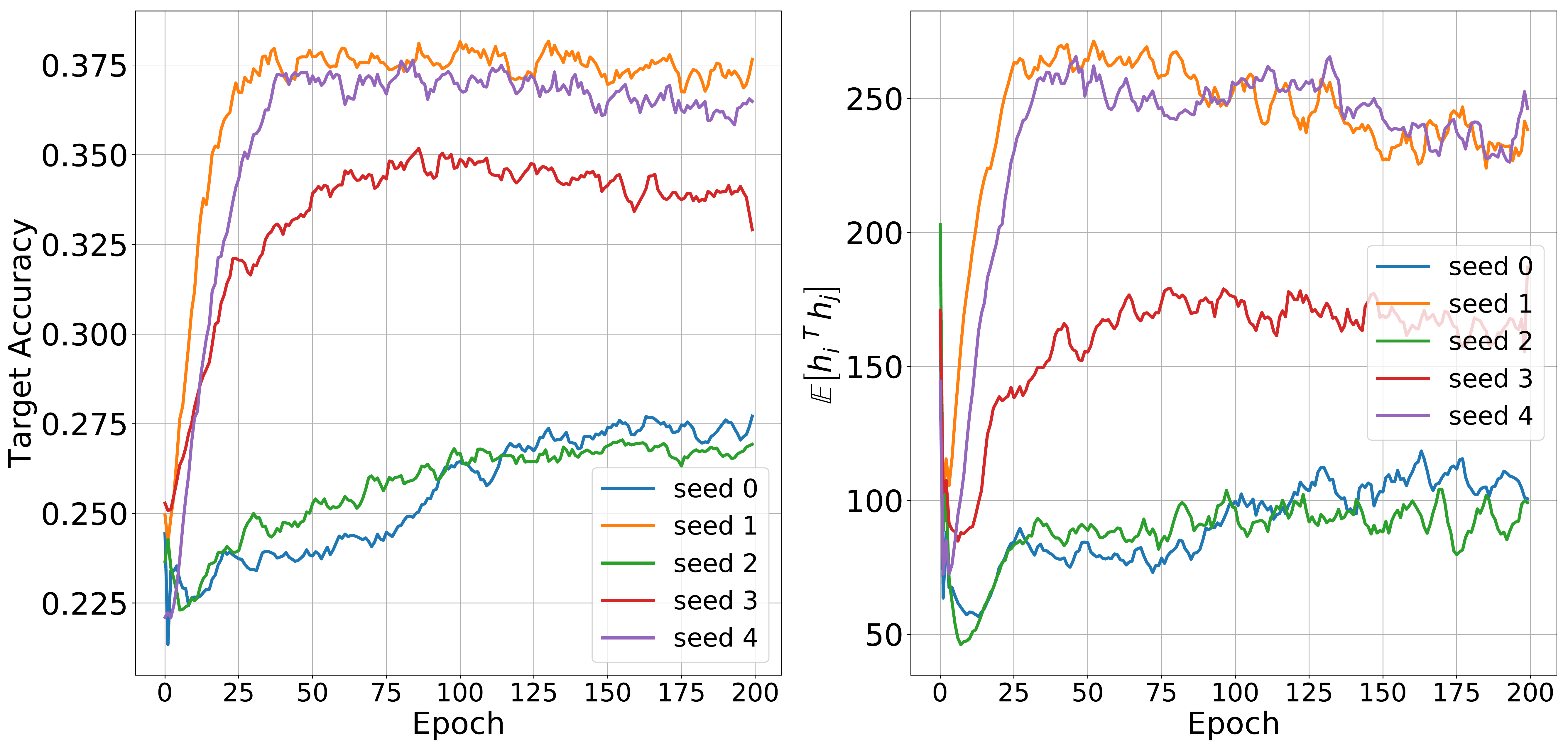}
        \label{fig:model_selection:mini_5w1s1s_fo}
    }
    \subfloat[5-way, 5-shot, 1-step]{%
        \includegraphics[width=0.33\linewidth]{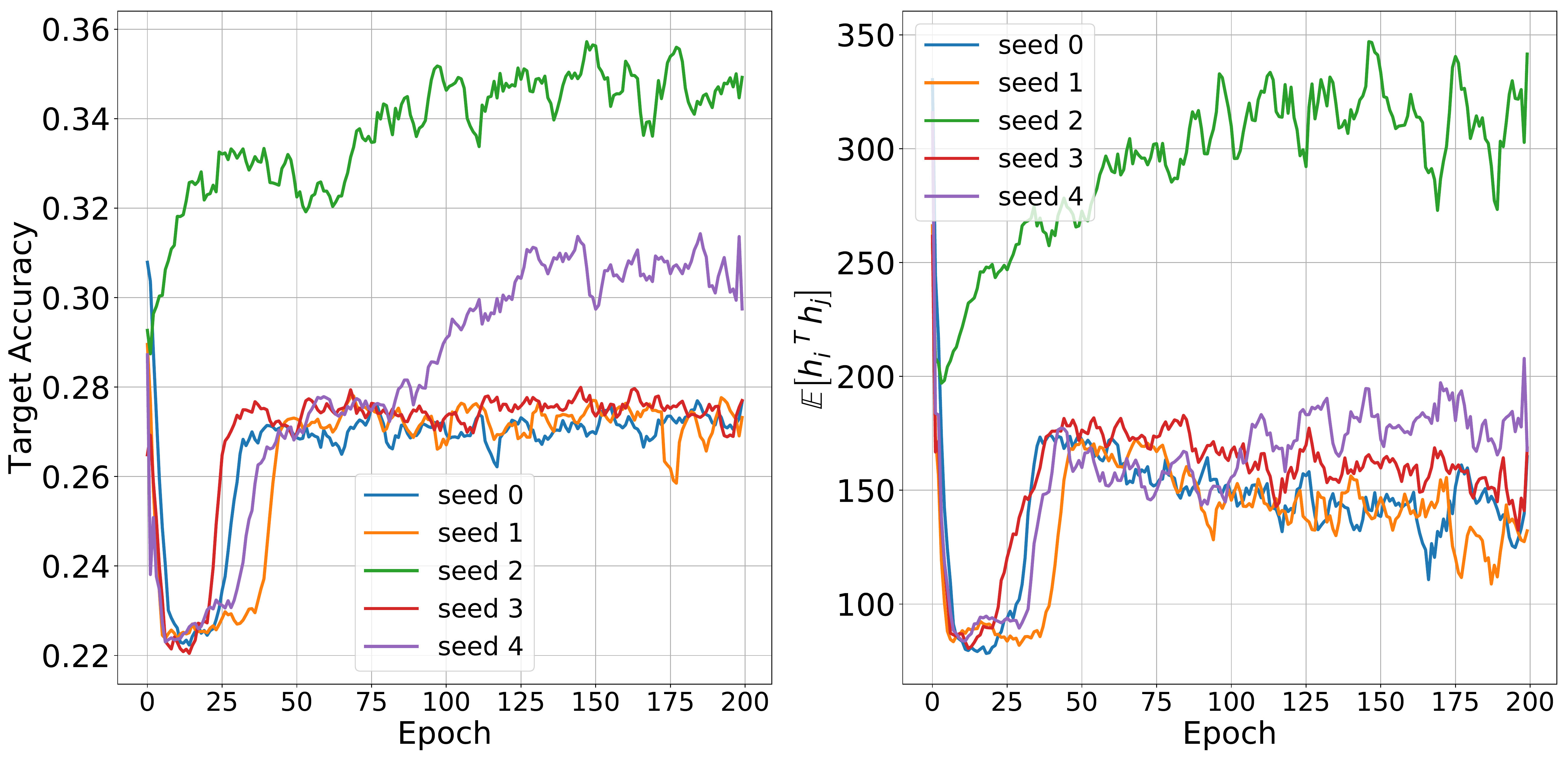}
        \label{fig:model_selection:mini_5w5s1s_fo}
    }
\caption{Comparison between average inner product between representation vectors at the final layer $L$ of the feature extractor, and average target accuracy, in few-shot learning. The expected inner product is a linear combination of the four aggregated moments defined in Sec.~\ref{sec:neural_activation_dynamics}. We use different regimes of MAML and First-Order MAML on MiniImagenet. Here the expression $\mathbb[\mathbf{h}_i^T\mathbf{h}_j]$ is the expected inner product between representations for the target inputs. Note : here $\mathbf{h} \doteq \varphi_L(\mathbf{x})$.}
\label{fig:exp:inner_product:few-shot_learning}
\end{figure}

\begin{figure}[H]
\centering
    \subfloat[Quickdraw to Quickdraw]{%
        \includegraphics[width=0.33\linewidth]{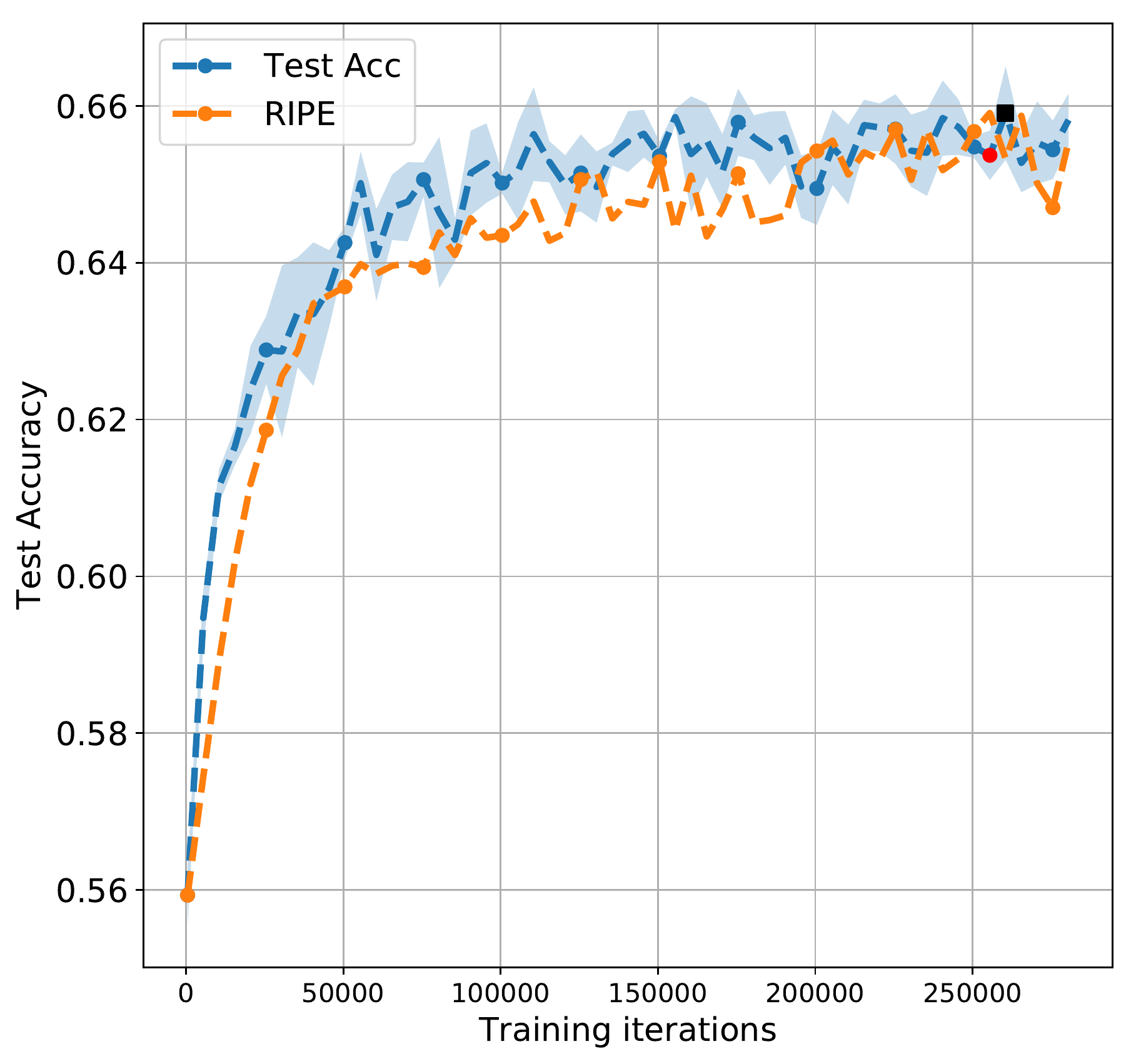}
        \label{fig:erip:quickdraw}
    }
    \subfloat[Quickdraw to Omniglot]{%
        \includegraphics[width=0.33\linewidth]{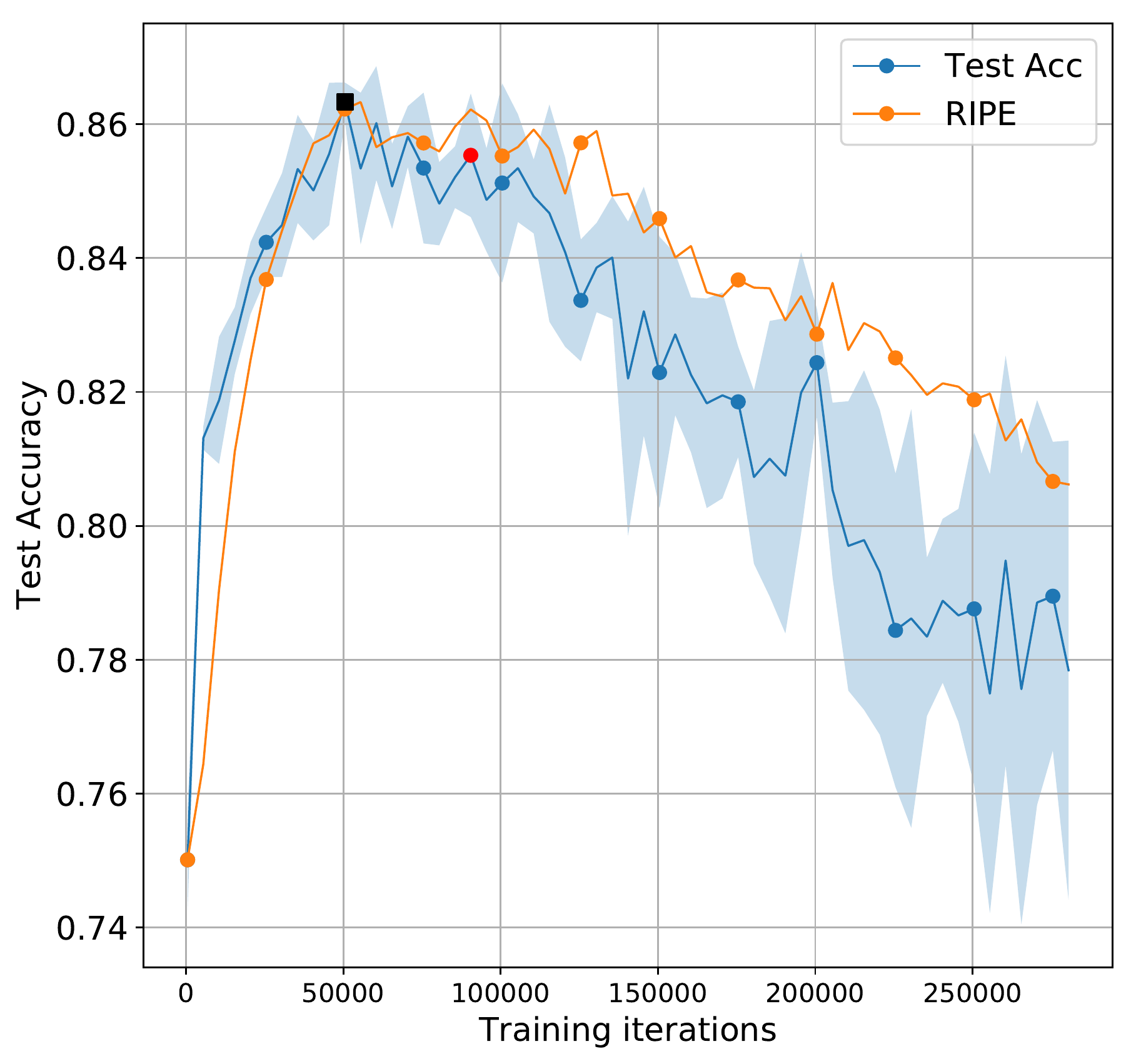}
        \label{fig:erip:miniimagenet}
    }
    \subfloat[Quickdraw to Imagenet]{%
        \includegraphics[width=0.33\linewidth]{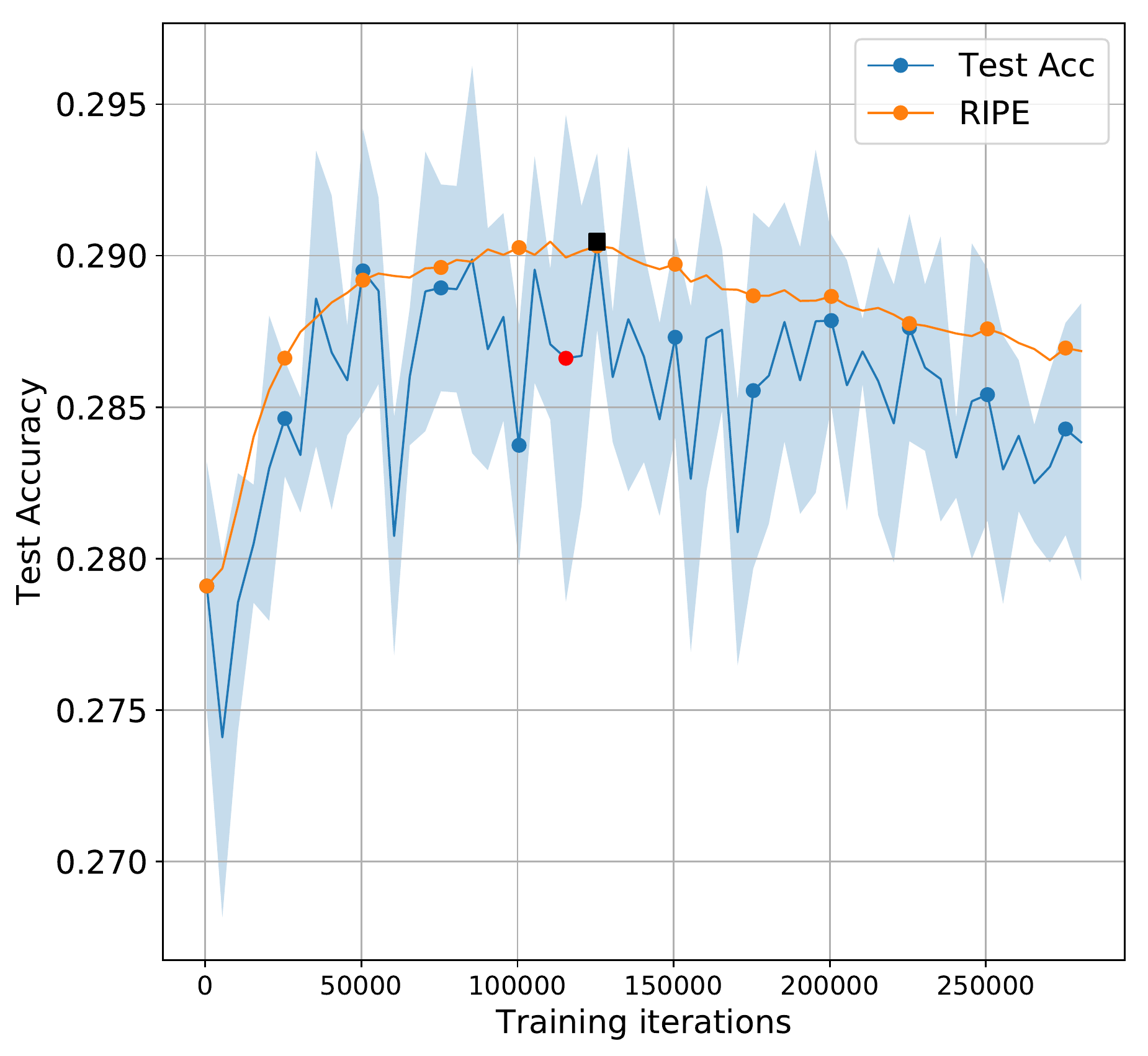}
        \label{fig:erip:aircraft}
    }
\caption{Measuring the expected representational inner product in Few-Shot Transfer Learning. MAML, 5-way 1-shot, training dataset : Quickdraw. The estimated early-stopping time of the metric $\hat{t}$ coincides well with the true optimal early-stopping time $t^*$. Measuring the correlation (Pearson) between  $t^*_{\psi}$ and $t^*$ gives $R=0.925$ with a p-value near 0. Considering the gap between 1) the average performance of validation early-stopping across the three settings (58.69\%); and 2) the maximum generalization across the three settings (61\%), the average performance of the metric is at 59.7\%, closing nearly half of the gap (43.74\% of the gap). Note: RIPE stands for representational inner product, an abbreviation for this specific linear combination of the aggregated moments.}
\label{fig:expected_inner:few-shot_transfer_learning}
\end{figure}

In this section we present experimental results to support the our characterization of the neural activation dynamics introduced in Sec.~\ref{sec:neural_activation_dynamics}. In Fig.~\ref{fig:exp:inner_product:few-shot_learning}, we show that target generalization in few-shot learning is proportional to a linear combination of the four aggregated moments $\hat{m}_1$, $\hat{m}_2$, $\hat{m}_3$ and $\hat{m}_4$, introduced in Sec.~\ref{sec:neural_activation_dynamics}. We then show in Fig.~\ref{fig:expected_inner:few-shot_transfer_learning} a similar observation for the case of few-shot transfer learning, where the target accuracy comes from three different datasets, and which is again a linear combination of the four aggregated moments, when computed on the target inputs.

\subsubsection{Neural activation dynamics: Different levels of the feature extractor can reveal the variation of generalization}\label{sec:appendix:exp_results:neural_activation_dynamics:obs_2}

\begin{figure}[H]
\centering
    \includegraphics[width=0.24\linewidth]{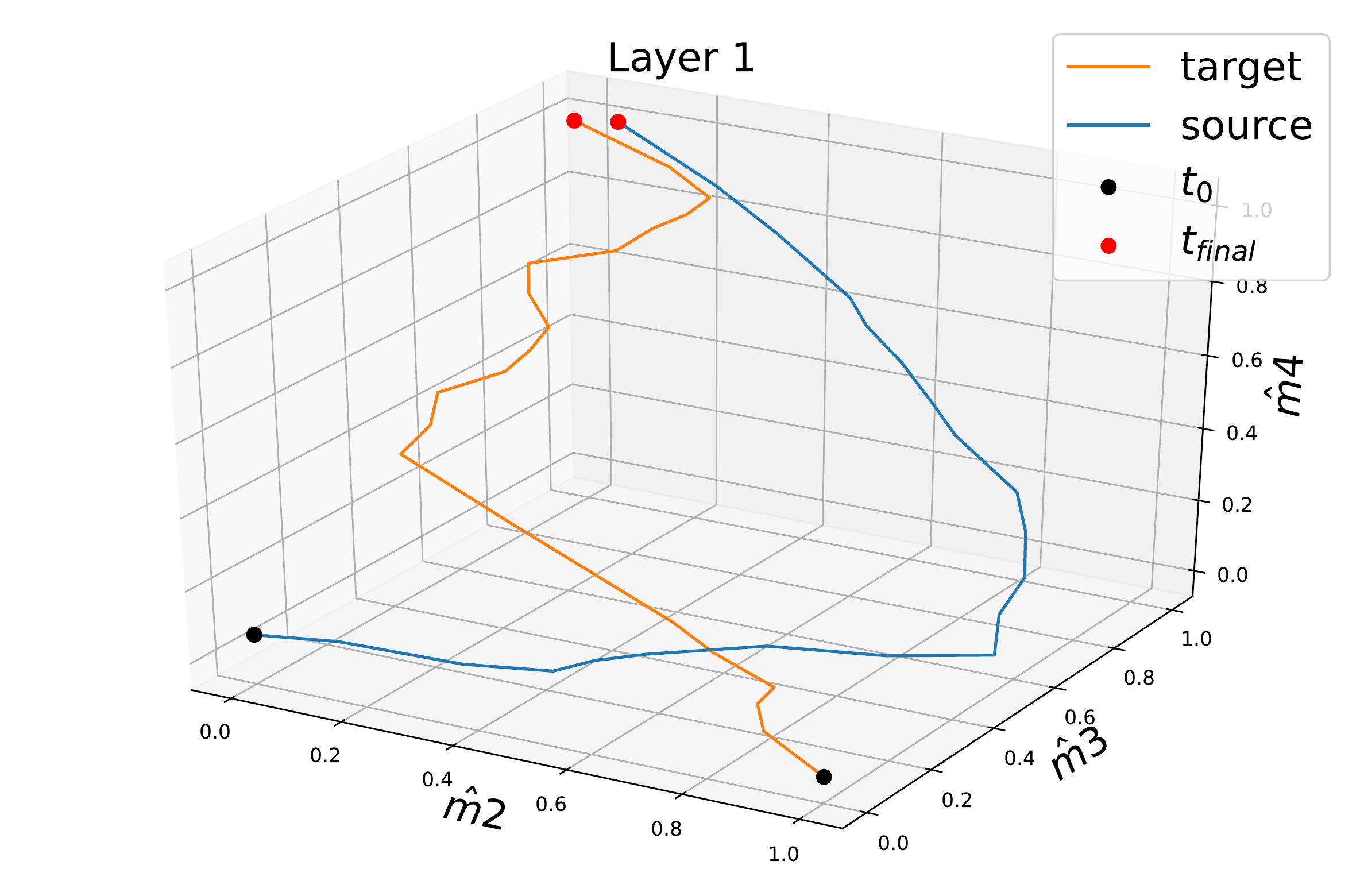}
    \includegraphics[width=0.24\linewidth]{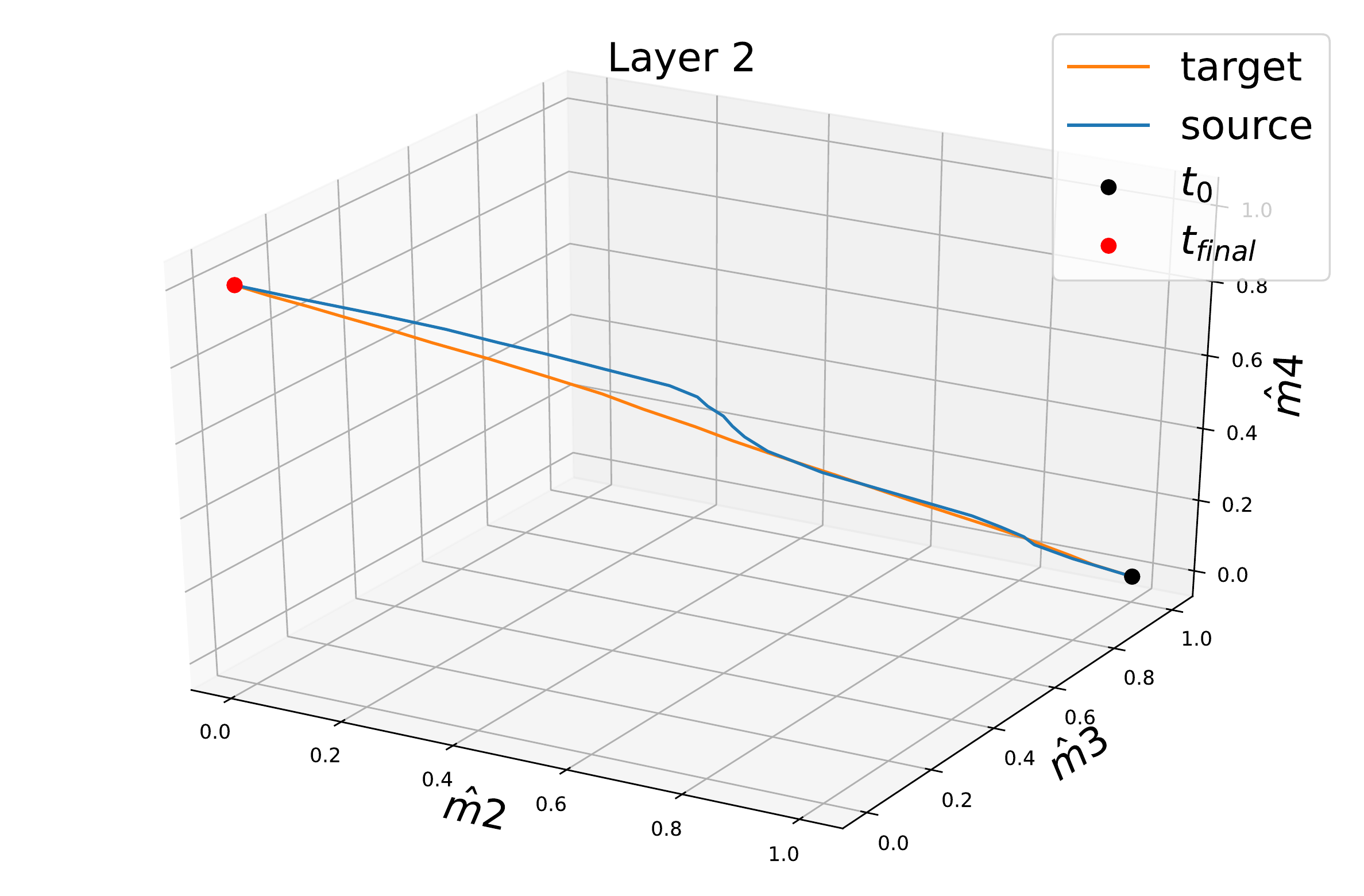}
    \includegraphics[width=0.24\linewidth]{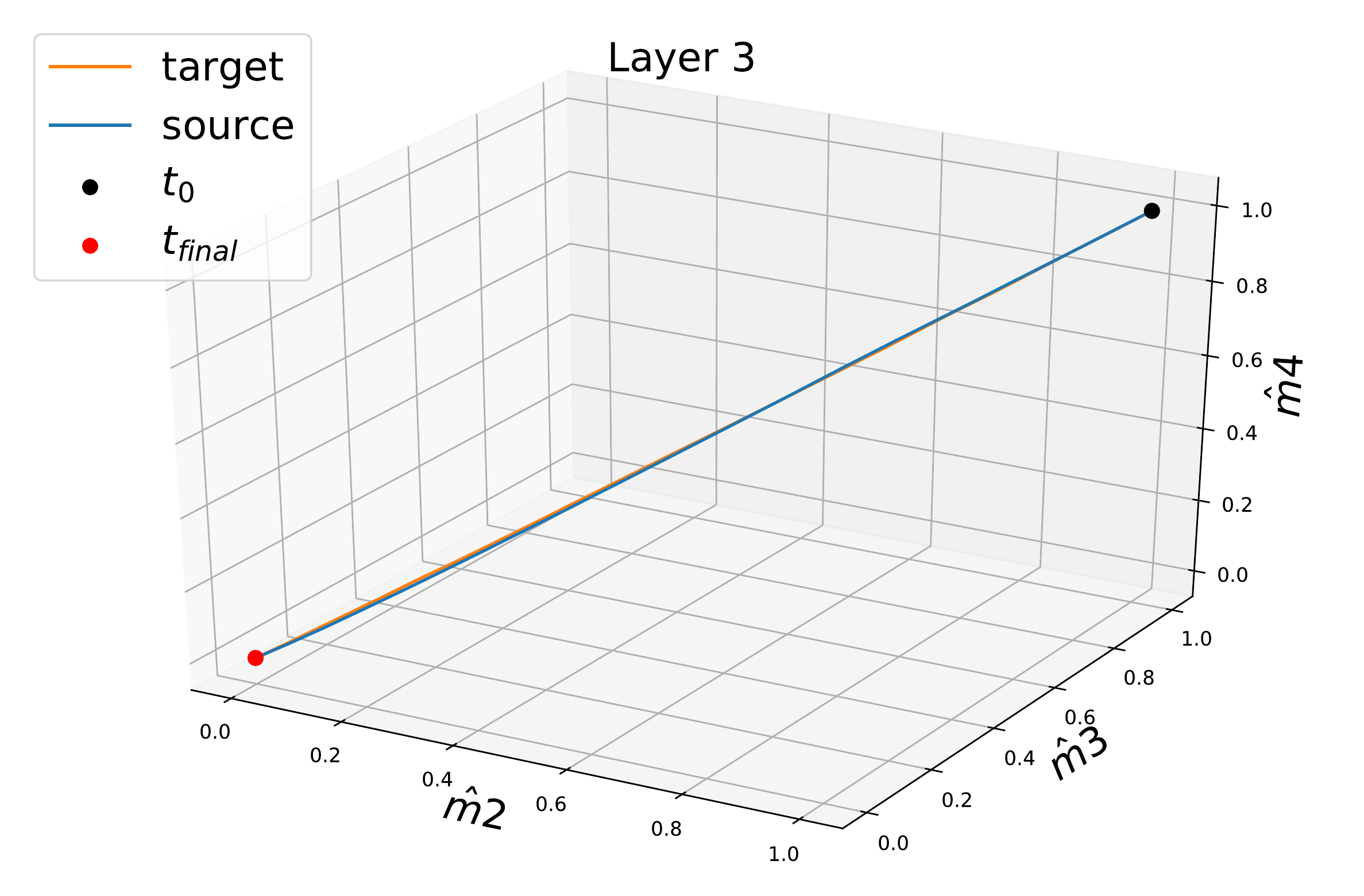}
    \includegraphics[width=0.24\linewidth]{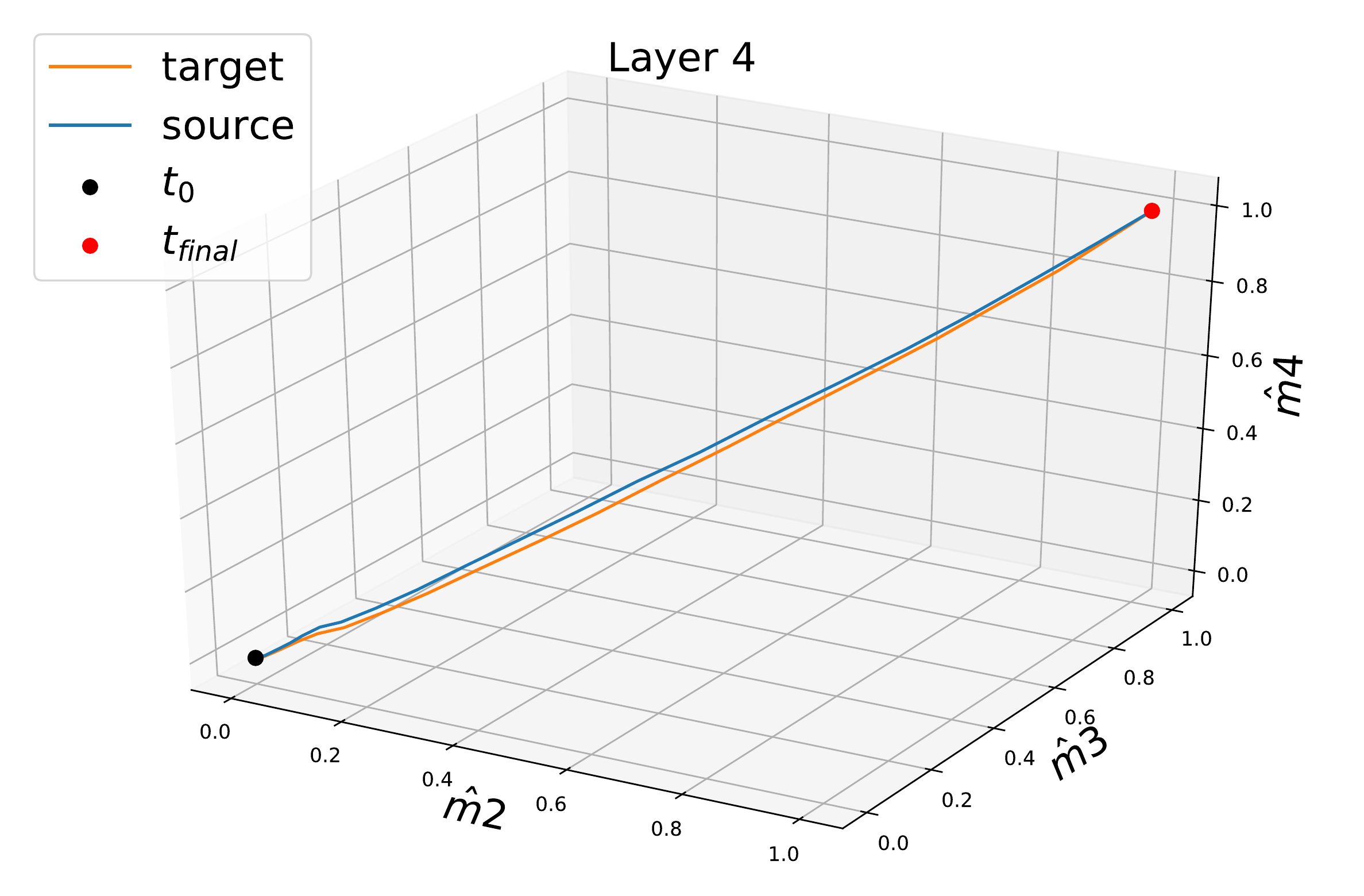}
    \newline
    \includegraphics[width=0.24\linewidth]{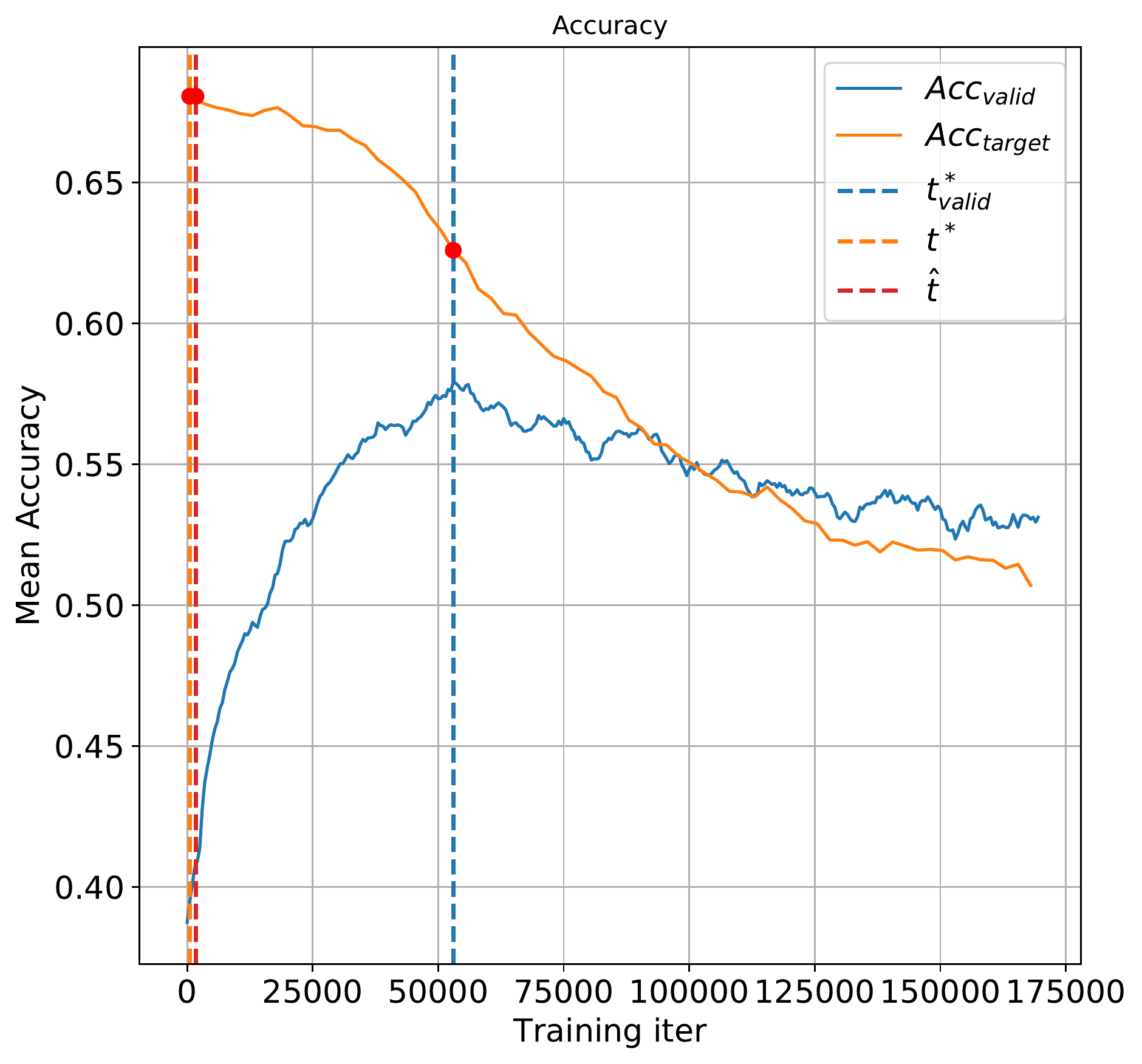}
\caption{MAML : Birds to Omniglot. The divergence between the target and source activation trajectories happens at layer 1, mostly along $\hat{m}_2$, and is detected from the beginning of training. This accurately corresponds to when meta-overfitting occurs. The resulting early-stopping time $\hat{t}$ is close to the optimum $t^*$, which drastically improves generalization, compared to validation-based early-stopping. Note that here we only show three aggregated moments for 3D visualization purposes, each time omitting the dimensions with the least amount of divergence. The trajectory curves have also been normalized here for better visualization, which doesn't affect how ABE works.}
\label{fig:critical-depth:maml-birds-to-omniglot}
\end{figure}

\begin{figure}[H]
\centering
    \includegraphics[width=0.24\linewidth]{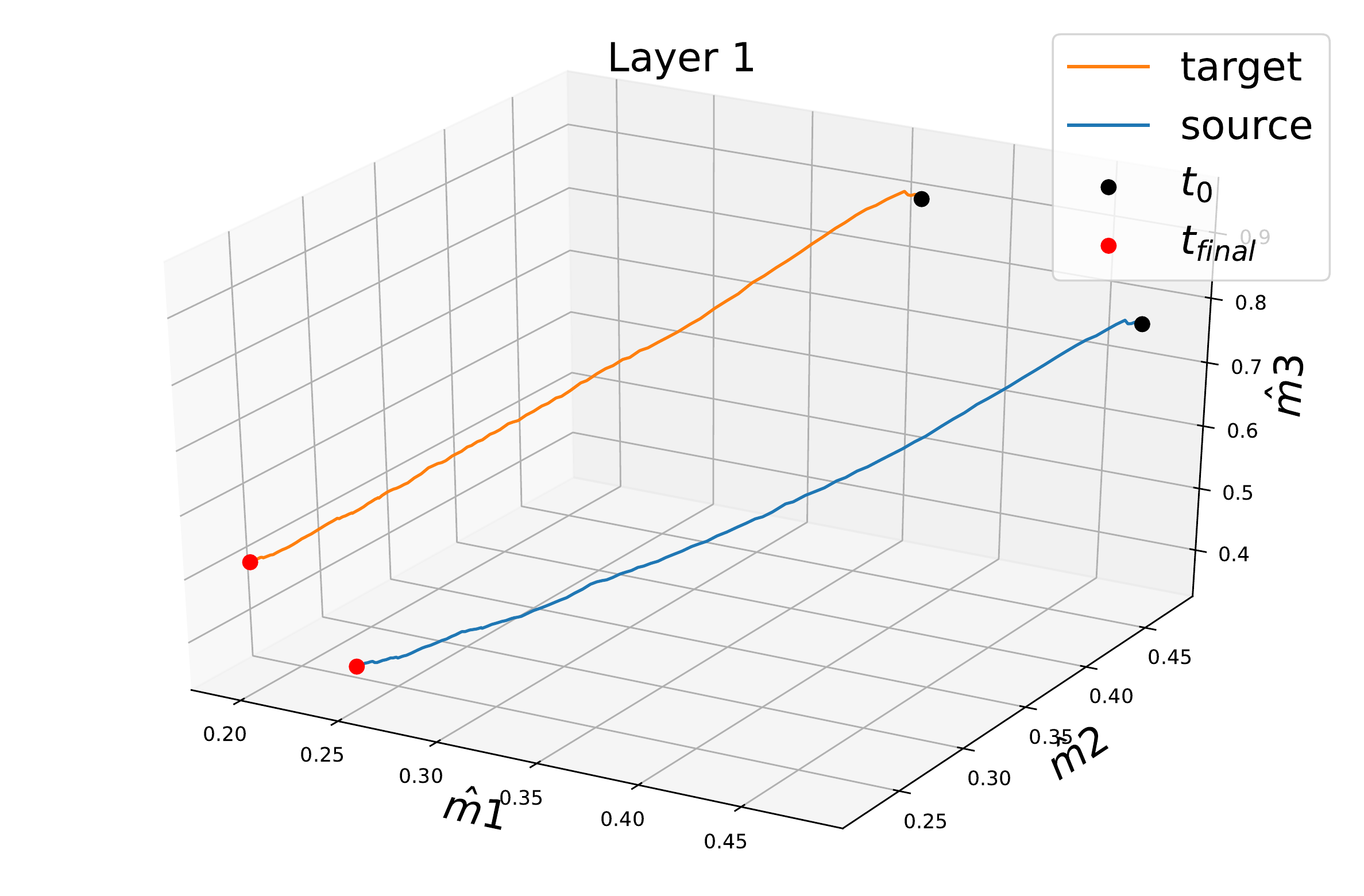}
    \includegraphics[width=0.24\linewidth]{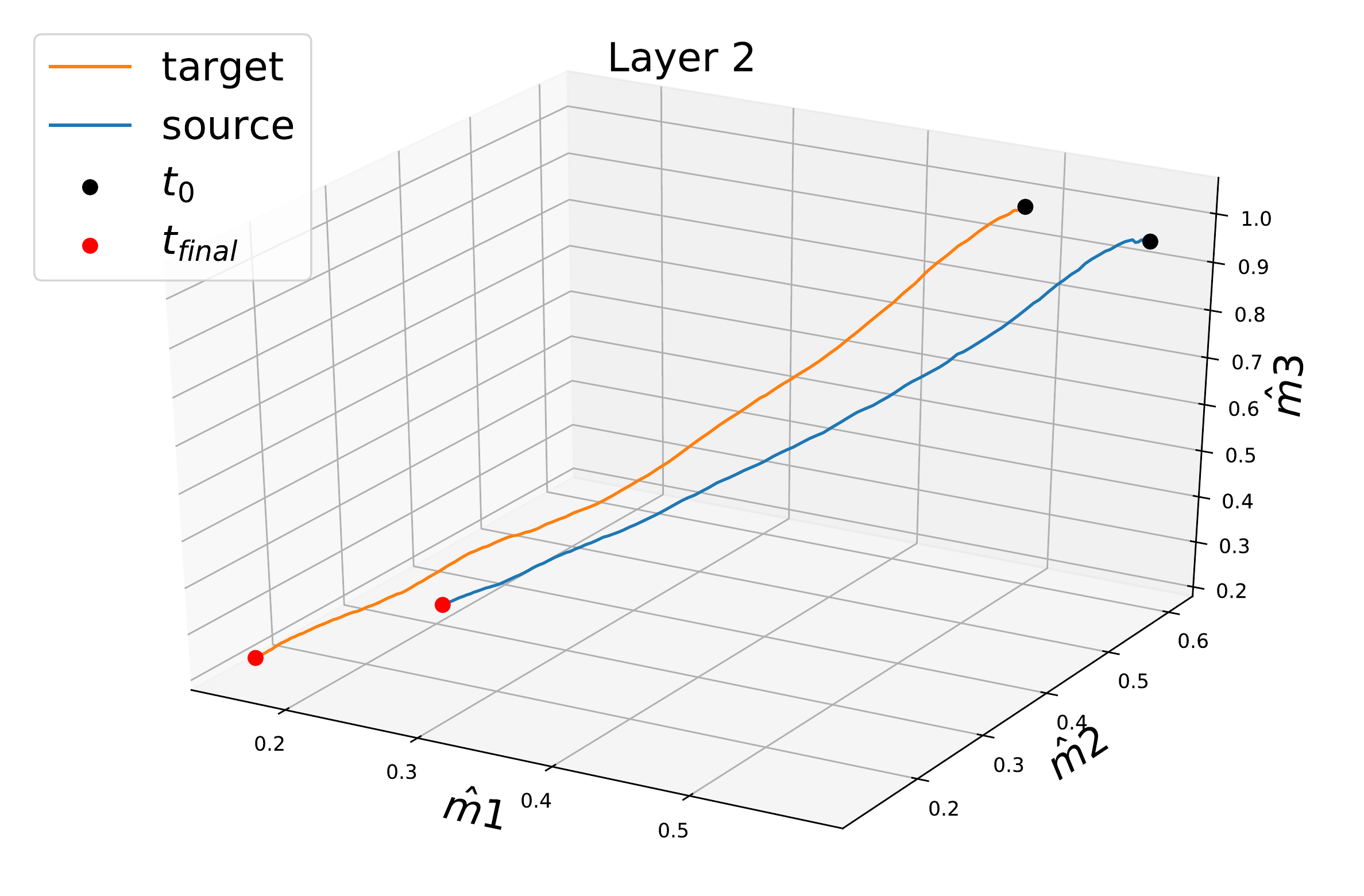}
    \includegraphics[width=0.24\linewidth]{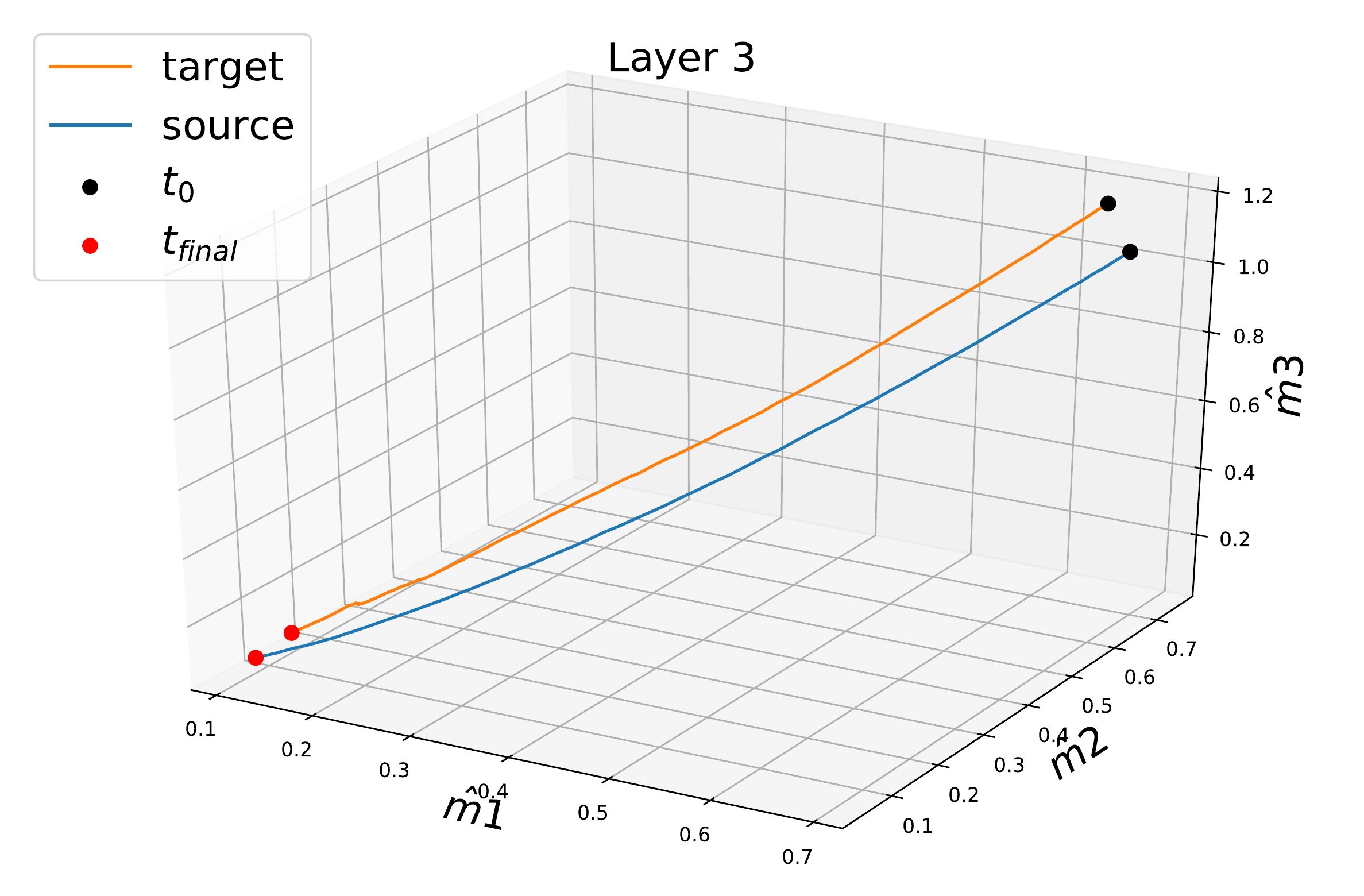}
    \includegraphics[width=0.24\linewidth]{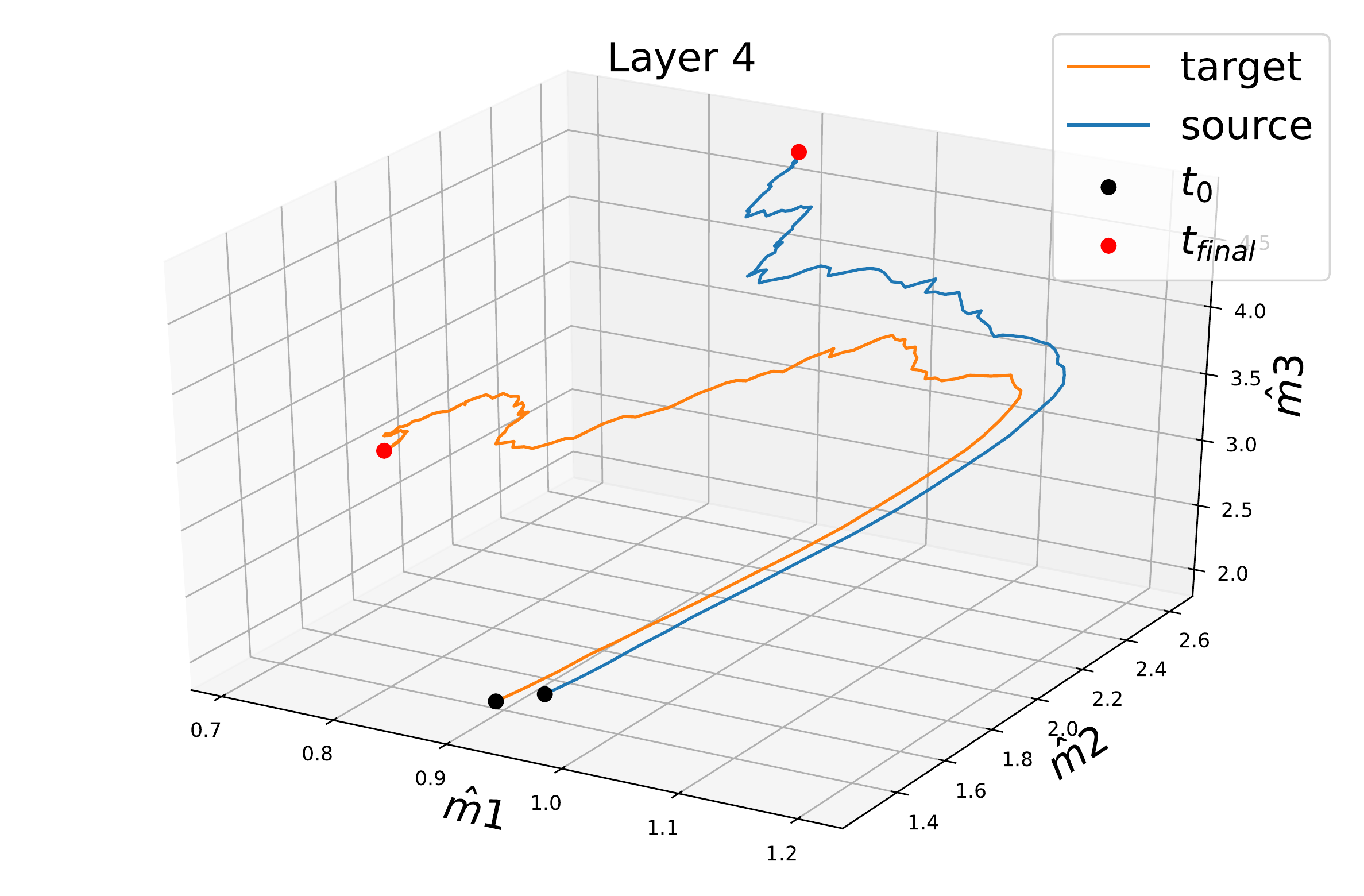}
    \newline
    \includegraphics[width=0.24\linewidth]{figures_paper/trajectories_per_layer/maml_quickdraw-omniglot_performance.pdf}
\caption{MAML : Quickdraw to Omniglot. The divergence between the target and source activation trajectories happens at layer 4, mostly along $\hat{m}_2$. This accurately corresponds to when meta-overfitting occurs. The resulting early-stopping time $\hat{t}$ is close to the optimum $t^*$, which drastically improves generalization, compared to validation-based early-stopping. Note that here we only show three aggregated moments for 3D visualization purposes, each time omitting the dimensions with the least amount of divergence.}
\label{fig:critical-depth:maml-quickdraw-to-omniglot}
\end{figure}

In this section, we present two few-shot transfer learning experiments about the critical layer, i.e. the depth $l$ where the target and source activations have the strongest divergence. There show that even with the same neural architecture, the critical layer may happen at a different level depending on the setting, here the source - target dataset pair. In Fig. \ref{fig:critical-depth:maml-quickdraw-to-omniglot}, divergence between the target and source activation trajectories is the strongest at the last layer of the feature extractor network, while in Fig. \ref{fig:critical-depth:maml-birds-to-omniglot}, this happens at the first layer of the feature extractor. Moreover, in both instances, the activation dynamics at the critical layer correspond the most with target accuracy, compared to the other layers.

\subsubsection{Different functions of the neural activation dynamics can reveal the variation of generalization}\label{sec:appendix:exp_results:neural_activation_dynamics:obs_3}

\begin{figure}[H]
    \centering
    \subfloat[Matching Network, ResNet-18, 5-way 5-shot]{%
        \includegraphics[width=0.49\linewidth]{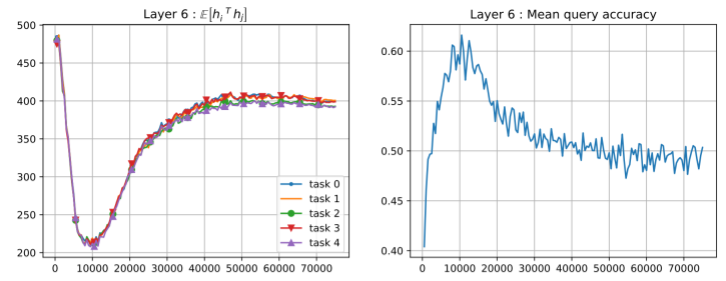}
    }
    \subfloat[Prototypical Network, ResNet-18, 5-way 1-shot]{%
        \includegraphics[width=0.49\linewidth]{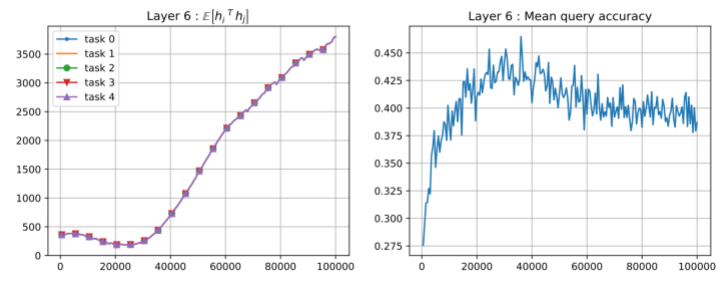}
        \label{fig:model_selection:mini_5w1s5s_fo_pn}
        }
\caption{Strong negative correlation between generalization and metrics on the representation space, where the minimum of the metric coincides with the maximum of generalization: Few-Shot Learning settings with MiniImagenet, 5-way 1-shot, with a ResNet-18. The metric is the expected inner product (left subfigures, represented by $\mathbb{E}[\mathbf{h}_i \; ^T \mathbf{h}_j]$ where $\mathbf{h}_i$ stands for the representation vector for an input example $\mathbf{x}_i$). The metric is measured at the output of the feature extractor (6th block of the ResNet), and we show its measurement on 5 distinct tasks. The generalization (right subfigures) is averaged over 50 tasks.}
\label{fig:inverted_correlation}
\end{figure}

In this section we present further experiments on the correspondence between target accuracy and a linear combination of the aggregated moments, and that this combination may be different depending on the meta-learning setting involved. In Fig.~\ref{fig:inverted_correlation} we present experiments where target again correlates with the expected inner product between activations, with its negative counterpart. We also show this phenomenon with a different neural architecture, a deep Residual Network (ResNet-18), as used in \citep{Triantafillou2020Meta-Dataset}.

\begin{figure}[H]
    \centering
    \subfloat[Exclusive correlation between a specific metric and generalization]{%
        \includegraphics[width=0.25\linewidth]{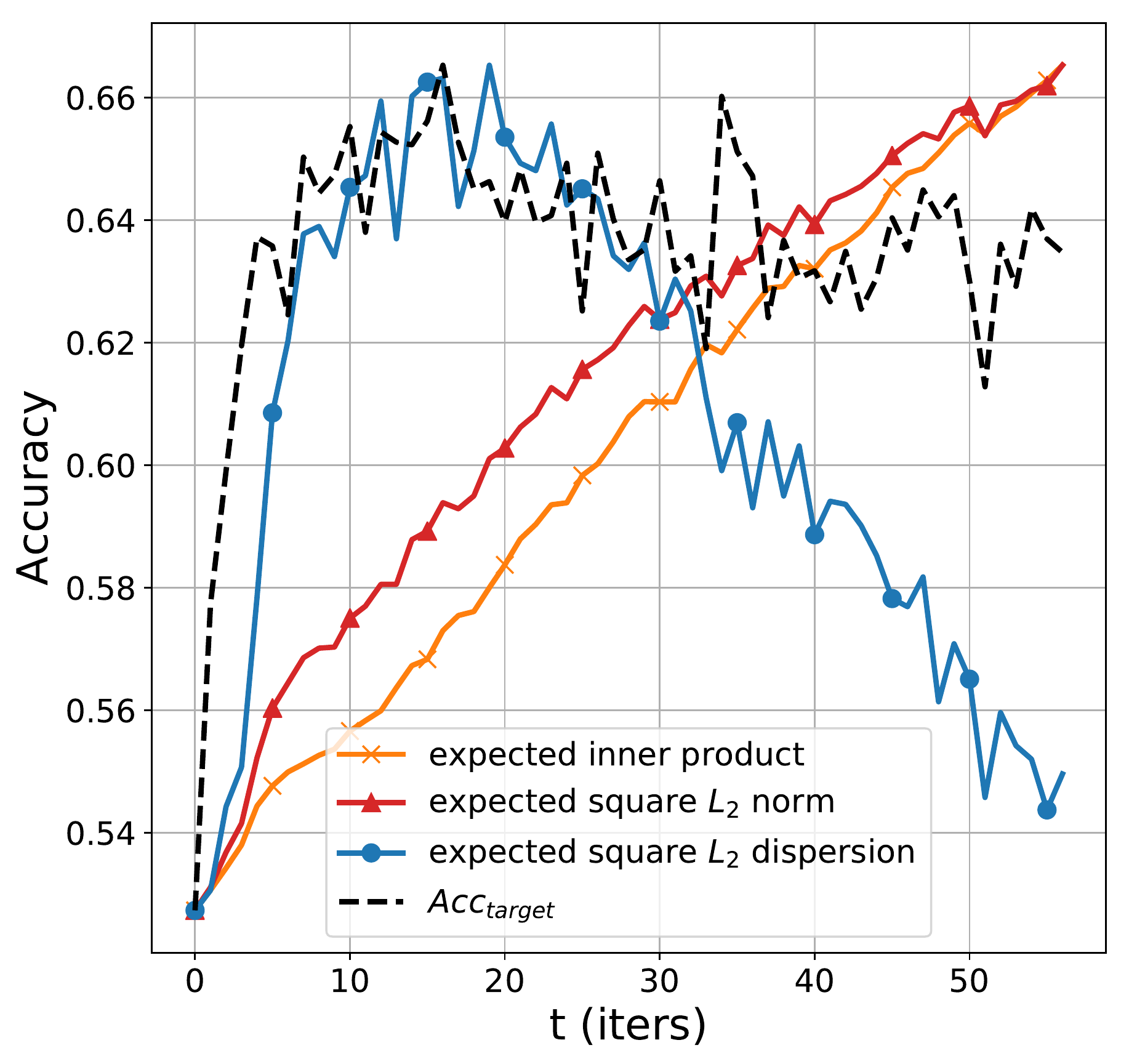}
        \label{fig:various_metrics:a}
    }
    \subfloat[Expected $l_2$ Norm]{%
        \includegraphics[width=0.25\linewidth]{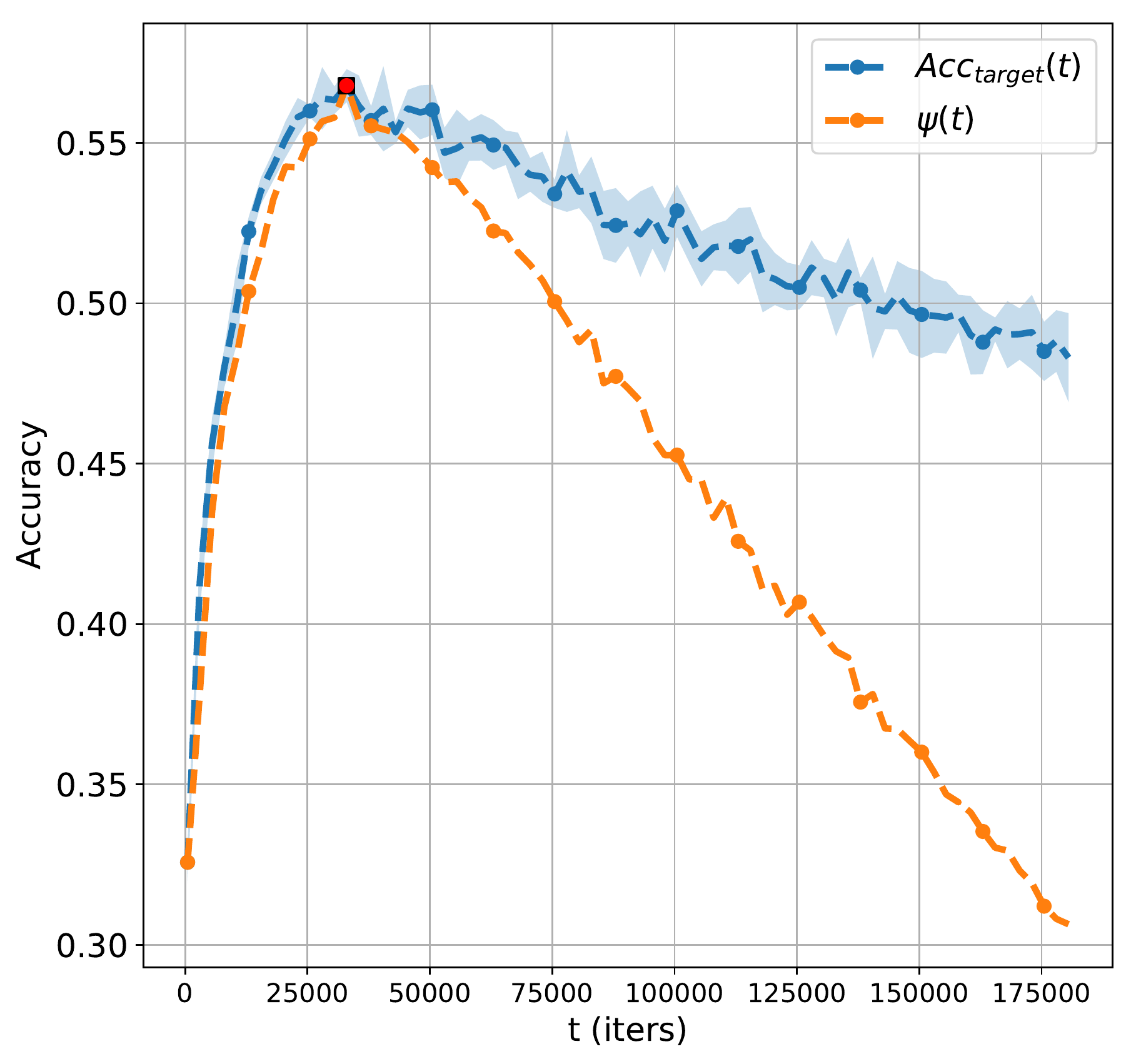}
    }
    \subfloat[Expected $l_2$ Dispersion]{%
        \includegraphics[width=0.25\linewidth]{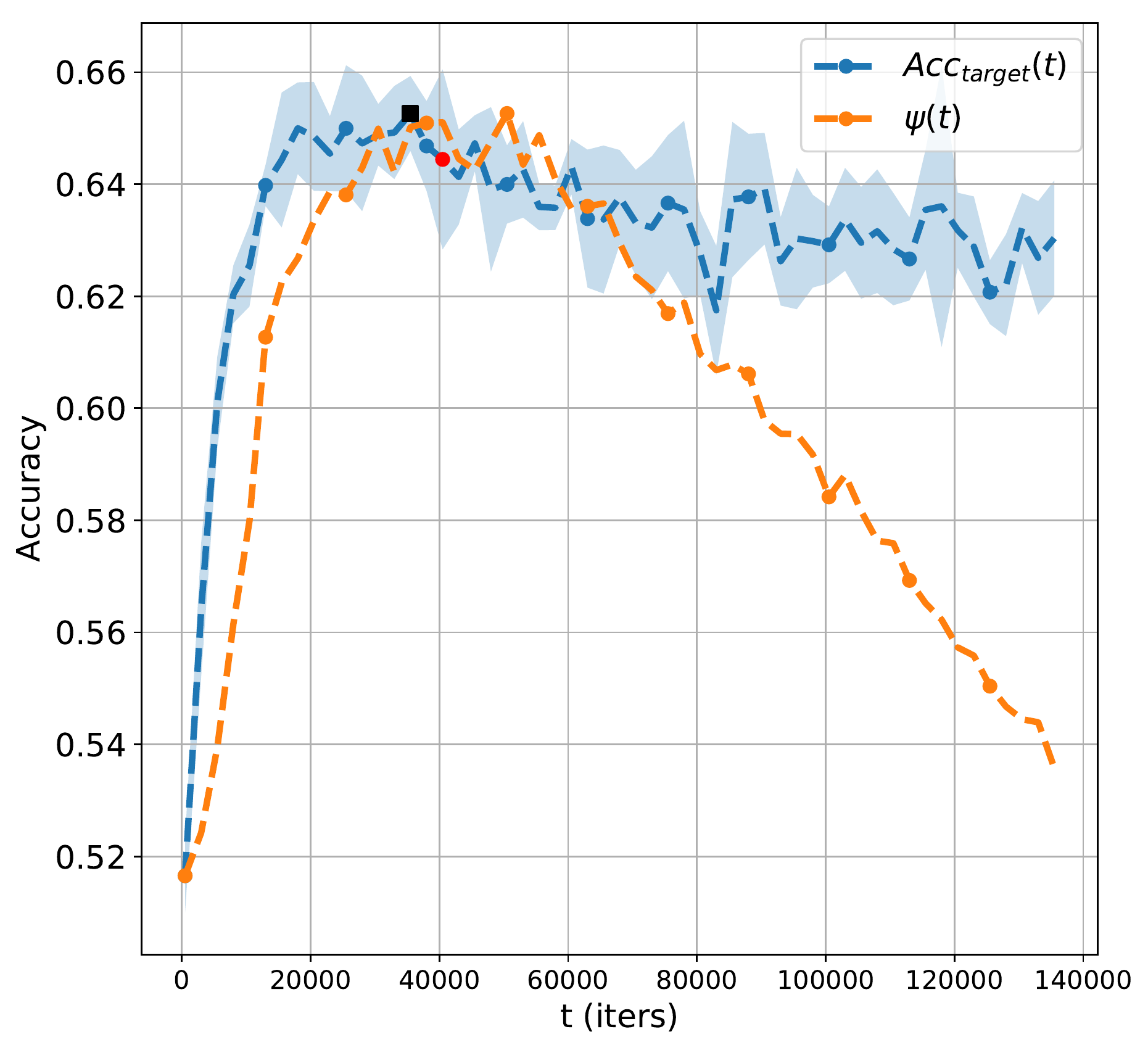}
    }
    \subfloat[Expected Feature-Wise Variance]{%
        \includegraphics[width=0.25\linewidth]{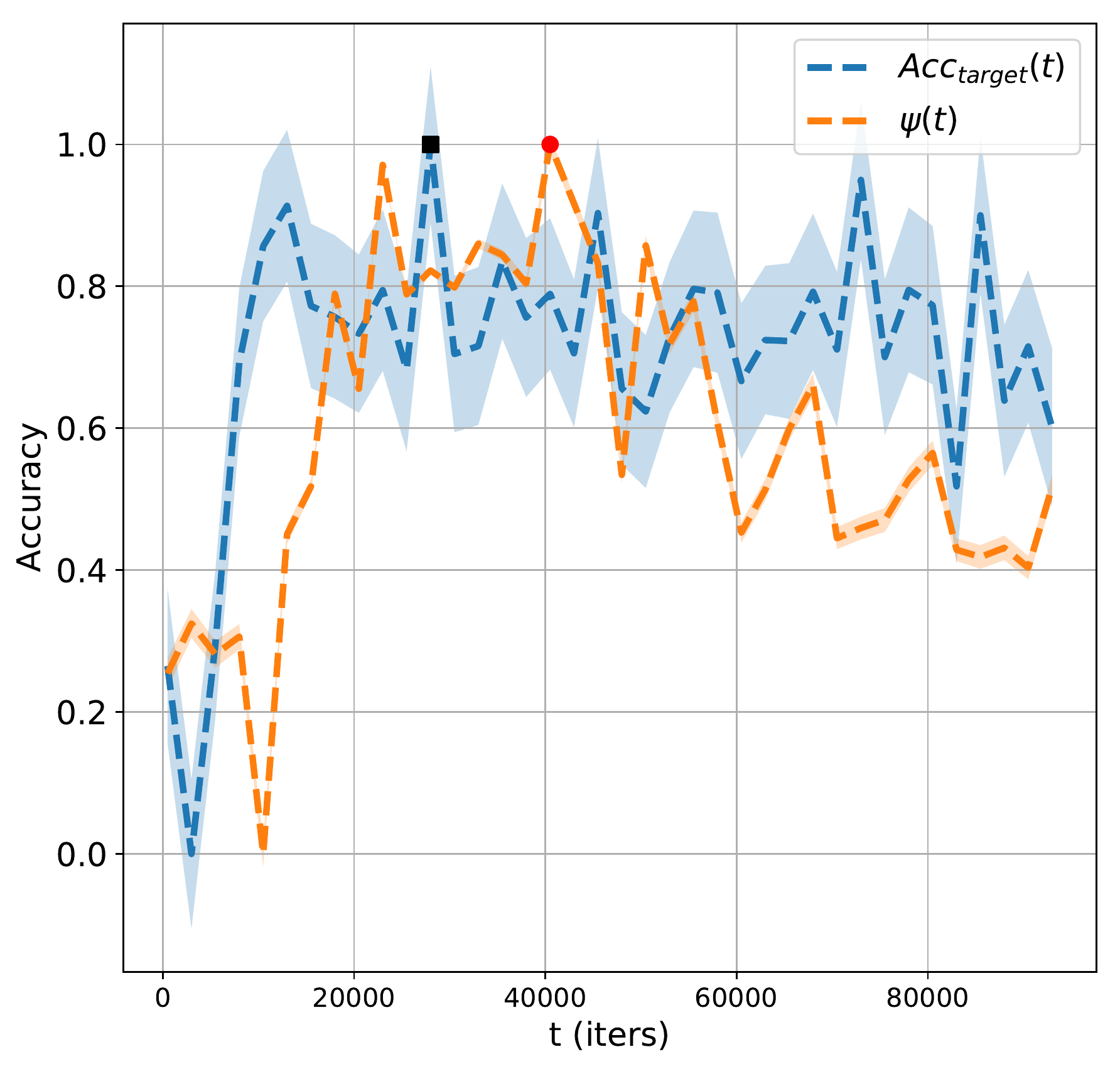}
    }    
    \caption{Different metrics of the representation space may have a strong correlation with generalization, other than the expected inner product. a) Prototypical, VGG Flower, 5-way 1-shot : out of three metrics which in other cases may be related with generalization (as in b), c), d)), here only the expected $l_2$ dispersion has a strong relation with generalization. b) Expected $l_2$ norm; c) Expected square $l_2$ dispersion, Prototypical Network, VGG Flower; d) Expected feature-wise variance, Prototypical Network, Omniglot to Quickdraw. These results motivate our approach of considering a family of functions $\psi$ (the linear combinations of $\hat{m}_1$ to $\hat{m}_4$) in which we must find the optimal function $\psi^*$ given the setting, rather than trying to discover a single universal metric that would correlate to generalization in all scenarios.}
    \label{fig:various_metrics}
\end{figure}

\subsubsection{Critical layers and critical moments}\label{sec:appendix:exp_results:critical_layer_and_moment}

\begin{figure}[H]
    \centering
    \subfloat[Critical layers]{%
        \includegraphics[width=0.25\linewidth]{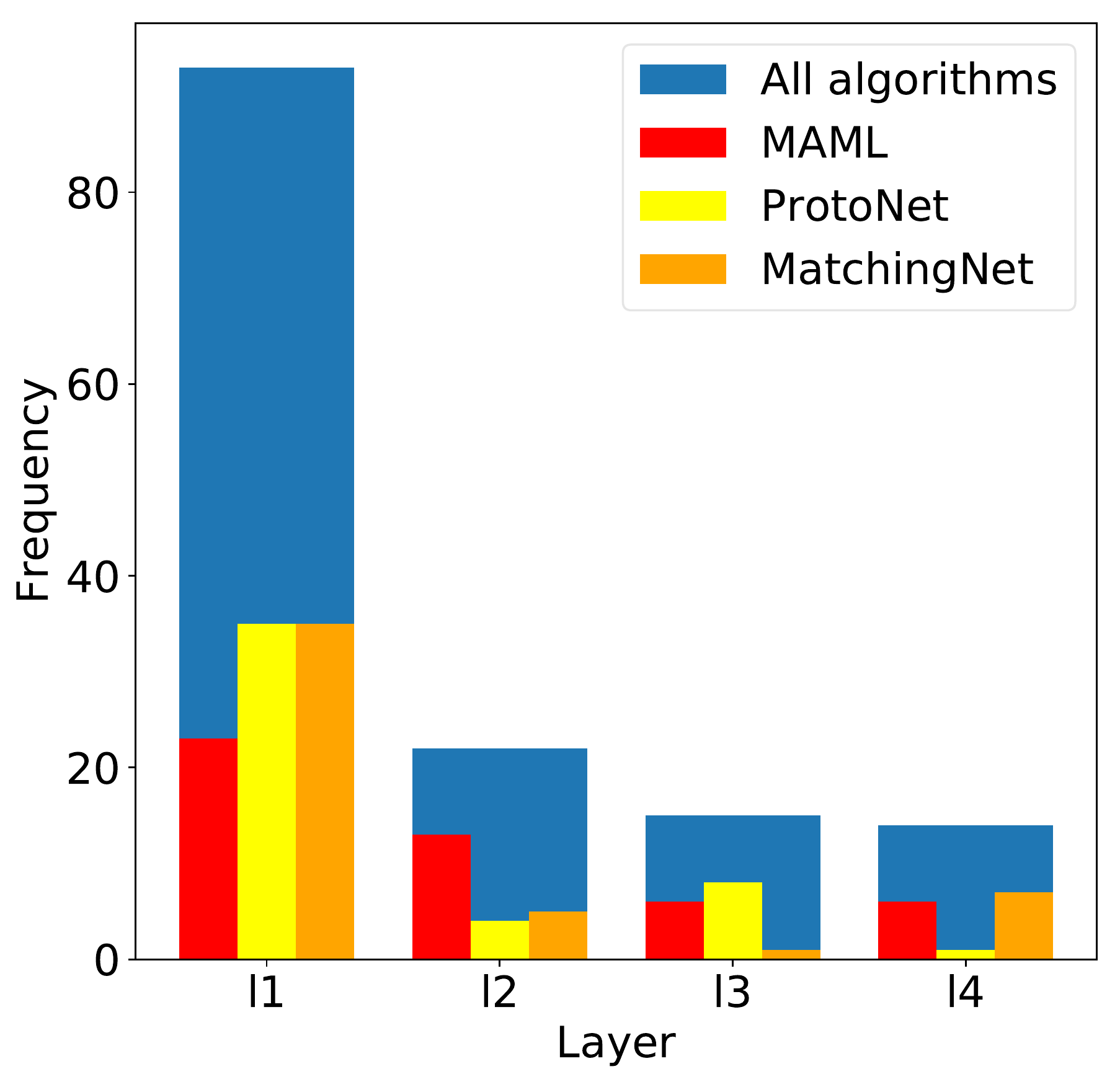}
        \label{fig:hist:critical_layers}
    }
    \hspace{0.03 in}
    \subfloat[Critical moments]{%
        \includegraphics[width=0.25\linewidth]{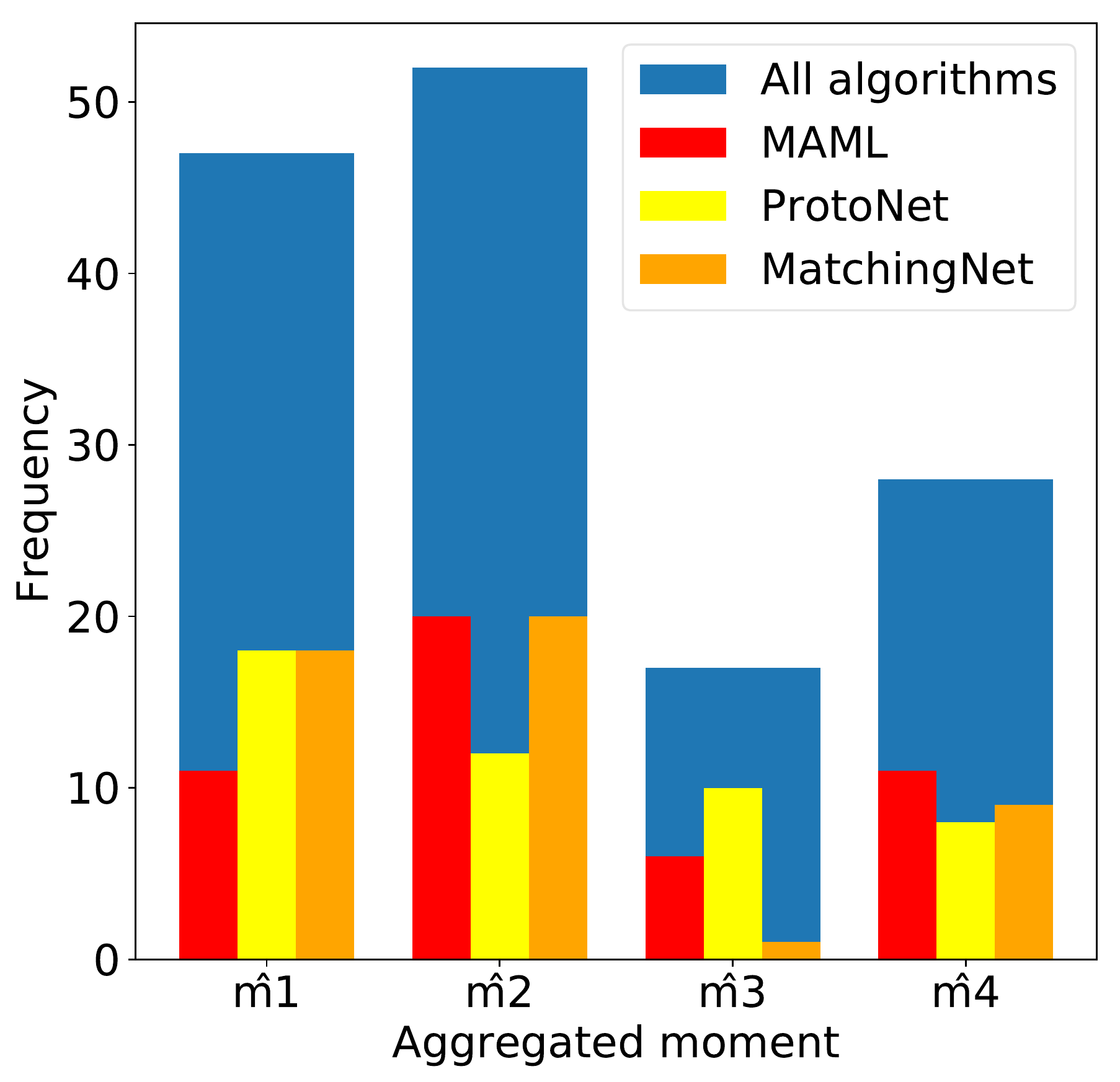}
        \label{fig:hist:critical_moments}
    }
    \caption{Critical layers and critical moments histograms, across all experiments. For a given experiment, the critical layer is the one where target and source activation distribution trajectories diverged the most. Results show that any layer of the feature extractor may be critical, but they happen more often at shallow depths (a). Moreover, the critical moment is the one by which the target and source activation distribution trajectories diverged the most. Results also indicate that all moments may be critical (b).}\label{fig:hist:critical_layers_and_moments}
\end{figure}

In Fig.~\ref{fig:hist:critical_layers_and_moments} we present observations on the frequency, across all of our experiments, of which layer and which aggregated moments are critical, \textit{i.e.} where we observed the strongest divergence between the trajectories of the target and source activations. In Fig.~\ref{fig:hist:critical_layers}, we observe that the strongest divergence may happen at all layers, but tends to happen more frequently in earlier layers. This further motivates the idea of ``looking under the hood'' to infer how target generalization is evolving, rather than simply looking at the representations of the last layer. Moreover, Fig.~\ref{fig:hist:critical_moments} suggests that all of the aggregated moments that we introduced, from $\hat{m}_1$ to $\hat{m}_4$, might be critical in driving the divergence between the activation trajectories. This further motivates our use of these four moments to characterize the trajectories of neural activations.

\subsection{Performance on Residual Networks}\label{sec:appendix:exp_results:resnet}

We added experiments with an additional architecture: a deep residual network of 18 layers (ResNet-18), as used in the original Meta-Dataset paper. We used the few-shot transfer learning setting where Mini-Imagenet is used for meta-training (source), and Omniglot is the target dataset. We ran this experiment for all meta-training algorithms considered in our work, namely: MAML, prototypical network, and matching network. Our experiments show that our method outperforms validation-based early-stopping, for all three algorithms. See Tab. \ref{tab:resnet}.

\begin{table}[H]
\centering
\begin{tabular}{|cccc|}
\hline
\multicolumn{4}{|c|}{ResNet-18} \\ \hline
\multicolumn{1}{|c|}{Acc (\%)} & \multicolumn{1}{c|}{MAML} & \multicolumn{1}{c|}{Proto Net} & Matching Net \\ \hline
\multicolumn{1}{|c|}{Optimal} & \multicolumn{1}{c|}{55.93} & \multicolumn{1}{c|}{66.86} & 64.93 \\ \hline \hline
\multicolumn{1}{|c|}{\begin{tabular}[c]{@{}c@{}}Validation \\ Baseline\end{tabular}} & \multicolumn{1}{c|}{52.38} & \multicolumn{1}{c|}{61.36} & 57.30 \\ \hline
\multicolumn{1}{|c|}{ABE (ours)} & \multicolumn{1}{c|}{\textbf{55.40}} & \multicolumn{1}{c|}{\textbf{62.50}} & \textbf{62.88} \\ \hline
\end{tabular}
\caption{Generalization performance of ABE, using a ResNet-18 as the neural architecture. For each setting, ``Optimal" is the maximum achievable generalization performance, if early-stopping had been optimal (with an oracle). We used Mini-Imagenet as the source dataset and Omniglot as the target dataset. The first row shows the maximum achievable target accuracy. Results show that for each source dataset and Meta-Learning algorithm, our method consistently outperforms the validation baseline.}
\label{tab:resnet}
\end{table}

\subsection{The issue of using a validation set for early-stopping in meta-learning}\label{sec:appendix:exp_results:issue_meta-val}

\begin{figure}[H]
\centering
    \subfloat[MAML - CU Birds]{%
        \includegraphics[width=0.33\linewidth]{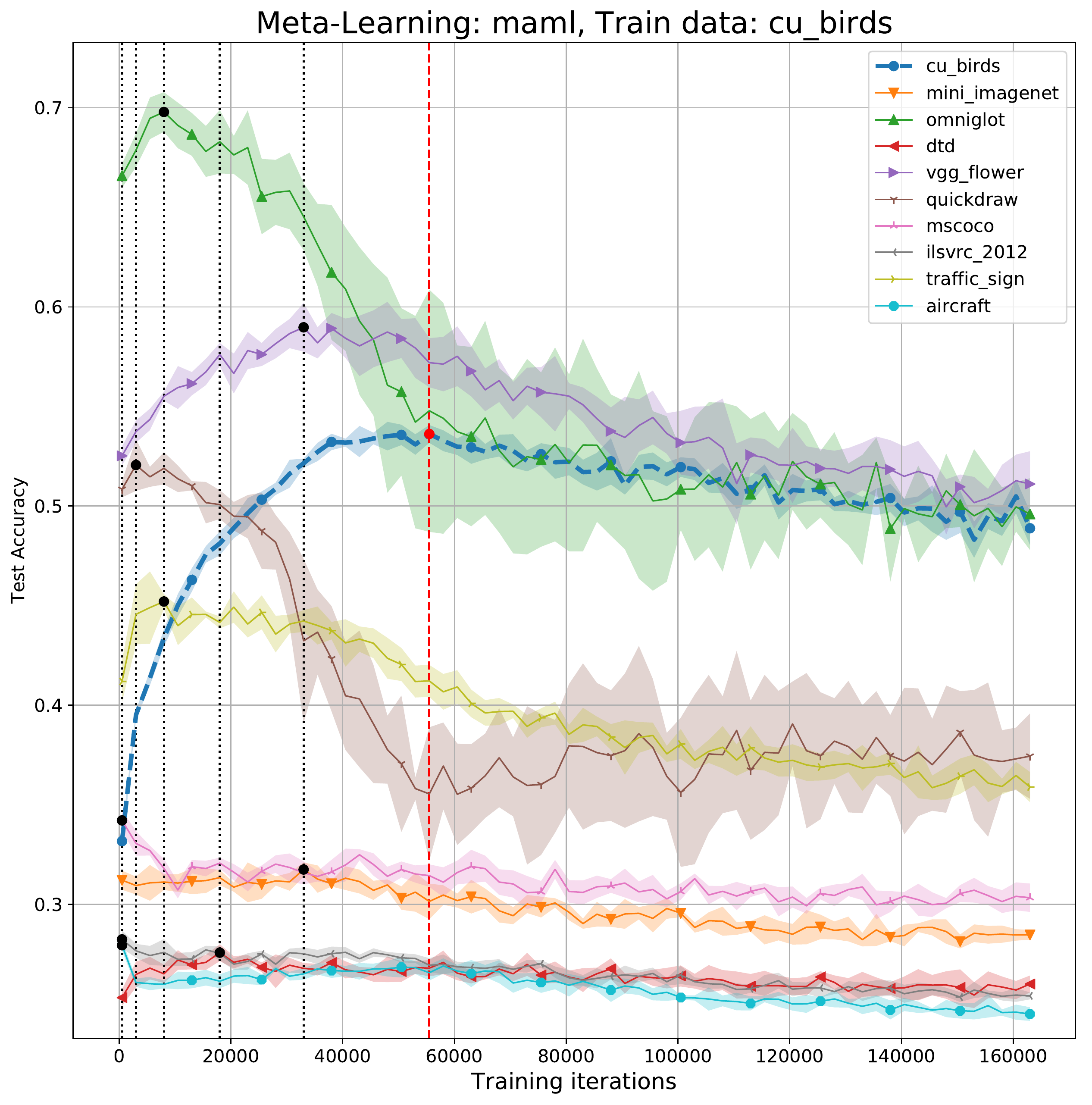}
        \label{fig:exp:issue_meta-val:maml:cu_birds}
    }
    \subfloat[Matching Network - CU Birds]{%
        \includegraphics[width=0.33\linewidth]{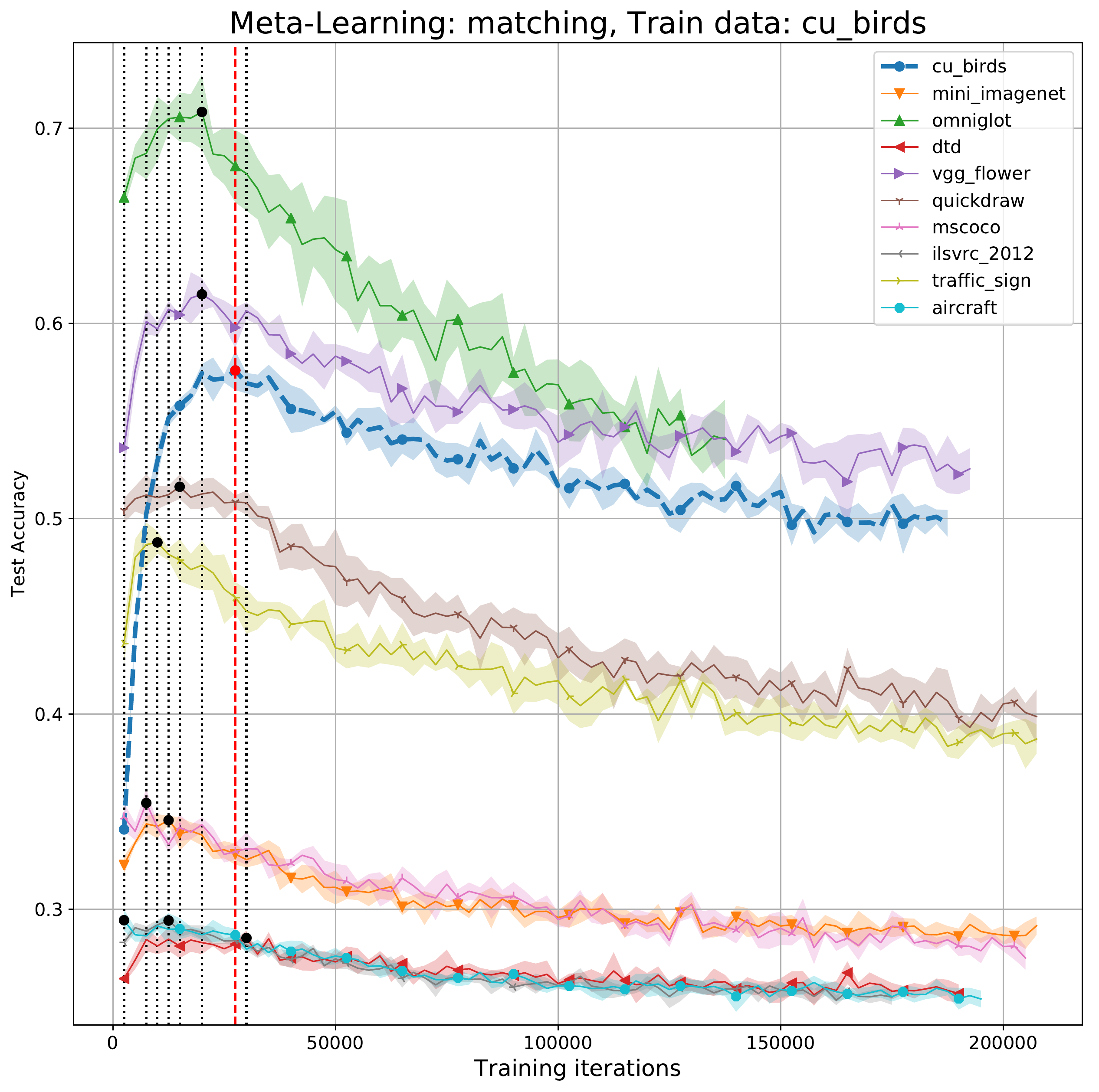}
        \label{fig:exp:issue_meta-val:mn:cu_birds}
    }
    \subfloat[Prototypical Network - CU Birds]{%
        \includegraphics[width=0.33\linewidth]{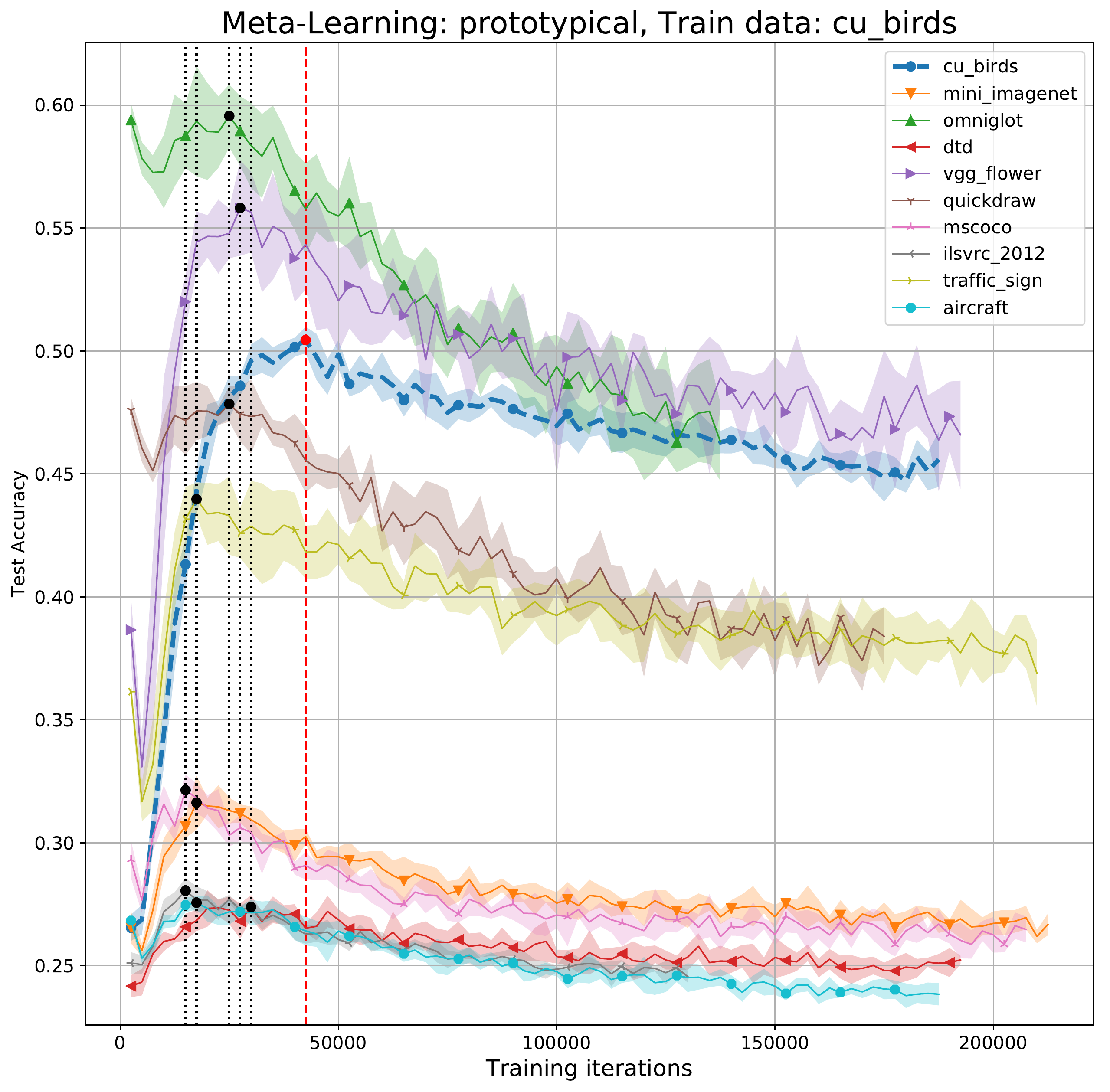}
        \label{fig:exp:issue_meta-val:pn:cu_birds}
    } \newline
    \subfloat[MAML - MiniImagenet]{%
        \includegraphics[width=0.33\linewidth]{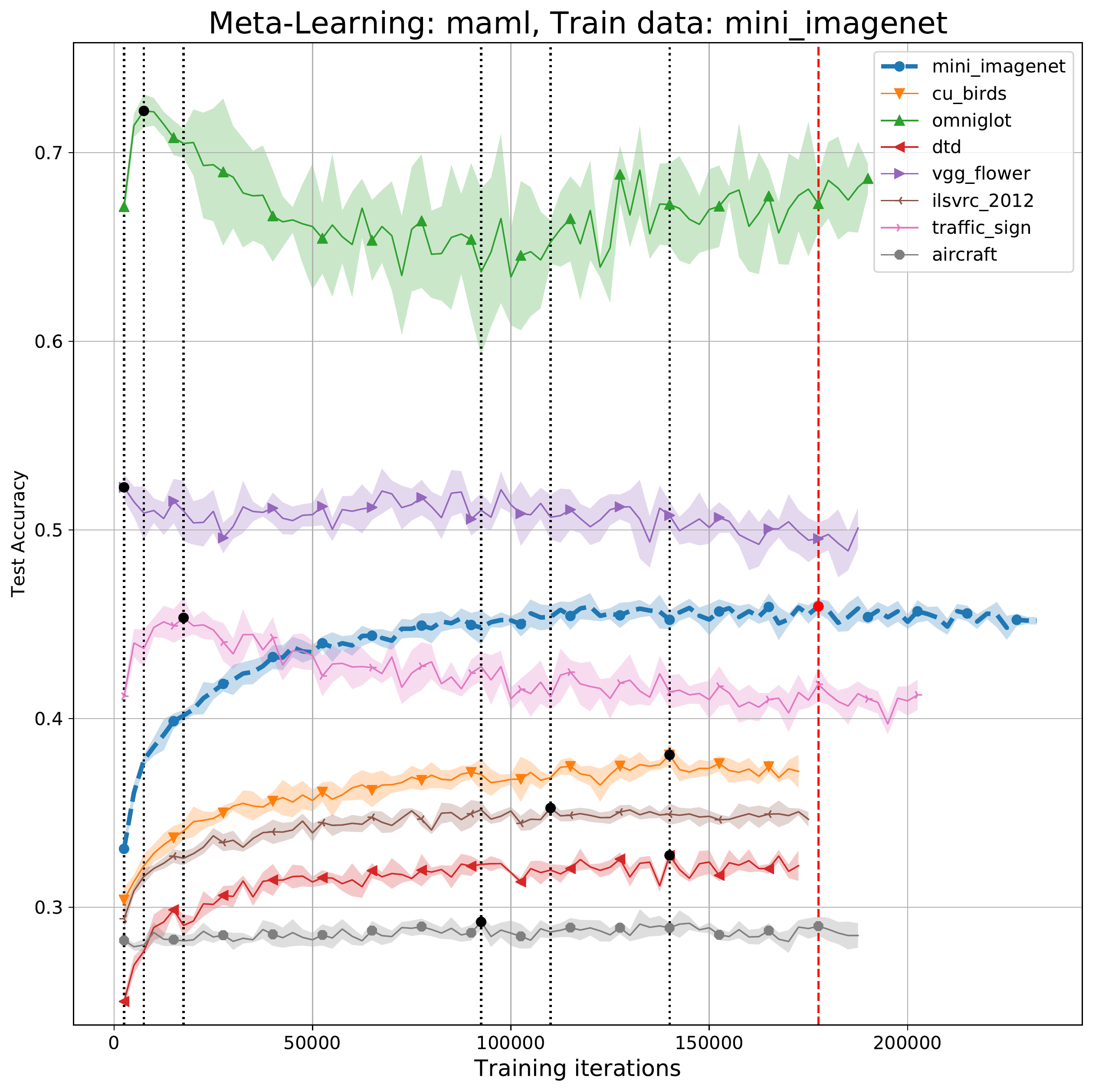}
        \label{fig:exp:issue_meta-val:maml:miniimagenet}
    }
    \subfloat[Matching Network - MiniImagenet]{%
        \includegraphics[width=0.33\linewidth]{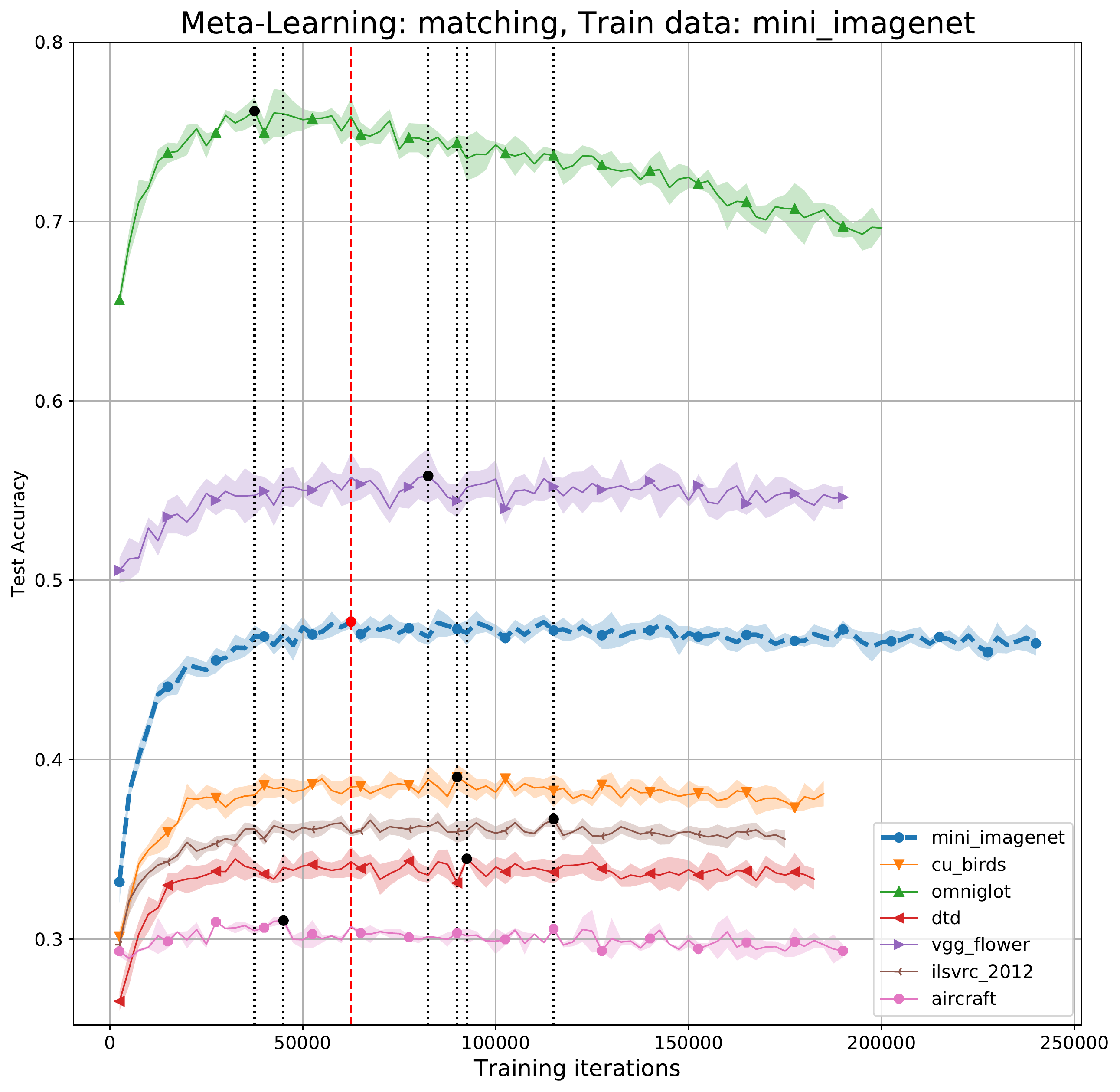}
        \label{fig:exp:issue_meta-val:mn:miniimagenet}
    }
    \subfloat[Prototypical Network - MiniImagenet]{%
        \includegraphics[width=0.33\linewidth]{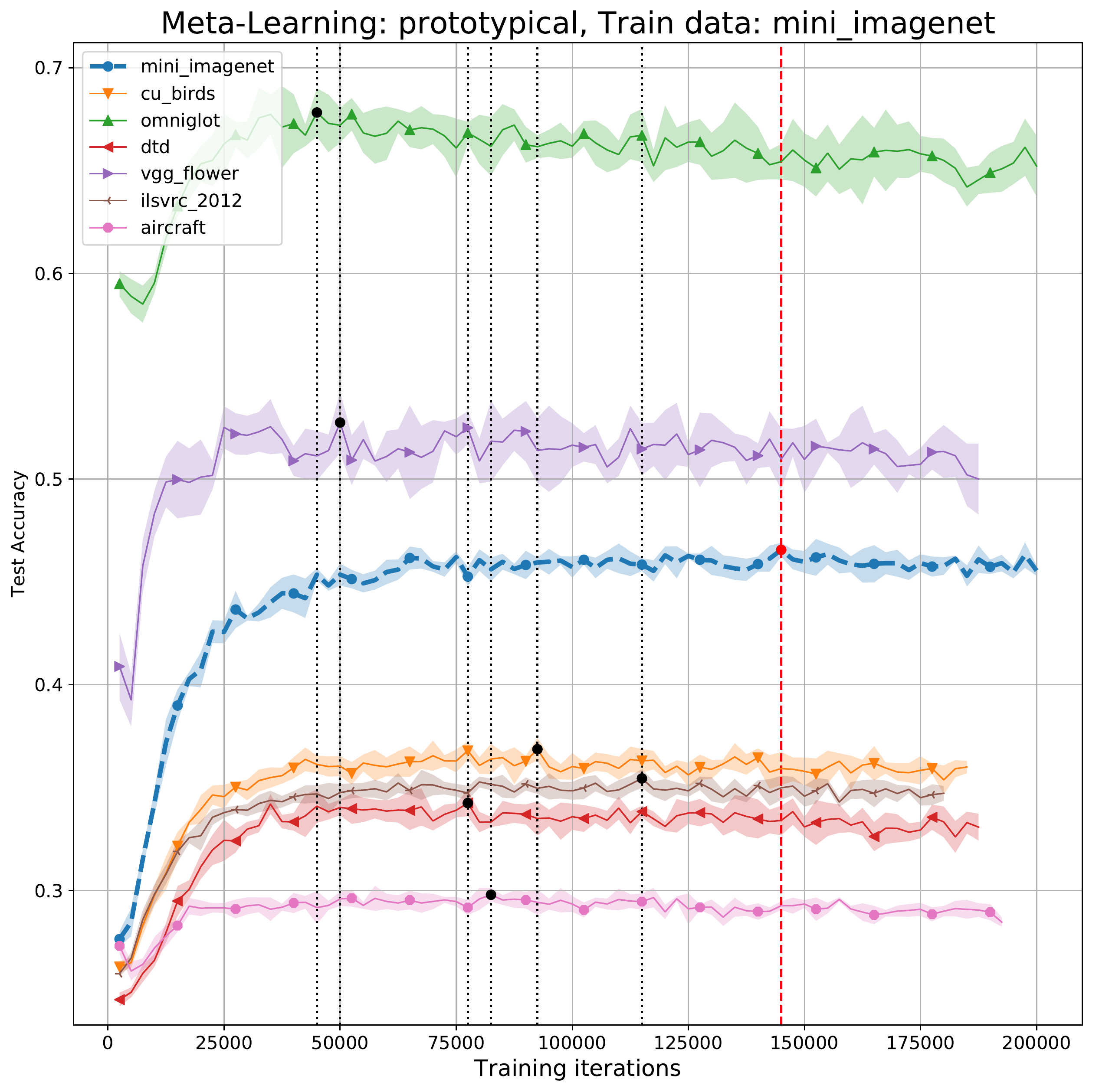}
        \label{fig:exp:issue_meta-val:pn:miniimagenet}
    }
\caption{Illustrating the issue of using meta-validation for early-stopping in Meta-Learning: MAML, Matching Network, and Prototypical Network, trained on both MiniImagenet and CU Birds, with various target datasets from the Meta-Dataset benchmark. The validation early-stopping time $t^*_{valid}$ (red dashed line) leads to sub-optimal generalization performances for the various target datasets (from the Meta-Dataset benchmark), each having their own optimal early-stopping time $t^*$ (black dashed lines) at different times. We also observe $t^* \leq t^*_{valid}$ across the different settings.}
\label{fig:exp:issue_meta-val}
\end{figure}

As Fig. \ref{fig:exp:issue_meta-val} suggests, the optimal stopping time for different target datasets can vary significantly. Here we quantitatively evaluate how they vary. For each experiment, i.e. each tuple (algorithm, source dataset), we computed the stopping times on the different target datasets, measured their standard deviation, and report this deviation as a percentage of the validation stopping time. Results show that those stopping times vary significantly, as some target datasets may peak very early on, while others can peak close to the validation stopping time. See the Tab. \ref{tab:variability_target_stopping_times}, where we summarize those results on a per-algorithm basis for compactness of the presentation. Those results suggest that using some extra ood dataset for validation (and early-stopping) may still to poor generalization, because of the large discrepancies between the different optimal stopping times.

\begin{table}[H]
\centering
\begin{tabular}{|c|c|c|c|}
\hline
Algorithm & MAML & Proto Net & Matching Net \\ \hline
\begin{tabular}[c]{@{}c@{}}Standard deviation\\ of target stopping-time\\ relative to validation stopping-time\\ (averaged over all source datasets)\end{tabular} & 38.97 \% & 30.05 \% & 20.77 \% \\ \hline
\end{tabular}%
\caption{Variation of the optimal stopping times of the different target datasets, considering a fixed source dataset. For each experiment, i.e. each tuple (algorithm, source dataset), we computed the stopping times on the different target datasets, measured their standard deviation, and report this deviation as a percentage of the validation stopping time. This variations is significant, and show that for a fixed source dataset, and a target dataset of interest, even if we had access to an extra dataset to use as a proxy for our true target dataset (because it is ood), its stopping time may be significantly different than the optimal stopping time of our target dataset of interest.}
\label{tab:variability_target_stopping_times}
\end{table}

\subsection{Additional baseline : Early-stopping on a third distribution}

Another question arises: what happens if we perform early-stopping using a third distribution (assuming we have access to it)? Should its distributional shift with the training distribution result in a better early-stopping time?
Having access to some extra dataset for validation should be considered as an oracle method, since we cannot assume easy availability of that extra dataset. Nevertheless, for the sake of analysis, we computed how such oracle method, which we call the “ OOD-Validation Oracle “, would perform in terms of generalization. For each experiment, i.e. each tuple (algorithm, source dataset, target dataset), we measured, on average, what is the performance when early-stopping from another dataset than the target dataset. Thus for each experiment, we average the performance over all such extra datasets, and we compute the standard deviation of those performances, since the extra datasets will have different stopping times. We then average this over all experiments, and present them on a per-algorithm basis, for compactness of the presentation. See Tab. \ref{tab:ood-validation_oracle}. The results show that, even with this oracle, our method generally performs better, as ABE shows better average performance for two out of the three algorithms, and shows less variance in its performance (for ABE, within each experiment, we compute the standard deviation across all the 50 target tasks that are used to test our method for early-stopping). This highlights the importance of considering the specific target problem at hand.

\begin{table}[H]
\centering
\begin{tabular}{|c|c|c|c|}
\hline
Acc (\%)              & MAML           & Proto Net      & Matching Net   \\ \hline
OOD-Validation Oracle & 39.59 $\pm$ 1.06 & 37.05 $\pm$ 0.68 & 42.48 $\pm$ 0.56 \\ \hline
\textbf{ABE (ours)}   & 40.51 $\pm$ 0.35 & 37.35 $\pm$ 0.13 & 42.09 $\pm$ 0.38 \\ \hline
\end{tabular}
\caption{OOD-Validation Oracle: Standard deviations for the oracle are computed on the obtained generalization performances, across the different ood dataset being used for early-stopping, then averaged across all experiments and presented on a per-algorithm basis. For results for ABE, the standard deviations in performances are computed across the different independent target tasks being used for early-stopping, and those deviations are averaged across all experiments. Those results show that our method generally performs better than the oracle, and show less variability in performance, meaning that using a single task gives a steady performance.}
\label{tab:ood-validation_oracle}
\end{table}

\subsection{Additional baseline : Support loss of a single task}\label{sec:appendix:exp_results:baseline_single-task_support_loss}

In addition, since ABE has access to the unlabelled data from a single target task, we compare its performance to another baseline : tracking the support loss of the single target task, after fine-tuning on its  examples (since we use 1-shot, we can randomly label the examples). This captures how well can the model classifies the support examples of the task. We use the cross-entropy loss instead of the classification accuracy which can easily saturate even after a single step. We performed this analysis using MAML (as it is not applicable with the other two algorithms). We used all of the source-target dataset pairs considered so far. For each setting, we use, just like with ABE, a fixed set of 50 target tasks, and keep their sets of support examples. At the end of each training epoch, for each of the 50 tasks, we feed the support examples to the model to get predictions. Since we use 1-shot of examples, we randomly label them,  compute the support loss, and finetune the model (only the final classification layer, and freeze the feature extractor). After finetuning, we reevaluate the support loss using the adapted parameters of the classifier. Thus for each task, we track this loss throughout the entire training experiment, we early-stop at the minimum value of the loss, and evaluate the target generalization at that point in time. We average this performance over the 50 target tasks.

\begin{figure}[H]
    \centering
    \includegraphics[width=0.5\linewidth]{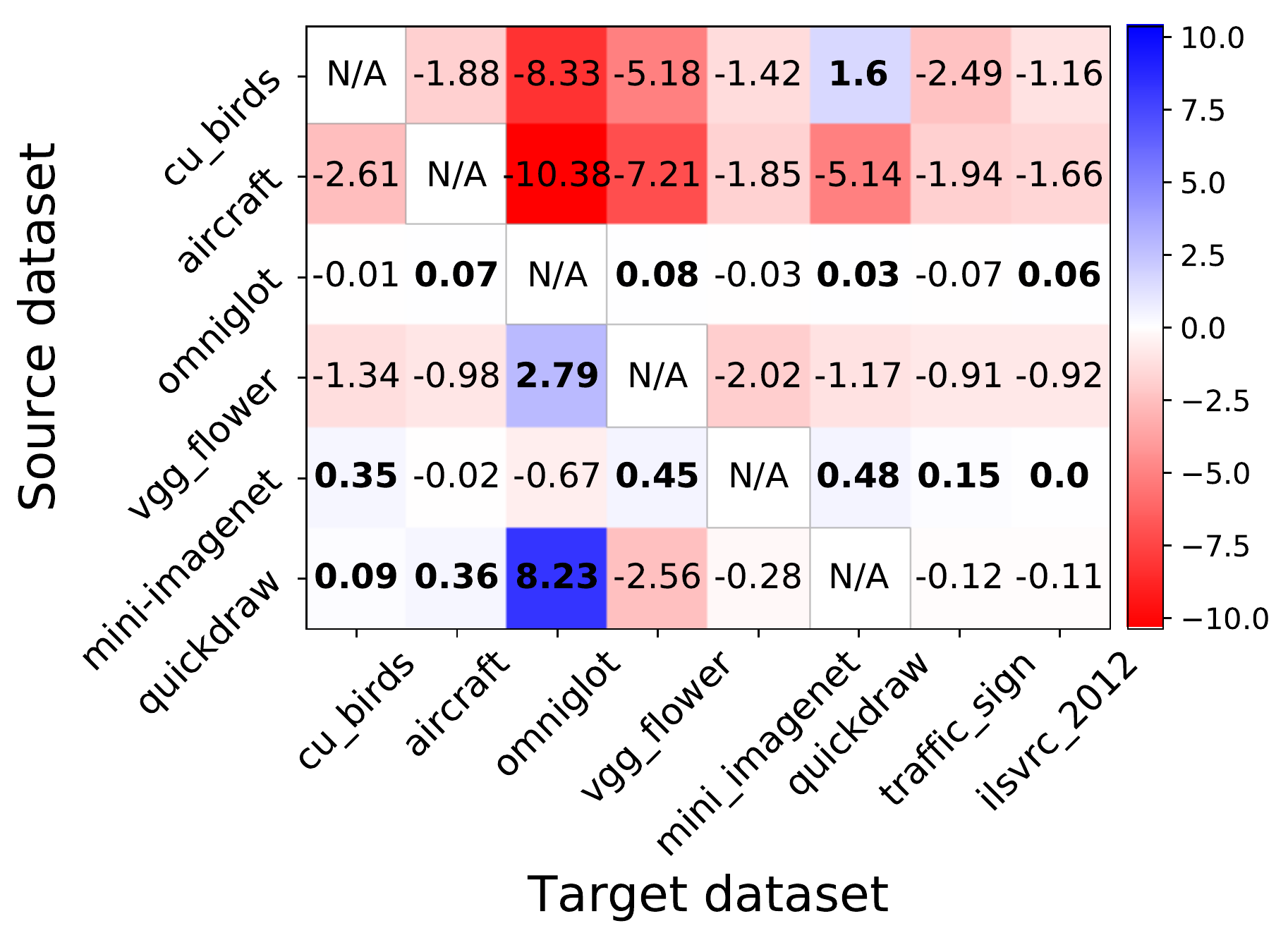}
    \caption{Performance when using the support loss of a single target task for early-stopping. Difference in generalization performance between this baseline and the baseline of validation early-stopping.}
    \label{fig:performance_heatmap:single-task_baseline:horizontal}
\end{figure}

Results show that this baseline doesn't not perform as well as ABE, and in fact performs worse than the validation-based early-stopping, as it increases the generalization gap by 54.1\% on average, while with MAML, our method closes the generalization gap by 71.4\% on average. We also observed a much greater variance between the estimated stopping times for this baseline, when using different tasks, and thus a higher standard deviation in performance of 0.74\% (in accuracy, on average) compared to only 0.35\% for ABE used with MAML. See results in Fig. \ref{fig:performance_heatmap:single-task_baseline:horizontal} where there are mostly degradation of generalization (red color) and very few instances of improvement (blue color), as opposed to the left most subfigure of  Fig. \ref{fig:performance_heatmap:method:horizontal} of Sec. \ref{sec:performance}.

 This reinforces some of the insights of our work -- to infer target generalization, one might have to " look under the hood " and inspect the activations at the lower layers of the feature extractor. We also observed a much greater variance between the estimated stopping times for this baseline, when using different tasks, as compared to ABE.

\subsection{Using more examples to perform activation-based early-stopping}\label{sec:appendix:exp_results:ABE_more_data}

\begin{figure}[H]
\centering
    \subfloat[Mean Accuracy]{%
        \includegraphics[width=0.5\linewidth]{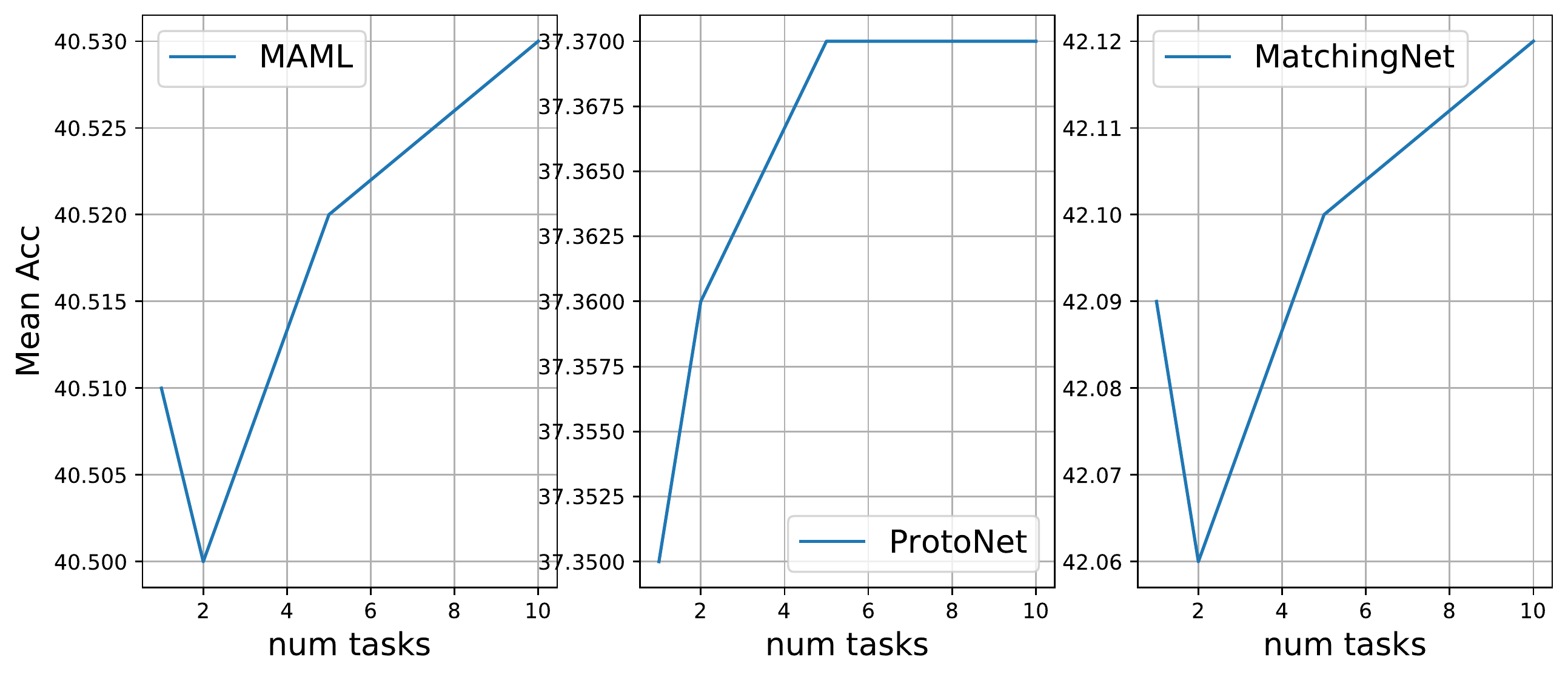}
        \label{fig:exp:more_data:acc_mean}
    }
    \subfloat[STD of Accuracy]{%
        \includegraphics[width=0.5\linewidth]{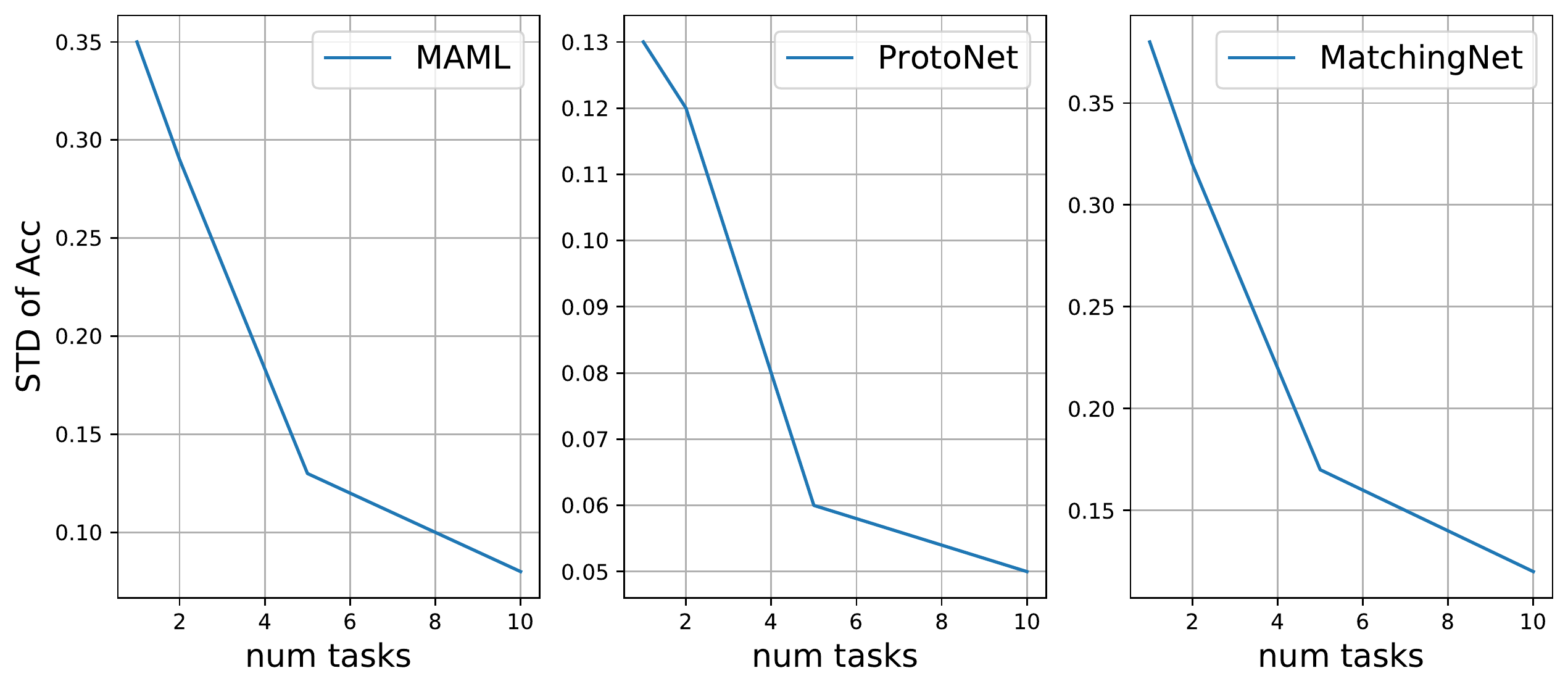}
        \label{fig:exp:more_data:acc_std}
    }
\caption{Mean and standard deviation performance of increasing the number of examples from additional tasks while using ABE.}
\label{fig:exp:more_data}
\end{figure}

We also study the effects of using more examples for ABE. Results are shown in Fig. \ref{fig:exp:more_data}. We observe that using examples from a larger pool of tasks tends to improve performance while reducing variance. Despite these promising results, using additional examples increases computational complexity and assumes access to several target tasks, which might not be always practical.

\end{document}

%% file: math_commands.tex
\usepackage{amsmath,amsfonts,bm}

\def\eqref#1{equation~\ref{#1}}

\def\1{\bm{1}}

\DeclareMathAlphabet{\mathsfit}{\encodingdefault}{\sfdefault}{m}{sl}
\SetMathAlphabet{\mathsfit}{bold}{\encodingdefault}{\sfdefault}{bx}{n}

